\documentclass[10pt,twocolumn,letterpaper]{article}

\usepackage{cvpr}              
\definecolor{cvprblue}{rgb}{0.21,0.49,0.74}
\usepackage[pagebackref,breaklinks,colorlinks,citecolor=cvprblue]{hyperref}

\usepackage{hyperref}
\usepackage{url}
\usepackage{amsmath}
\usepackage{amssymb}
\usepackage{wrapfig}
\usepackage{enumitem}
\usepackage{algorithm}
\usepackage{algpseudocode}
\usepackage{varwidth}          
\usepackage{xcolor}            
\usepackage{svg}
\usepackage{multirow}
\usepackage{booktabs}
\usepackage{subcaption}
\usepackage{graphicx}
\usepackage{colortbl}
\usepackage[normalem]{ulem} 
\usepackage{makecell}
\usepackage{siunitx}
\useunder{\uline}{\ul}{}
\usepackage{booktabs}
\usepackage{adjustbox}

\newcommand{\expt}{\mathop{\mathbb{E}}}

\def\paperID{} 

\title{Rethinking Preference Alignment for Diffusion Models\\ with Classifier-Free Guidance}

\author{
Zhou Jiang$^{1}$ \quad Yandong Wen$^{1}$ \quad Zhen Liu$^{2}$\thanks{Corresponding Author.} \\
$^{1}$Westlake University 
$^{2}$The Chinese University of Hong Kong, Shenzhen
}

\begin{document}
\maketitle

\begin{abstract}
Aligning large-scale text-to-image diffusion models with nuanced human preferences remains challenging. While direct preference optimization (DPO) is simple and effective, large-scale finetuning often shows a generalization gap. We take inspiration from test-time guidance and cast preference alignment as classifier-free guidance (CFG): a finetuned preference model acts as an external control signal during sampling. Building on this view, we propose a simple method that improves alignment without retraining the base model. To further enhance generalization, we decouple preference learning into two modules trained on positive and negative data, respectively, and form a \emph{contrastive guidance} vector at inference by subtracting their predictions (positive minus negative), scaled by a user-chosen strength and added to the base prediction at each step. This yields a sharper and controllable alignment signal. We evaluate on Stable Diffusion 1.5 and Stable Diffusion XL with Pick-a-Pic v2 and HPDv3, showing consistent quantitative and qualitative gains. Code and models will be made publicly available from \href{https://github.com/UGVly/PreferenceGuidanceDiffusion}{the URL}.
\end{abstract}

\section{Introduction}
\label{sec:intro}

Diffusion models~\citep{ddpm, song2019generative, song2021score} 
are one of the most prevalent generative models
for high-fidelity text-to-image (T2I) synthesis~\citep{sdxl, saharia2022photo}.
These models are typically trained from Internet-scale datasets which, due to the tremendous scale, are not carefully curated. A diffusion model pretrained on these datasets therefore deviate from what humans (in the majority voting sense) truly prefer in aspects such as aesthetic and instruction following~\citep{kirstain2023pick}.


The same issue is well studied in the field of large language model (LLM), in which na\"ively pretrained LLMs without any post-training steps do not follow instructions and are not able to chat naturally with human~\citep{ouyang2022human} . Typical approaches to align LLMs with human preference for LLMs include 1) reinforcement learning from human feedback (RLHF)~\citep{ouyang2022human}, which demands a reward model pretrained on a preference dataset and requires careful hyperparameter tuning, and 2) direct preference optimization (DPO)~\citep{rafailov2023direct}, the simpler alternative that bypasses reward modeling by essentially treating the alignment problem as a binary classification problem on positive-negative preference pairs. This simple solution of DPO can be easily adapted for aligning diffusion models with human preference~\citep{wallace2024diffusion} and has been widely used in applications other than T2I synthesis~\citep{wang2023mesh,blattmann2023stablevideo,wu2023tuneavideo,khachatryan2023text2video} . Nevertheless, DPO is generally considered less robust compared to RLHF: it is prone to overfitting, may produce non-smooth predictions on out-of-distribution text prompts, and even exhibit catastrophic forgetting behaviors~\citep{lin-etal-2024-limited}. While one may include in either the whole pretraining dataset or just the prompt set to regularize models, access to these pretraining sets is typically infeasible for large-scale models.

\begin{figure}
    \centering

    \includegraphics[width=\linewidth]{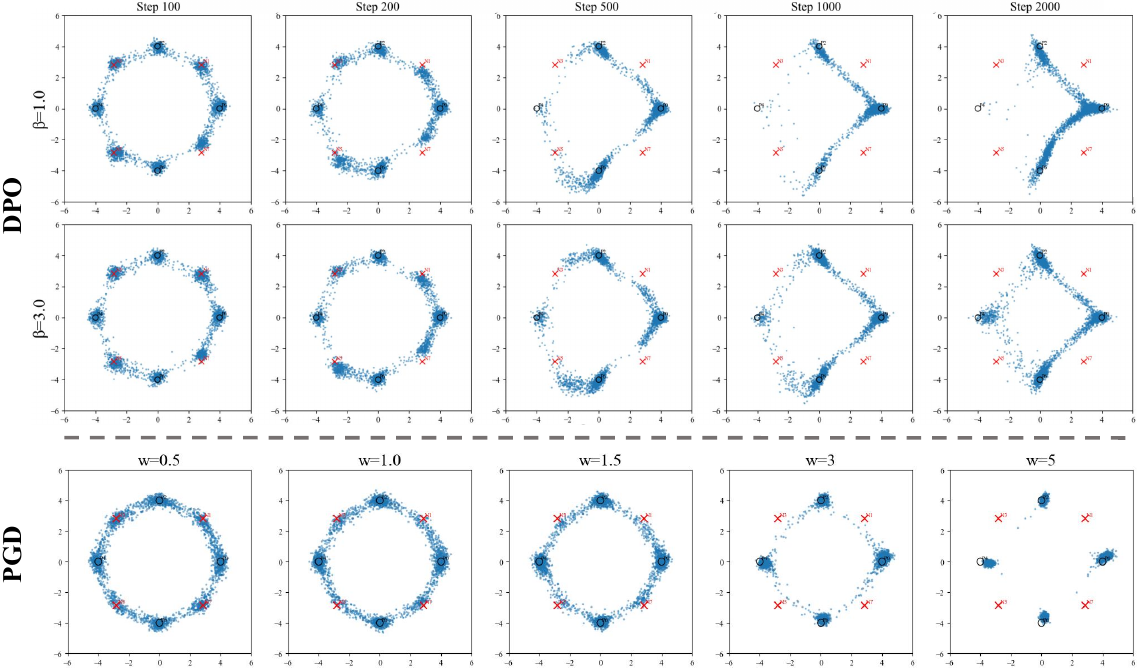}\\

\captionsetup{labelfont=footnotesize}
    \caption{\footnotesize Toy 2D experiment on DPO (top) and our proposed PGD (below) to demonstrate the overfitting issue in DPO training. Black circles indicate positive sample clusters and red crosses indicate negative sample clusters. $w$ is the guidance weight of PGD and $\beta$ is the DPO scale parameter.}
    \label{fig:toy_exp}
\end{figure}

We conduct a toy 2D experiment (Fig.~\ref{fig:toy_exp}) to illustrate how DPO fails. Specifically, we consider a toy 8-Gaussians dataset composed of 2D points sampled from 8 Gaussian balls, for which we label half of the Gaussian balls as positive samples and the rest of them as negative ones. With this dataset, we train an simple diffusion model parameterized by a 3-layer-MLP to perform Diffusion-DPO finetuning with randomly sampled preference pairs. Even on this toy dataset, DPO-tuned models deviate from ideal finetuned distributions, and, if trained for too long, easily suffers from overfitting and mode collapse.



We instead take inspiration from inference-time techniques for diffusion model adaptation. Specifically, we observe that classifier free guidance (CFG)~\citep{dhariwal2021diffusion}, the standard approach for sampling from conditional diffusion models by linearly combining between unconditional and conditional predictions, can be viewed as tempering the potentially overfitted conditional model with the more generalizable unconditional prior. Since the posterior distribution obtained through CFG typically exhibits strong performance, and the alignment objective from the control as inference perspective~\citep{levine2018reinforcement} is likewise to obtain a posterior distribution~\citep{rafailov2023dpo}, we are led to ask: \emph{can CFG be adopted to address the diffusion alignment problem?}



Motivated by this question, we view a finetuned diffusion model as the diffusion model conditioned on a virtual control signal from the preference dataset, while the base model serves as the unconditional model. From this perspective, sampling from the aligned diffusion model naturally becomes a CFG-style inference process, which gives rise to our first method, Preference-Guided Diffusion (PGD). With the guidance weight to amplify the difference between the control signal and the prior during test time, the control signal does not have to be a fully-finetuned model but one finetuned only with a few iterations, thus effectively preventing overfitting, as illustrated in the bottom row of Fig.~\ref{fig:toy_exp}.
Adopting this CFG perspective further suggests that finetuning should resemble conditional diffusion training, which does not rely on positive–negative pairs but instead uses the standard diffusion loss. To implement this idea, we finetune two models independently, one that generates positive samples and another that generates negative samples, and combine them at inference through CFG-style composition. We refer to this variant as contrastive Preference-Guided Diffusion (cPGD). In experiments, PGD and cPGD consistently outperform vanilla Diffusion-DPO. Notably, both methods achieve Pareto improvements, simultaneously yielding higher reward, lower FID and greater diversity in generated samples. Moreover, the approach in principle produces transferable plug-and-play modules that, once trained from base diffusion models, can be reused to align others.


In summary, our contributions are
\begin{itemize}[leftmargin=*,nosep]
\setlength\itemsep{0.39em}
    \item We propose to alleviate the generalization issue in Diffusion-DPO by treating diffusion model alignment as a special case of CFG-style inference.
    \item We introduce Preference-Guided Diffusion (PGD), which aligns the generated distribution with human preference through CFG-style guidance at inference time.
    \item We extend this view by considering finetuning as conditional diffusion training and propose contrastive PGD (cPGD).
    \item We empirically demonstrate that both variants achieve Pareto improvements over the Diffusion-DPO baseline.
\end{itemize}

\section{Related Work}

\noindent \textbf{Alignment with human preference.} 
Preference optimization has become central to aligning large generative models with human expectations. In large language models, reinforcement learning from human feedback (RLHF)~\citep{ouyang2022human} is the dominant framework, relying on a reward model trained from pairwise human preferences~\citep{christiano2017deep, stiennon2020learning}. While effective, RLHF requires careful hyperparameter tuning in both the reward model and reinforcement learning stages. In contrast, direct preference optimization (DPO)~\citep{rafailov2023direct}, its diffusion-specific extension Diffusion-DPO, and several related alternatives~\citep{azizzadenesheli2023mapo, xu2024implicit, lin-etal-2024-limited} offer a simpler approach: directly finetuning the model with a logistic regression objective on preference pairs, thereby eliminating the need for an explicit reward model. However, DPO methods are often less competitive than RLHF~\citep{ouyang2022human}, a limitation also observed in recent adaptations of preference optimization to text-to-image diffusion models~\citep{black2023training, lee2023aligning, blacktraining, fan2023dpok, xu2024imagereward, clarkdirectly, prabhudesai2023aligning, wallace2024diffusion, lialigning, yang2024dense, zhu2025dspo}. Building on this line of work, we propose a Diffusion-DPO variant that reformulates preference alignment as inference-time guidance to improve generalization.

\noindent \textbf{Guidance in diffusion models.} 
Controlling diffusion models can be broadly categorized into \emph{fine-tuning approaches} and \emph{inference-time guidance approaches}. 
Fine-tuning methods adapt model parameters to inject conditioning signals or domain knowledge. 
Representative examples include DreamBooth~\citep{ruiz2023dreambooth}, which personalizes text-to-image models with subject-specific data, and other adapter- or LoRA-style techniques~\citep{hu2021lora, gal2022image}. 
While effective, such methods require additional training and may incur overfitting or catastrophic forgetting when data is limited. 
In contrast, inference-time guidance requires no additional training and modifies the sampling process to incorporate conditioning. 
Classifier guidance~\citep{dhariwal2021diffusion} uses the gradient of an external classifier, but can lead to distributional shifts. 
Classifier-free guidance (CFG)~\citep{ho2022classifierfree} avoids this by training with randomly dropped conditions and linearly combining between unconditional and conditional predictions at inference, and has since become the de facto standard for controllable text-to-image generation. 
Numerous extensions build on this principle, e.g., language-model-based steering~\citep{nichol2021glide}, attention-based semantic guidance~\citep{chefer2023attend}, or plug-and-play conditioning modules~\citep{liu2023plug}. Our work draws direct inspiration from inference-time guidance. 
Instead of conditioning on textual prompts or class labels, we extend the CFG principle to \emph{preference alignment}, treating human preference as a conditioning signal that can be injected at inference to steer generation toward preferred outputs.




\section{Preliminaries}

\subsection{Diffusion models}

Diffusion models are a category of generative models that generate samples by sequentially denoising noisy samples. Specifically, a diffusion model defines a noising (forward) process $q(\mathbf{x}_t \mid \mathbf{x}_{t-1})=\mathcal{N}\!\big(\sqrt{\alpha_t}\,\mathbf{x}_{t-1},\, (1-\alpha_t)\mathbf{I}\big)$
where $\{\beta_t\}_{t=1}^T$ and $\alpha_t = 1-\beta_t$, $\bar\alpha_t=\prod_{s=1}^t \alpha_s$ are the noise schedule, typically set such that $q(x_T | x_0) \approx \mathcal{N}(0, I)$. Sampling from this diffusion model is through the denoising (reverse) process $p_\theta(\mathbf{x}_{t-1}\!\mid\!\mathbf{x}_t)
=\mathcal{N}\!\big(\boldsymbol{\mu}_\theta(\mathbf{x}_t,t),\,\sigma_t^2\mathbf{I}\big)$, with the mean computed using $\boldsymbol{\mu}_\theta(\mathbf{x}_t,t)
=\frac{1}{\sqrt{\alpha_t}}\!\left(\mathbf{x}_t-\frac{\beta_t}{\sqrt{1-\bar\alpha_t}}\,
\epsilon_\theta(\mathbf{x}_t,t)\right)$
and $\sigma_t^2$ set to the posterior variance $\tilde{\beta}_t=\frac{1-\bar\alpha_{t-1}}{1-\bar\alpha_t}\beta_t$.

Training a diffusion model amounts to simply minimize the diffusion loss
\begin{equation}
\mathcal{L}(\theta)
=
\mathop{\mathbb{E}}_{t,\mathbf{x}_0,\epsilon}
\Big[\,
w_t
\big\|
    \epsilon
    -\epsilon_\theta\!\big(\sqrt{\bar\alpha_t}\,\mathbf{x}_0+\sqrt{1-\bar\alpha_t}\,\epsilon,\ t\big)
\big\|_2^2
\Big]
\end{equation}
where $w_t$ is a weighting scalar with one of the common choices being $w(t) = 1$. Such as objective is equivalent to matching the model output $\epsilon_\theta(x,t)$ with the ground-truth score function $\nabla \log p_t(x)$ and we therefore use $\nabla \log \pi(x,t;\theta)$ interchangeably with $\epsilon_\theta(x,t)$.

\subsection{Direct Preference Optimization}

Given a preference dataset $\mathcal{D}$, where $x_+$ and $x_-$ denote the preferred and dispreferred samples conditioned on a prompt $c$, direct preference optimization (DPO)~\citep{rafailov2023dpo} performs logistic regression on the relative log-odds:

{\scriptsize
\begin{align}
    L_{\text{DPO}}
    = - \expt_{(x_+,x_-,c)\sim\mathcal{D}}
    \bigg[
        \log \sigma\!\Big(
            \beta \log \tfrac{\pi_\theta(x_+\mid c)}{\pi_{\text{ref}}(x_+\mid c)}
            -
            \beta \log \tfrac{\pi_\theta(x_-\mid c)}{\pi_{\text{ref}}(x_-\mid c)}
        \Big)
    \bigg],
\end{align}
}

\noindent where $\sigma(\cdot)$ is the sigmoid function and $\beta>0$ is an inverse-temperature. Equivalently, DPO can be viewed as maximum likelihood under an implicit reward model (with normalization constant $Z$)~\citep{rafailov2023direct}:
$
    r(x,c) \;=\; \beta \log \frac{\pi^{*}(x \mid c)}{\pi_{\text{ref}}(x \mid c)} + \log Z,
$
together with the Bradley--Terry (BT) preference model~\citep{bradley1952rank} that, under the optimal policy $\pi^{*}$, yields

{\footnotesize
\begin{align}
    p(x_+ \succ x_- \mid c)
    \;=\;
    \sigma\!\Big(
        \beta \log \tfrac{\pi^{*}(x_+\mid c)}{\pi_{\text{ref}}(x_+\mid c)}
        -
        \beta \log \tfrac{\pi^{*}(x_-\mid c)}{\pi_{\text{ref}}(x_-\mid c)}
    \Big).
\end{align}
}

\paragraph{DPO for diffusion models.}
A na\"{\i}ve application of DPO to diffusion models would treat $x_+$ and $x_-$ as entire reverse-diffusion trajectories from $t{=}T$ to $t{=}0$. This is intractable because computing trajectory-level likelihood ratios requires integrating over all intermediate noise steps. Diffusion-DPO~\citep{wallace2024diffusion} circumvents this by applying Jensen’s inequality to obtain an upper bound on the trajectory loss. Decomposing the joint log-likelihood into a sum of per-step transition log-likelihoods leads to a simple objective:
\begin{align}
L
= - \expt_{\substack{(x_0^+,x_0^-,c)\sim\mathcal{D}\\ t\sim U\{1,\dots,T\}}}
\bigg[&
    \log \sigma\!\Big(
        \beta \log \tfrac{\pi_\theta(x_{+}^{(t-1)} \mid x_{+}^{(t)},c)}{\pi_{\text{ref}}(x_{+}^{(t-1)} \mid x_{+}^{(t)},c)}  \notag  \\
        &-
        \beta \log \tfrac{\pi_\theta(x_{-}^{(t-1)} \mid x_{-}^{(t)},c)}{\pi_{\text{ref}}(x_{-}^{(t-1)} \mid x_{-}^{(t)},c)}
    \Big)
\bigg],
\end{align}

where $\pi_\theta(x_{t-1}\mid x_t,c)$ denotes the one-step reverse-diffusion transition under $\pi_\theta$.

Approximating the transition log-likelihoods with standard diffusion losses then yields:
\begin{align}
L&_{\text{Diff-DPO}}(\theta)
= - \expt_{\substack{(x_0^+,x_0^-,c)\sim\mathcal{D}\\ t\sim U\{1,\dots,T\}}}
\Big[ \log \sigma \!\Big( 
- \beta T \, \omega_t \,
\big(\notag \\
& \|\epsilon^+ - \epsilon_\theta(x_+^{(t)}, t,c)\|^2 
- \|\epsilon^+ - \epsilon_{\text{ref}}(x_+^{(t)}, t,c)\|^2 -\notag \\
&\|\epsilon^- - \epsilon_\theta(x_-^{(t)}, t,c)\|^2
+ \|\epsilon^- - \epsilon_{\text{ref}}(x_-^{(t)}, t,c)\|^2
\big)\Big)\Big].
\end{align}

where $\epsilon^{+}$ and $\epsilon^{-}$ are the Gaussian noises used to form $x_{+}^{(t)}$ and $x_{-}^{(t)}$ from $x_0^{+}$ and $x_0^{-}$, $\epsilon_{\theta}$ and $\epsilon_{\text{ref}}$ are the model and reference noise predictors, $\omega_t$ is the loss weighting factor, and $T$ is the number of diffusion steps.


\subsection{Conditional Generation and Classifier-Free Guidance}

With a conditional diffusion model trained on the dataset $\{(x_i, c_i)\}_{i=1}^N$, the inference process typically adopts classifier-free guidance (CFG)~\citep{ho2022classifierfree} that samples instead with the composed score estimate: $\hat{\epsilon}(\mathbf{x}_t,t,\mathbf{c})=\epsilon_u + w\cdot(\epsilon_c-\epsilon_u),$
where $\epsilon_u=\epsilon_\theta(\mathbf{x}_t,t,\varnothing),
\epsilon_c=\epsilon_\theta(\mathbf{x}_t,t,\mathbf{c})$ are the unconditional and conditional score estimate (respectively), $w$ is a positive guidance weight that is usually greater than $1$, and $\varnothing$ is the null condition. In practice, $\epsilon_u$ is trained by setting the embedding of the condition input to zero. The CFG inference process approximately generates samples from the posterier distribution $p(x) p^w(c | x)$ with $p(x)$ being the prior distribution, or equivalently in log-likelihood, $\log p(x) + w\log p(c|x)$.

\section{Method}

\subsection{Preference-Guided Inference}

Let $\pi_{\mathrm{ref}}$ be a reference policy and $\pi_{\mathrm{DPO}}$ be a DPO-tuned policy. By treating the DPO-tuned policy as $\pi(x | \mathcal{D})$ and the reference policy as $\pi(x | \varnothing)$ as in CFG, we immediately obtain the CFG-style score function for inference, for which we term preference-guided diffusion (PGD):
\begin{align}
    \nabla \log 
    &
    \pi_\text{PGD}(x) = \nabla \log \pi_\text{ref}(x) +
    \notag \\ 
    &
    w\Big(\nabla \log \pi_\text{DPO}(x) - \nabla \log \pi_\text{ref}(x)\Big),
    \label{eqn:pgd}
\end{align}

where the guidance weight $w$ determines the trade-off between our confidence on the reward and other metrics such as prior preservation and sample diversity. Since $\pi_\text{ref}$ can be understood as some prior pretrained on unlabeled datasets, once we have trained $\nabla \log \pi_\text{DPO}(x)$, we are able to virtually align any other base model $\pi_\text{ref}'(x)$ by simply replacing $\pi_\text{ref}(x)$ with it.

\subsection{Contrastive PGD as Dynamically-Reweighted Guidance}

\begin{figure*}[t]
\centering
\begin{minipage}{0.46\linewidth}
  \centering
\vspace{-3pt}   
  \includegraphics[width=\linewidth, height=\linewidth]{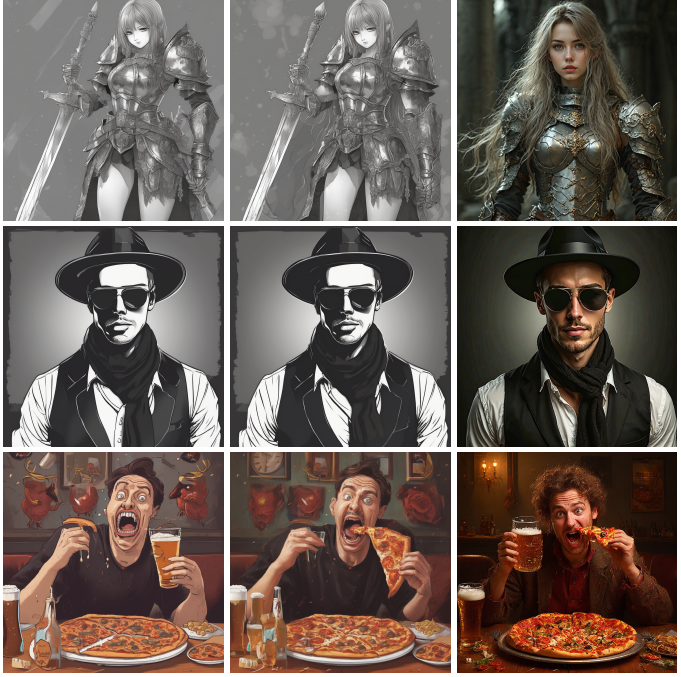} \\
  \makebox[0.32\linewidth][c]{Base}%
  \makebox[0.32\linewidth][c]{DPO}%
  \makebox[0.32\linewidth][c]{PGD}
    \vspace{3pt} 
    \captionsetup{labelfont=footnotesize}
  \vspace{-3mm}
  \caption{
  \footnotesize
  Comparison of base, DPO, and PGD: PGD retains base fidelity while leveraging DPO-learned preferences. 
  }
  \label{fig:pgd}
\end{minipage}\hfill
\begin{minipage}{0.46\linewidth}
  \centering
\vspace{-3pt}   
  \includegraphics[width=\linewidth, height=\linewidth]{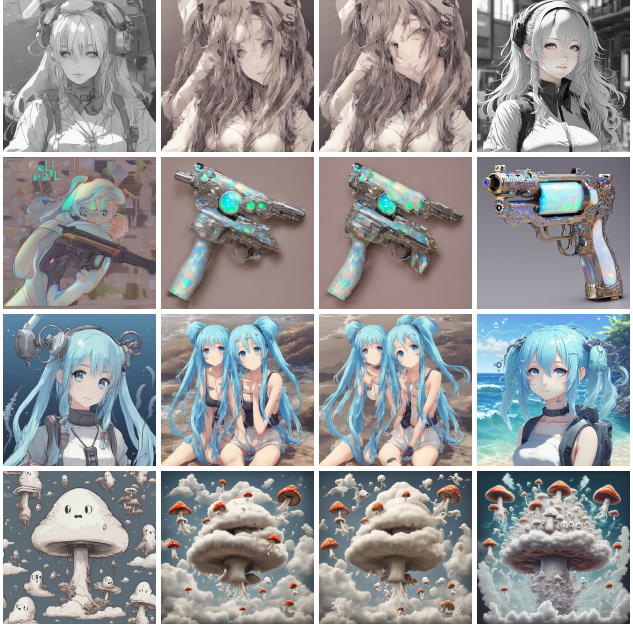}
  \makebox[0.25\linewidth][c]{Base}%
  \makebox[0.25\linewidth][c]{pSFT}%
  \makebox[0.25\linewidth][c]{nSFT}%
  \makebox[0.25\linewidth][c]{cPGD}
  
    \captionsetup{labelfont=footnotesize}
  \vspace{-3mm}
  \caption{
  \footnotesize
  Illustration of cPGD. pSFT and nSFT denote inference with the model finetuned on positive and negative samples, respectively.
  }
  \label{fig:cpgd}
\end{minipage}

\vspace{-2mm}
\end{figure*}

The connection between CFG and diffusion model alignment prompts us to think whether finetuning should also be done in a way similar to conditional diffusion model training, which directly encourage the negative score functions to point towards the data points. However, our preference dataset contains both positive samples and negative ones. These negative samples act as ``repelling'' forces that pushes the negative score function away from them. Inspired by that much of alignment can be turned into an inference-time manner, we propose to postpone this ``repelling'' behavior to inference-time as well. Specifically, we finetune another copy of the base model so that it generates negative samples. Formally speaking, with $\mathcal{D}_+$ representing the set of positive samples and $\mathcal{D}_-$ the set of negative ones, we independently finetune two models (with parameters $\theta_+$ and $\theta_-$, respectively) with diffusion losses ($x_t(x_0, \epsilon) = \sqrt{\bar\alpha_t}\,x_0+\sqrt{1-\bar\alpha_t} \epsilon$):
\begin{align}
    L_\text{pos}(\theta_+) = 
    & 
    \mathop{\mathbb{E}}_{\epsilon, t, x_0 \sim \mathcal{D}_+} 
    \Big\|
        \epsilon
        -\epsilon_{\theta_+} \big(
            x_t(x_0, \epsilon), t
        \big)
    \Big\|^2
\\
    L_\text{neg}(\theta_-) = 
    & 
    \mathop{\mathbb{E}}_{\epsilon, t, x_0 \sim \mathcal{D}_-} 
    \Big\|
        \epsilon
        -\epsilon_{\theta_-} \big(
            x_t(x_0, \epsilon), t
        \big)
    \Big\|^2
\end{align}
Intuitively, the difference between two models characterizes the implicit reward model. Therefore we may write the residual parameterization $\nabla \log \pi_\text{finetuned}(x,t) = \nabla \log \pi(x,t; \theta_+) - \nabla \log \pi(x,t; \theta_-) + \nabla \log \pi_\text{ref}(x,t)$. It follows that the resulting PGD formulation, to which we refer with contrastive PGD (cPGD), is
\begin{align}
    \nabla \log &\pi_\text{PGD}(x, t) = \nabla \log \pi_\text{ref}(x, t) + \notag \\ &w\Big(\nabla \log \pi(x, t; \theta_+) - \nabla \log \pi(x, t; \theta_-)\Big).
    \label{eqn:cpgd}
\end{align}
\noindent \textbf{Alternative perspective of cPGD.} While it may seem a bit arbitrary to replace the DPO loss on the preference dataset with two diffusion losses on positive-only and negative-only datasets respectively, cPGD essentially performs dynamic reweighting of DPO loss gradients. For simplicity, let's consider the general DPO case (without Diffusion-DPO approximations). If we plug the residual parametrization of the finetuned model into the DPO loss gradient, we observe (with $\theta = (\theta_+, \theta_-)$):
\begin{align}
    \nabla_\theta L_\text{DPO} 
    &= 
    - \mathop{\mathbb{E}}_{(x_+, x_-) \sim \mathcal{D}}
    \Big[
        \beta\sigma\Big(
            \log \pi(x;\theta_-) - \log \pi(x;\theta_+)
        \Big) \notag \\ 
    \cdot \Big( 
        &\nabla_{\theta_+} \log \pi(x;\theta_+) - \nabla_{\theta_-} \log \pi(x;\theta_-)
    \Big)
    \Big].
\end{align}
Suppose for each sample pair $(x_+, x_-)$ we dynamically reweight the loss function by {\footnotesize \( 1 / \Big[\beta\sigma\Big(\log \pi(x;\theta_-) - \log \pi(x;\theta_+)\Big)\Big] \) } and obtain
\begin{align}
    & \nabla_\theta L_\text{reweight}
    \notag \\
= &
    - \mathop{\mathbb{E}}_{(x_+, x_-)\sim \mathcal{D}}
    \Big[
        \nabla_{\theta_+} \log \pi(x;\theta_+) 
        - \nabla_{\theta_-} \log \pi(x;\theta_-)
    \Big]\notag
\\
= &
    \mathop{\mathbb{E}}_{x_-\sim \mathcal{D}_-}
    \Big[
        \nabla_{\theta_-} \log \pi(x;\theta_-)
    \Big] 
    - \mathop{\mathbb{E}}_{x_+\sim \mathcal{D}_+}
    \Big[
        \nabla_{\theta_+} \log \pi(x;\theta_+)
    \Big]
\end{align}
which is exactly the gradient of the cPGD training objectives once we take into consideration the fact that $\log \pi$ is parameterized by a diffusion model. \citet{wu2025dft} show that such reweighting can be seen as an interpolation between supervised finetuning gradients and
vanilla policy gradients, and that it can be helpful to alleviate the overfitting issue due to the small scale of finetuning datasets.

\begin{figure*}
    \centering

    \includegraphics[width=0.8\linewidth]{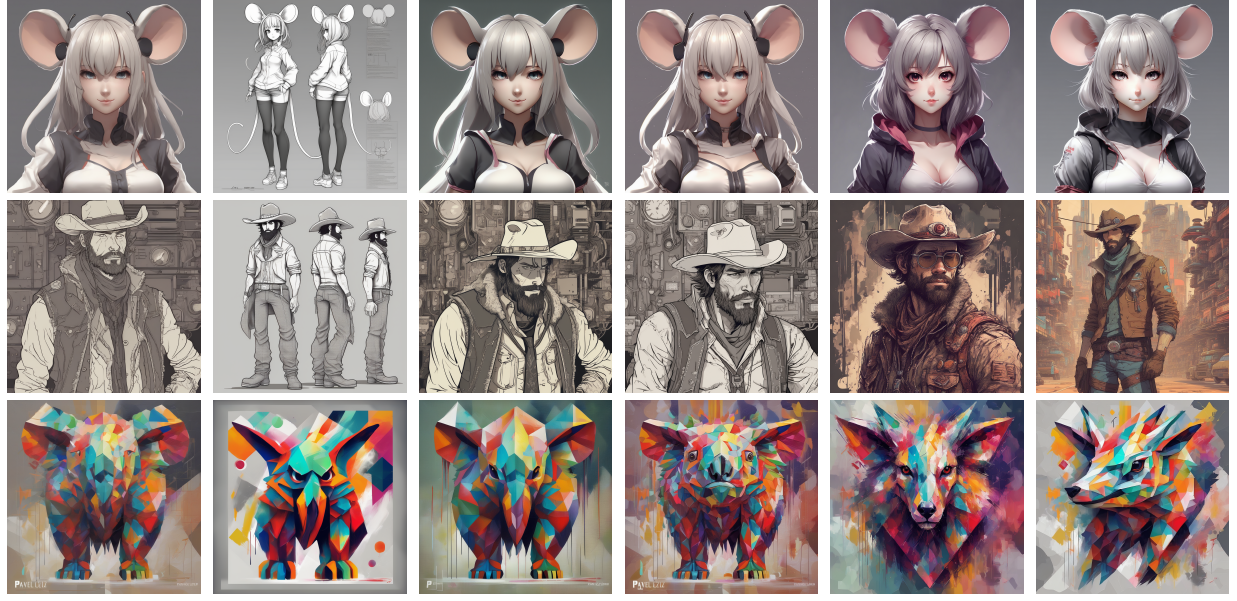}\\

\makebox[0.13\linewidth][c]{SDXL}%
\makebox[0.13\linewidth][c]{DPO}%
\makebox[0.13\linewidth][c]{MaPO}%
\makebox[0.13\linewidth][c]{NPO}
\makebox[0.13\linewidth][c]{PGD}
\makebox[0.13\linewidth][c]{cPGD}
\vspace{-1mm}
\captionsetup{labelfont=footnotesize}
    \caption{\footnotesize Comparison of preference-optimization methods on SDXL. Columns show outputs from the base model (SDXL), DPO, MaPO, NPO, PGD, and cPGD. PGD and cPGD achieves the highest rewards and is the most effective in aligning with human preference implied in the Pick-a-Pic v2 dataset.}
    \label{fig:m1_figure}
\vspace{-4mm}
\end{figure*}

\begin{table*}[ht]
\centering

\captionsetup{labelfont=footnotesize}
\resizebox{\linewidth}{!}{%
\begin{tabular}{cc|cccccc|cccccc|c}
\toprule
 & & \multicolumn{6}{c|}{Pick-a-Pic v2 test (424 prompts)} & \multicolumn{6}{c|}{Parti-Prompts (1632 prompts)} & \\
\multirow{-2}{*}{Base Model} &\multirow{-2}{*}{\makecell{Inference \\ Strategy}}  & \textbf{ PS} $\uparrow$ & \textbf{ HPSv2} $\uparrow$ & \textbf{ HPSv3} $\uparrow$ & \textbf{ Aes} $\uparrow$ & \textbf{ CLIP} $\uparrow$ & \textbf{ IR} $\uparrow$  & \textbf{ PS} $\uparrow$ & \textbf{ HPSv2} $\uparrow$ & \textbf{ HPSv3} $\uparrow$ & \textbf{ Aes} $\uparrow$ & \textbf{ CLIP} $\uparrow$ & \textbf{ IR} $\uparrow$  & \multirow{-2}{*}{\textbf{Avg.} $\uparrow$} \\
\midrule
                         & --                 & 50.0           & 50.0           & 50.0           & 50.0           & 50.0           & 50.0           & 50.0           & 50.0           & 50.0           & 50.0           & 50.0           & 50.0           & 50.0 \\
                          & NPO                & 58.7           & 59.2           & 69.1           & \textbf{ 52.1 } & 37.5           & 53.5           & 55.0           & 56.4           & 62.1           & 55.1           & 39.0           & 51.5           & 54.1 \\
                         & PGD              & {\ul\ 78.8 }    & {\ul\ 79.0 }    & {\ul\ 73.6 }    & {\ul\ 51.9 }    & {\ul\ 63.7 }    & {\ul\ 69.3 }    & \textbf{ 78.7 } & {\ul\ 78.4 }    & \textbf{ 78.2 } & \textbf{ 68.8 } & {\ul\ 54.0 }    & {\ul\ 69.1 }    & {\ul70.3} \\
\multirow{-4}{*}{\cellcolor{white}{SDXL}}   & cPGD              & \textbf{ 80.0 } & \textbf{ 80.2 } & \textbf{ 77.1 } & 50.9           & \textbf{ 64.9 } & \textbf{ 69.8 } & {\ul\ 75.8 }    & \textbf{ 80.3 } & {\ul\ 77.8 }    & {\ul\ 57.2 }    & \textbf{ 62.0 } & \textbf{ 74.0 } & \textbf{70.8} \\

\midrule
                            & --                 & 71.7           & 77.6           & 67.9           & 53.3           & {\ul\ 61.6 }    & {\ul\ 65.8 }    & 64.0           & 70.3           & 64.0           & 57.5           & 58.5           & 69.7           & 65.2 \\
                             & NPO                & 76.9           & {\ul\ 81.8 }    & 81.4           & 53.8           & 57.8           & 70.8           & 70.6           & 78.4           & 77.5           & {\ul\ 63.4 }    & 56.3           & 70.8           & 70.0 \\
                     & PGD              & \textbf{ 83.3 } & \textbf{ 85.4 } & \textbf{ 85.6 } & {\ul\ 59.7 }    & \textbf{ 62.3 } & \textbf{ 73.6 } & \textbf{ 80.8 } & \textbf{ 83.9 } & \textbf{ 81.7 } & \textbf{ 67.6 } & {\ul\ 57.5 }    & \textbf{ 76.1 } & \textbf{74.8} \\
\multirow{-4}{*}{DPO-SDXL}     & cPGD              & {\ul\ 80.9 }    & 77.6           & {\ul\ 84.7 }    & \textbf{ 63.9 } & 58.7           & 64.9           & {\ul\ 73.9 }    & {\ul\ 79.2 }    & {\ul\ 72.8 }    & 59.4           & \textbf{ 64.1 } & {\ul\ 74.4 }    & {\ul71.2} \\

\midrule
                            & --                 & 55.9           & 65.3           & 61.8           & 68.2           & 50.2           & 68.2           & 52.0           & 64.4           & 58.5           & {\ul\ 72.4 }    & 48.2           & 65.0           & 60.8 \\
                 & PGD              & \textbf{ 80.4 } & \textbf{ 81.6 } & \textbf{ 81.6 } & \textbf{ 75.7 } & {\ul\ 51.4 }    & \textbf{ 72.2 } & \textbf{ 78.9 } & {\ul\ 77.8 }    & \textbf{ 79.6 } & \textbf{ 77.8 } & {\ul\ 53.6 }    & {\ul\ 72.7 }    & \textbf{73.6} \\
\multirow{-3}{*}{MaPO-SDXL} & cPGD              & {\ul\ 77.4 }    & {\ul\ 78.8 }    & {\ul\ 72.9 }    & {\ul\ 69.1 }    & \textbf{ 59.4 } & {\ul\ 72.4 }    & {\ul\ 72.5 }    & \textbf{ 81.1 } & {\ul\ 76.9 }    & 70.1           & \textbf{ 58.2 } & \textbf{ 74.4 } & {\ul71.9} \\
\midrule
                            & --                 & 89.4           & 83.0           & 96.0           & {\ul\ 81.8 }    & 33.3           & 78.8           & 87.8           & 85.5           & 92.2           & {\ul\ 88.1 }    & 31.7           & 74.9           & 76.9 \\
                 & PGD              & {\ul92.2}    & {\ul\ 86.1 }    & {\ul\ 96.5 }    & \textbf{ 82.1 } & {\ul\ 42.5 }    & {\ul\ 81.4 }    & {\ul\ 91.4 }    & {\ul\ 87.7 }    & \textbf{ 93.9 } & \textbf{ 88.4 } & {\ul\ 48.0 }    & {\ul\ 77.3 }    & {\ul\ 80.6 } \\
\multirow{-3}{*}{SPO-SDXL}   & cPGD              & \textbf{ 92.9 } & \textbf{ 88.4 } & \textbf{ 96.7 } & 78.8           & \textbf{ 53.8 } & \textbf{ 84.4 } & \textbf{ 92.0 } & \textbf{ 90.3 } & {\ul\ 93.9 }    & 83.6           & \textbf{ 50.4 } & \textbf{81.3} & \textbf{82.2} \\
\bottomrule
\end{tabular}
}
\vspace{-2mm}
\caption{\footnotesize Win rates of preference optimization methods against the SDXL model on the Pick-a-Pic v2 test set and the Parti-Prompts benchmark. Model checkpoints for other methods are provided by their respective authors. The 1st-best results are \textbf{bolded} and the 2nd-best results are \uline{underlined}.}
\label{tab:combined-sdxl-winrate}
\vspace{-1mm}
\end{table*}

\begin{table*}[ht]
\centering
\vspace{-1mm}
\resizebox{\linewidth}{!}{%
\begin{tabular}{cc|cccccc|cccccc|c}
\toprule
 & & \multicolumn{6}{c|}{Pick-a-Pic v2 test (424 prompts)} & \multicolumn{6}{c|}{Parti-Prompts (1632 prompts)} & \\
\multirow{-2}{*}{Base Model} &\multirow{-2}{*}{\makecell{Inference \\ Strategy}} & \textbf{ PS} $\uparrow$ & \textbf{ HPSv2} $\uparrow$ & \textbf{ HPSv3} $\uparrow$ & \textbf{ Aes} $\uparrow$ & \textbf{ CLIP} $\uparrow$ & \textbf{ IR} $\uparrow$  & \textbf{ PS} $\uparrow$ & \textbf{ HPSv2} $\uparrow$ & \textbf{ HPSv3} $\uparrow$ & \textbf{ Aes} $\uparrow$ & \textbf{ CLIP} $\uparrow$ & \textbf{ IR} $\uparrow$ & \multirow{-2}{*}{\textbf{Avg.} $\uparrow$} \\
\midrule
                             & --                 & 50.0           & 50.0           & 50.0           & 50.0           & 50.0           & 50.0           & 50.0           & 50.0           & 50.0           & 50.0           & 50.0           & 50.0           & 50.0 \\
{}                & PGD                & \textbf{ 78.3 } & 71.2           & 67.9           & 62.3           & 58.5           & 63.7           & \textbf{ 68.0 } & 65.0           & 59.6           & 58.1           & 55.0           & 56.9           & {\ul63.7} \\
\multirow{-3}{*}{SD1.5}      & cPGD               & 76.9           & \textbf{ 71.7 } & \textbf{ 71.9 } & \textbf{ 63.2 } & \textbf{ 59.9 } & \textbf{ 72.2 } & 66.4           & \textbf{ 76.9 } & \textbf{ 68.9 } & \textbf{ 68.1 } & \textbf{ 58.4 } & \textbf{ 71.0 } & \textbf{68.8} \\
\midrule
                             & --                 & 76.4           & 67.7           & 66.3           & 65.1           & 55.9           & 60.6           & 67.3           & 64.8           & 64.5           & 62.2           & 53.6           & 61.0           & 63.8 \\
                 & PGD                & \textbf{ 79.2 } & {\ul\ 70.3 }    & {\ul\ 66.3 }    & {\ul\ 66.3 }    & {\ul\ 59.4 }    & {\ul\ 63.2 }    & \textbf{ 74.2 } & {\ul\ 67.4 }    & {\ul\ 65.0 }    & {\ul\ 63.1 }    & {\ul\ 55.5 }    & {\ul\ 62.9 }    & {\ul66.1} \\
\multirow{-3}{*}{DPO-SD1.5}  & cPGD               & {\ul\ 79.0 }    & \textbf{ 81.8 } & \textbf{ 75.5 } & \textbf{ 71.7 } & \textbf{ 62.3 } & \textbf{ 76.2 } & {\ul\ 73.8 }    & \textbf{ 74.1 } & \textbf{ 69.6 } & \textbf{ 69.0 } & \textbf{ 59.0 } & \textbf{ 67.3 } & \textbf{71.6} \\
\midrule
                             & --                 & 72.6           & 78.1           & 76.2           & 68.6           & 58.5           & \textbf{ 75.0 } & 66.6           & {\ul\ 78.3 }    & {\ul\ 71.9 }    & {\ul\ 68.8 }    & 53.3           & 71.3           & 69.9 \\
              & PGD                & \textbf{ 81.6 } & \textbf{ 83.3 } & \textbf{ 80.2 } & {\ul\ 70.3 }    & \textbf{ 61.1 } & {\ul\ 77.4 }    & \textbf{ 72.1 } & \textbf{ 80.3 } & \textbf{ 72.8 } & \textbf{ 72.4 } & {\ul\ 55.1 }    & \textbf{ 73.8 } & \textbf{73.4} \\
\multirow{-3}{*}{KTO-SD1.5}  & cPGD               & {\ul\ 76.7 }    & {\ul\ 80.2 }    & {\ul\ 75.9 }    & \textbf{ 70.5 } & {\ul\ 60.4 }    & 74.3           & {\ul\ 66.2 }    & 77.1           & 69.5           & 68.4           & \textbf{ 55.9 } & {\ul\ 72.2 }    & {\ul70.6} \\
\midrule
                             & --                 & 71.2           & 63.0           & 64.9           & 68.2           & 38.7           & 61.1           & 68.6           & 61.2           & 64.2           & 71.9           & 37.7           & 61.6           & 61.0 \\

               & PGD                & {\ul\ 79.7 }    & {\ul\ 70.3 }    & {\ul\ 69.6 }    & {\ul\ 69.8 }    & {\ul\ 44.3 }    & {\ul\ 67.9 }    & {\ul\ 74.2 }    & {\ul\ 66.5 }    & {\ul\ 66.2 }    & {\ul\ 72.2 }    & {\ul\ 46.7 }    & {\ul\ 67.0 }    & {\ul66.2} \\

\multirow{-3}{*}{SPO-SD1.5}  & cPGD               & \textbf{ 82.3 } & \textbf{ 81.8 } & \textbf{ 75.5 } & \textbf{ 71.2 } & \textbf{ 60.6 } & \textbf{ 76.2 } & \textbf{ 74.8 } & \textbf{ 71.9 } & \textbf{ 71.3 } & \textbf{ 73.9 } & \textbf{ 47.5 } & \textbf{ 72.9 } & \textbf{71.7} \\
\bottomrule
\end{tabular}
}
\vspace{-2mm}
\captionsetup{labelfont=footnotesize}
\caption{\footnotesize Win rates of preference optimization methods against the SD1.5 model on the Pick-a-Pic v2 test set and the Parti-Prompts benchmark. Model checkpoints for other methods are provided by their respective authors. The 1st-best results are \textbf{bolded} and the 2nd-best results are \uline{underlined}.}
\label{tab:combined-SD1.5-winrate}
\vspace{-3.5mm}
\end{table*}

\begin{figure*}[t]
\centering
\resizebox{0.9\linewidth}{!}{
\begin{tabular}{c c c}

 \multicolumn{3}{c}{\includegraphics[width=0.9\linewidth]{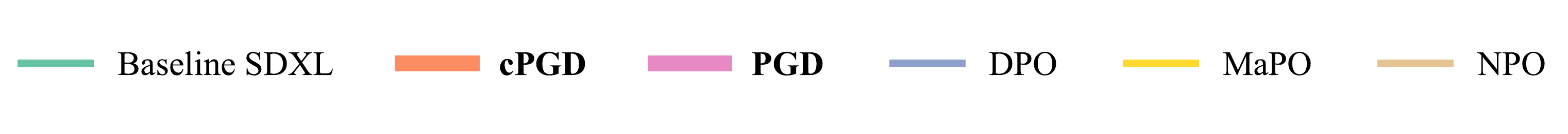}} \\[-3ex]

  \includegraphics[width=0.33\linewidth]{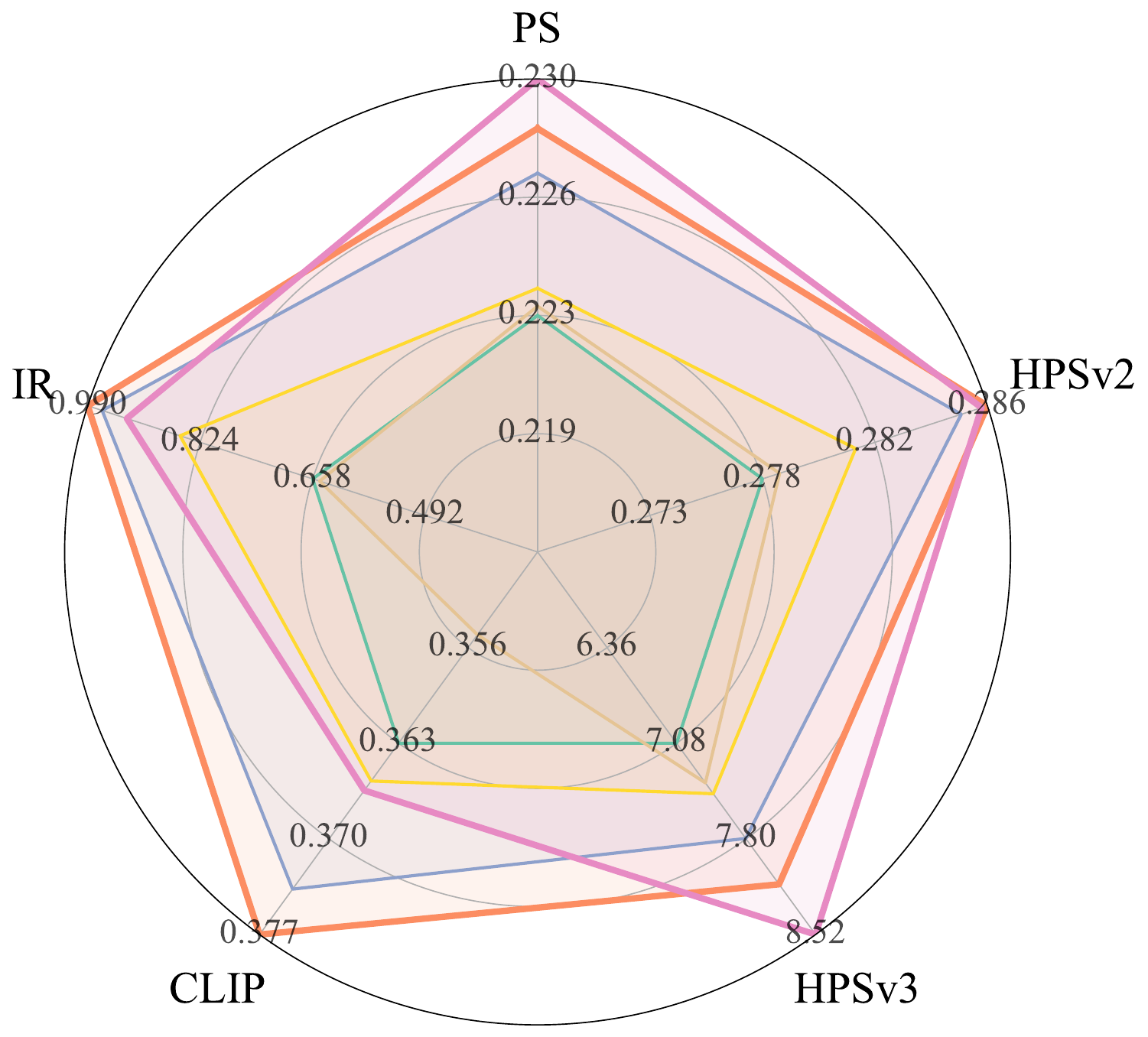} &
  \includegraphics[width=0.33\linewidth]{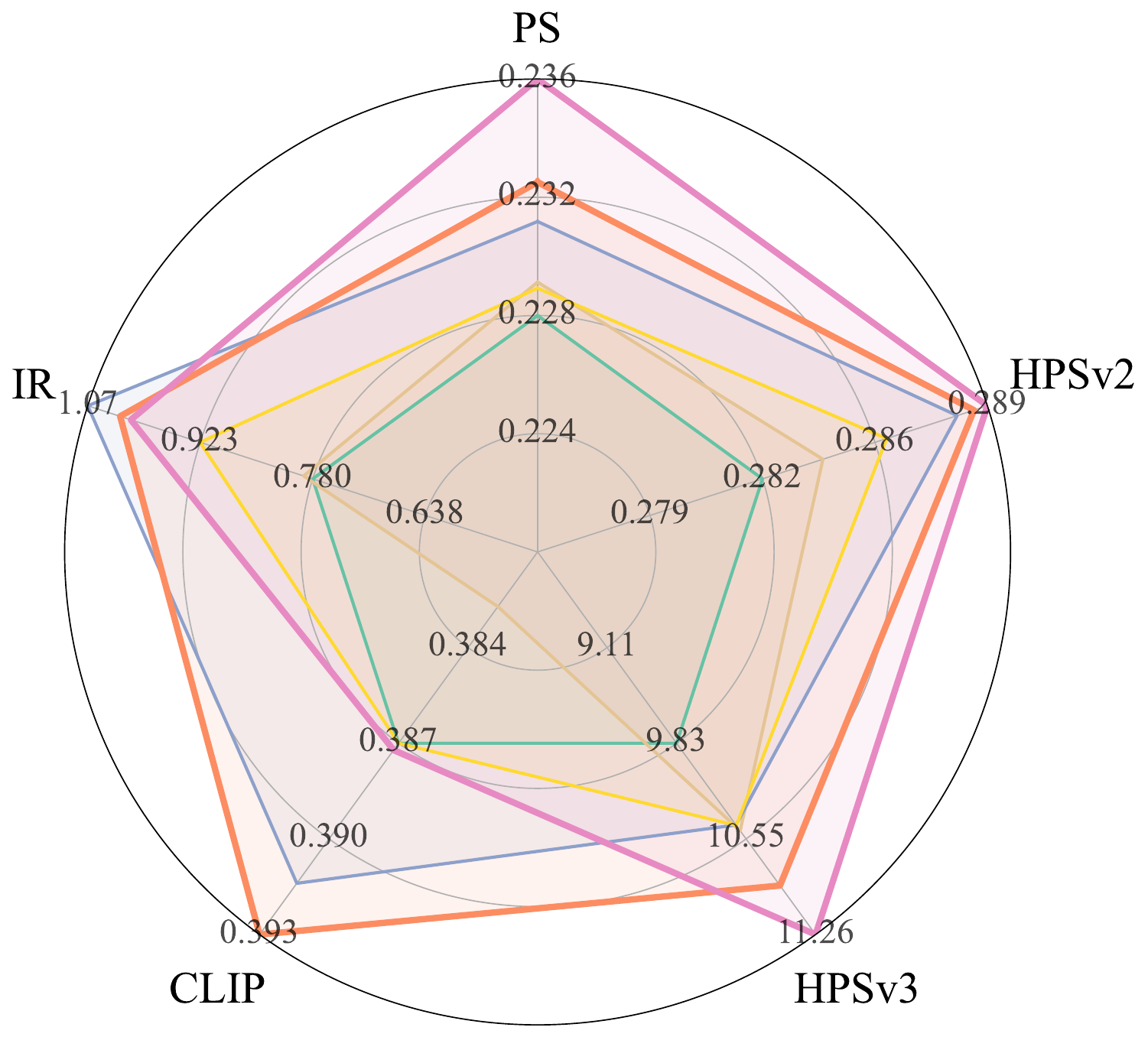} &
  \includegraphics[width=0.33\linewidth]{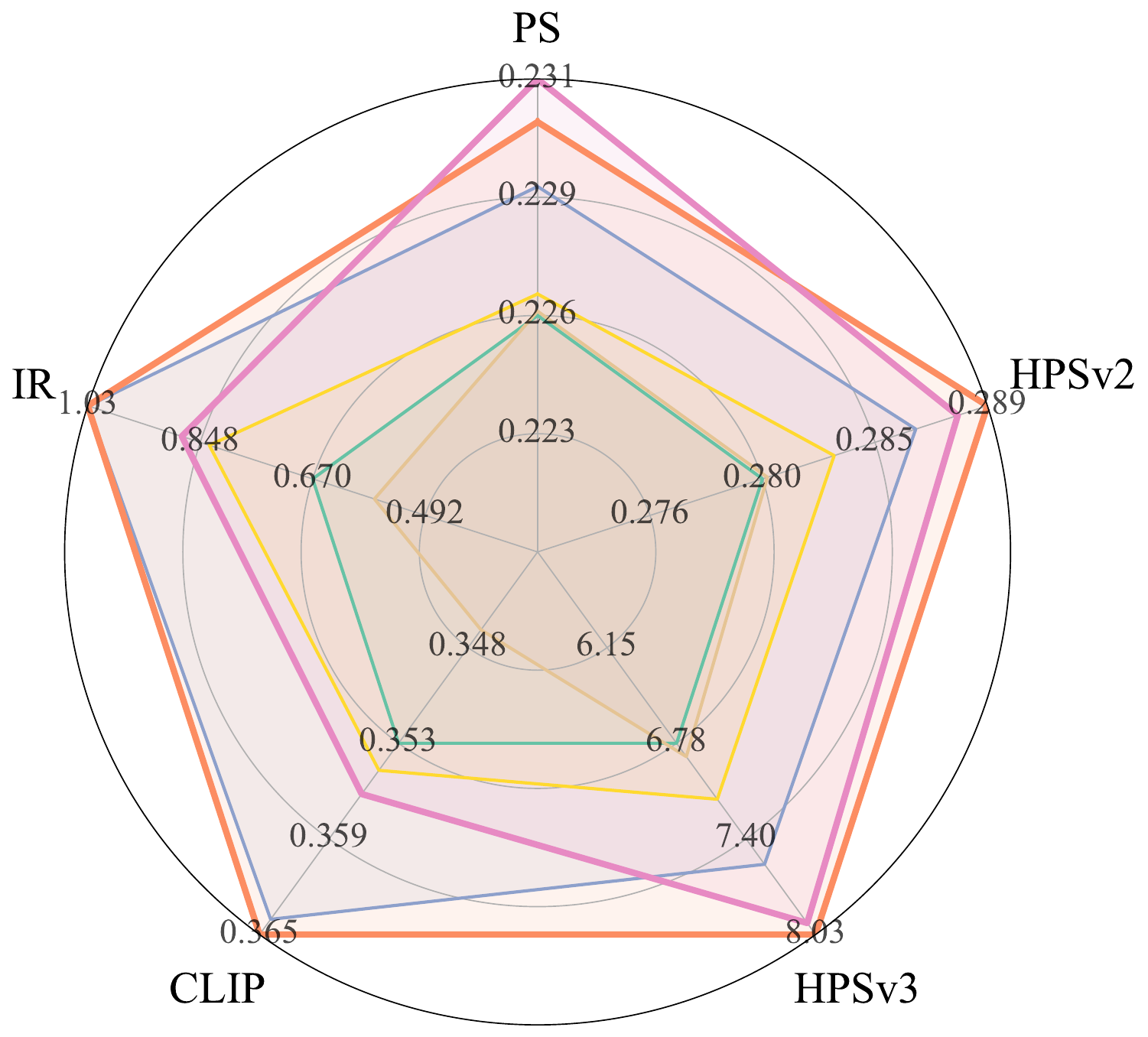} \\[1ex]

  Pic-a-Pic v2 test & HPDv2 & Parti-Prompts \\
\end{tabular}
}
\captionsetup{labelfont=footnotesize}
\caption{\footnotesize Overall comparison on SDXL. 
Radar axes report mean scores (higher is better): PickScore (PS), HPSv2, HPSv3, CLIP, and ImageReward (IR). Polygons closer to the outer rim indicate better aggregate performance across metrics.}
\label{fig:general_results}
\vspace{-3mm}
\end{figure*}

\section{Experiments}

\subsection{Experimental Setup}
\label{sec:exp_seting}

\noindent\textbf{Training datasets.} We consider consider two datasets: 1) Pick-a-Pic v2~\citep{kirstain2023pick}, which consists of approximately 900,000 image preference pairs derived from 58,000 unique prompts, and 2) HPDv3~\citep{hpdv3}, which comprises 1.08M text-image pairs and 1.17M annotated pairwise data. Our main experiments are done on Pick-a-Pic v2, while for ablation, we create a high-image-quality subset of HPDv3 besides the full dataset of HPDv3.

\noindent\textbf{Test prompts.} We consider the following prompt datasets for testing: the test split of Pick-a-Pic v2 (424 prompts), the HPDv2 test set~\citep{hpdv2} (400 prompts), and the Parti-Prompts benchmark~\citep{yu2022scaling} (1,632 prompts).

\noindent\textbf{Baselines.}
We benchmark our approaches against the following baselines:
(i) SFT-Pref, a supervised fine-tuning baseline using only the preferred images;
(ii) Diffusion-DPO~\citep{wallace2024diffusion}, an adaptation of the DPO method to diffusion models;
(iii) Diffusion-KTO~\citep{lialigning}, a variant that incorporates a Kullback–Leibler trade-off to Diffusion-DPO for unlocking the potential of leveraging readily available per-image binary signals;
(iv) MaPO~\citep{hong2024marginaware}, which refines preference optimization with margin-based pairwise consistency;
(v) Diffusion-NPO~\citep{diffusionnpo}, which explicitly models negative preferences to strengthen classifier-free guidance. Additionally, we consider SPO~\citep{liang2024step}, a hybrid method that trains auxiliary reward models in an online fashion during DPO finetuning; due to its online nature, we exclude SPO from the direct comparison between offline DPO variants.

\noindent\textbf{Reward models.}
We evaluate generated images with these reward models: PickScore (PS)~\citep{kirstain2023pick}, HPSv2~\citep{hpdv2}, HPSv3~\citep{hpdv3}, ImageReward (IR)~\citep{xu2024imagereward}, CLIP Score~\citep{radford2021learning}, and Aesthetics Score (Aes)~\citep{laionaes}. 

\noindent\textbf{Metrics.} Besides the absolute reward values, we compute win rates for different methods, which is the percentage of instances where the finetuned model outperforms the base model. Since win rates are considerably more robust than absolute reward values~\citep{hpdv2, kirstain2023pick} , we use win rate as our primary metric. Plus, we compute FID score and diversity score to measure the extent of prior preservation and sample diversity, respectively. Sample diversity is computed by measuring the average pairwise distances between CLIP embeddings~\cite{radford2021learning} of generated samples (see Appendix~\ref{sec:EvaluationMetrics}).

\noindent\textbf{Implement Details.}
We experiment with two base models: Stable Diffusion v1.5 (SD1.5)~\citep{rombach2022high} and Stable Diffusion XL base (SDXL)~\citep{sdxl}. An effective batch size of 2048 image pairs is used for all experiments. Following common practices, we set for each model a base learning rate and scale it linearly with the batch size. For SD1.5, we use AdamW optimizer with a base learning rate of 3e-8; for the larger SDXL model, we employ Adafactor optimizer with a base learning rate of 5e-9. Following \citet{wallace2024diffusion}, $\beta$ is set to 3000 and 5000 for SD1.5 and SDXL, respectively. It is worth noting, however, that our effective learning rate is smaller than that of \citet{wallace2024diffusion}. In their setting, the effective learning rate is 2.048e-5, whereas ours is only 9.6e-7, an order of magnitude smaller. For PGD, we use the finetuned model with training 2000 steps and for cPGD, we use models trained with 500 steps.

\subsection{Results}

\begin{figure}[ht]
\centering
\begin{minipage}{0.484\linewidth}
  \centering
  \includegraphics[width=\linewidth]{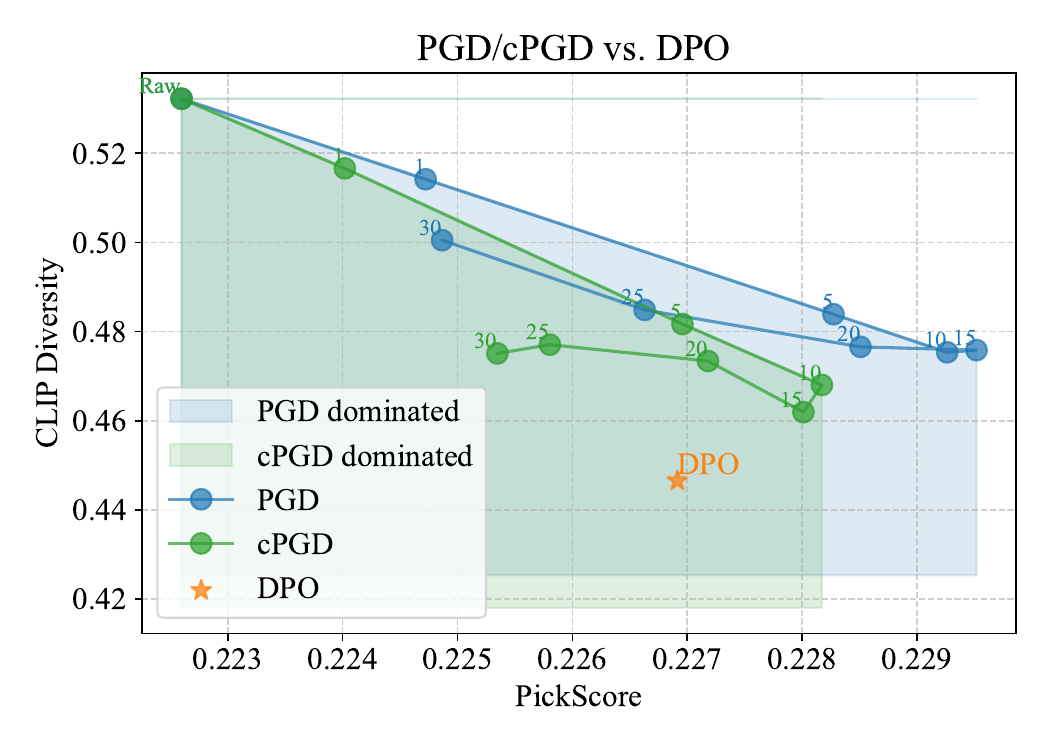}
  \caption{Reward–Diversity}
  \label{fig:diversity}
\end{minipage}%
\hspace{0.02\linewidth} 
\begin{minipage}{0.484\linewidth}
  \centering
  \includegraphics[width=\linewidth]{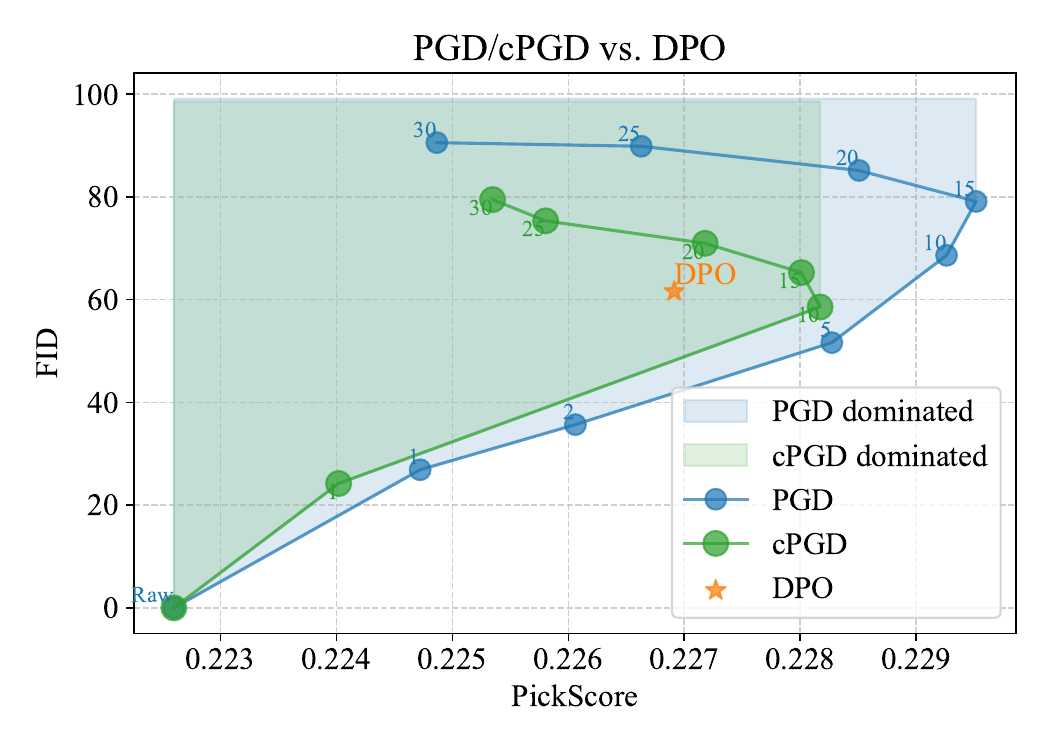}
  \caption{Reward–Fidelity}
  \label{fig:fid}
\end{minipage}

\vspace{-6mm}
\end{figure}

\noindent \textbf{General results.} As shown in Fig.~\ref{fig:general_results}, Table~\ref{tab:combined-sdxl-winrate} and Table~\ref{tab:combined-SD1.5-winrate}, we find that our proposed method PGD and cPGD generally outperform the baselines in achieving higher absolute reward values and win rates for different test prompt sets and different base models. While our methods generally achieve lower Aes scores, the behavior is less indicative because our training objective is to align with the human preference implied by text-image paired datasets, while Aes is an unconditional reward model that does not take text-image alignment into consideration.

\noindent \textbf{Diversity and prior preservation.} We further demonstrate the tradeoffs between reward, FID (measuring prior preservation) and diversity scores in Fig.~\ref{fig:fid} and Fig.~\ref{fig:diversity}. The blue regions are the combinations that are strictly dominated by the performance of our methods, the boundary of which is formed by the performance resulted from different choices of guidance weights.

\noindent \textbf{PGD vs. cPGD.} We find that cPGD is generally better on SD1.5 but comparable on SDXL. We hypothesize that such behavior is due to the distribution shift between the image distributions of the preference datasets and that of the base model. As the images in the preference datasets we used are generally better than those from SD1.5, the dynamic reweighting mechanism used in cPGD helps generalization.

\noindent \textbf{Transfer to other base models.} Inspired by the plug-and-play nature of our approach, we experiment with aligning base models that are finetuned with alternative DPO variants on the same preference datasets using our PGD/cPGD-finetuned modules (\textit{e.g.}, the second mega-row ``DPO-SDXL'' in Table~\ref{tab:combined-sdxl-winrate} demonstrate the performance when using DPO-tuned SDXL as the inference-time base model). We find that there is nearly consistent improvement compared to any original base model, which is made easy with the CFG-style inference rule in our PGD method.

\begin{figure*}[t]
\centering

\resizebox{0.99\textwidth}{!}{%
\includegraphics[]{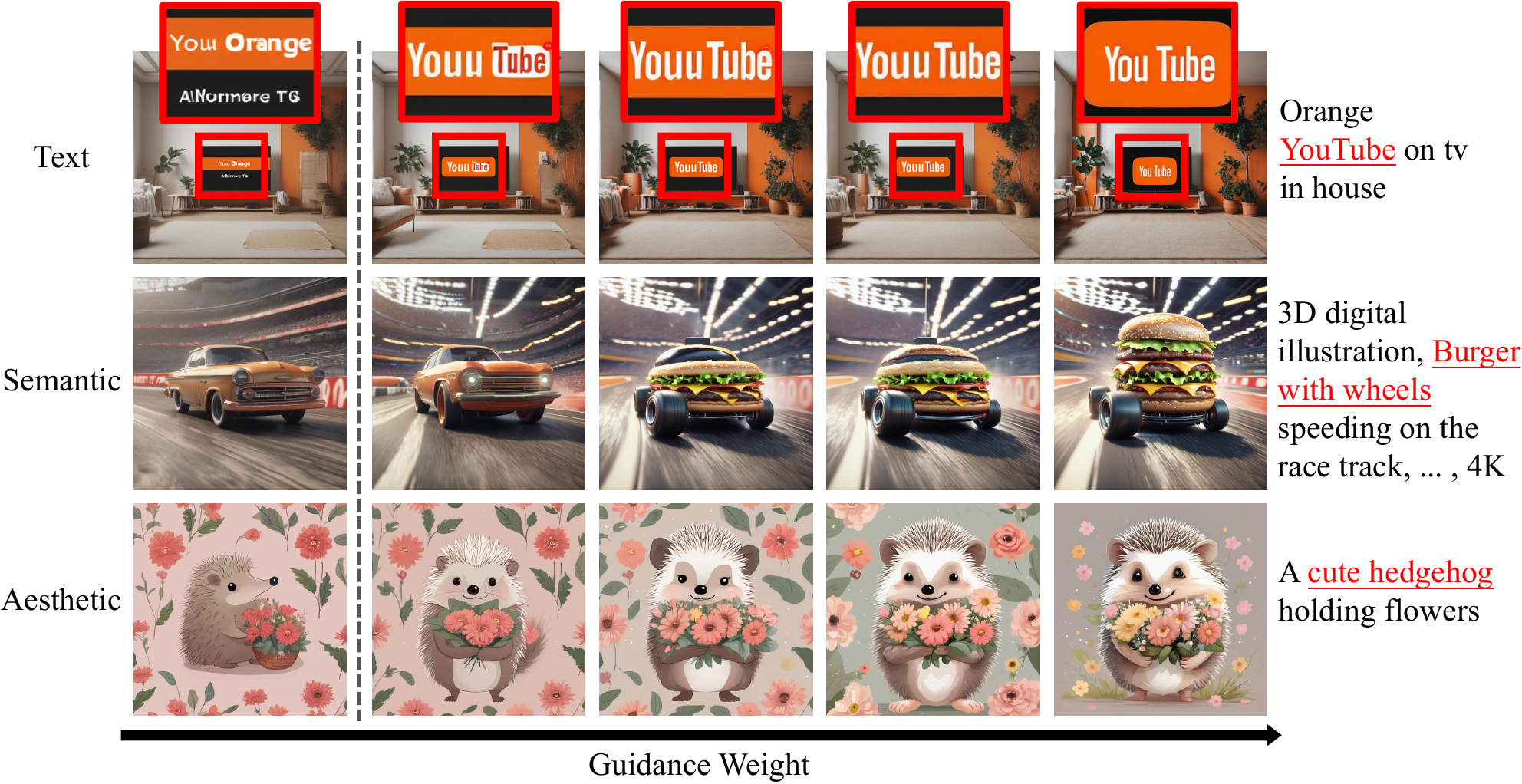} 
}
\vspace{-1mm}

\captionsetup{labelfont=footnotesize}
\caption{\footnotesize Qualitative effect of increasing guidance weight $w$ (left $\rightarrow$ right). 
Rows show text fidelity, semantic binding, and aesthetic style. 
Stronger $w$ improves alignment and legibility up to a mid range, after which overshooting/rigidity appears.}
\label{fig:weight_results}
\vspace{-3mm}
\end{figure*}

\begin{figure*}[t]
\vspace{1mm}
\centering
\vspace{-1pt}   

\resizebox{\linewidth}{!}{
\begin{tabular}{@{}c c c}

  \includegraphics[width=0.33\linewidth]{ 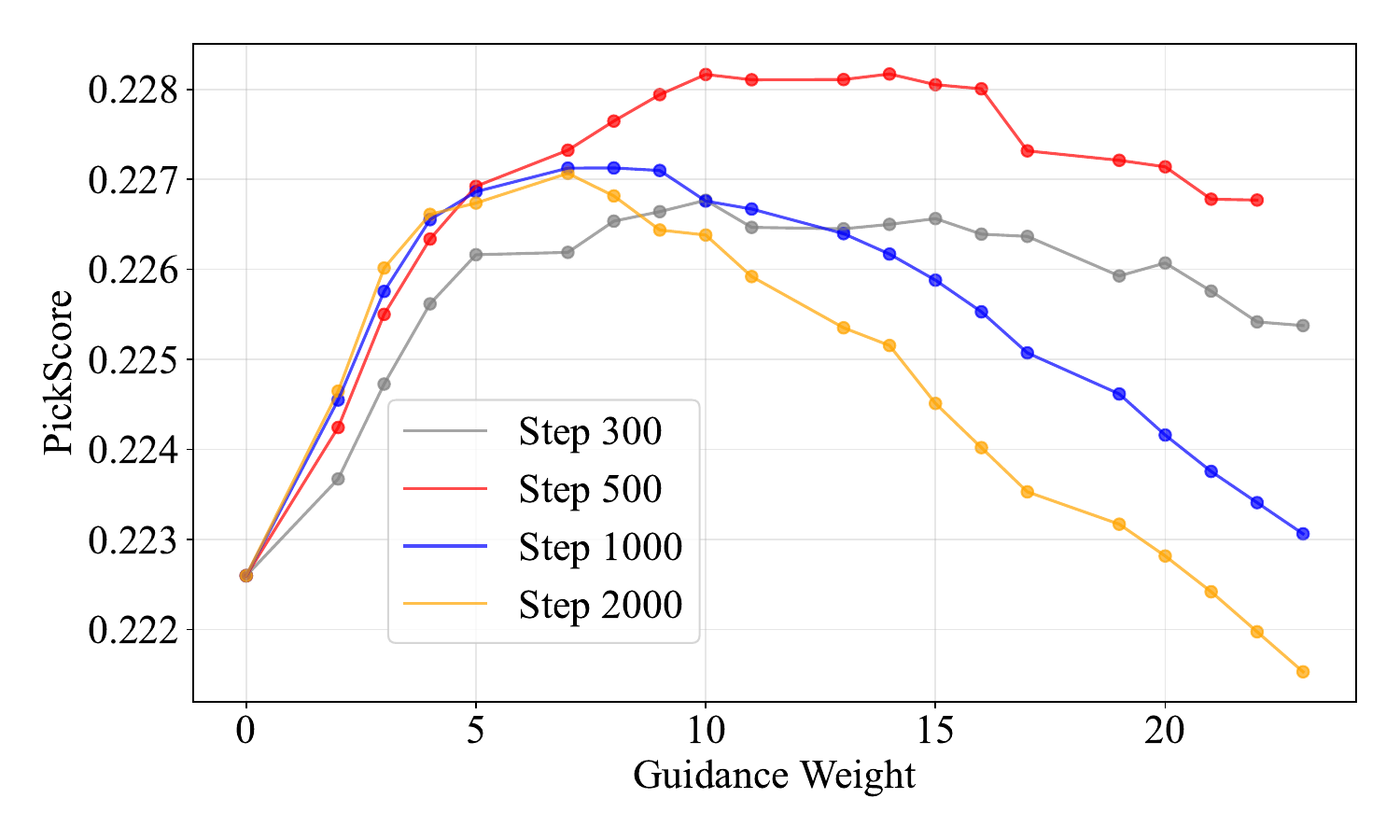} &
  \includegraphics[width=0.33\linewidth]{ 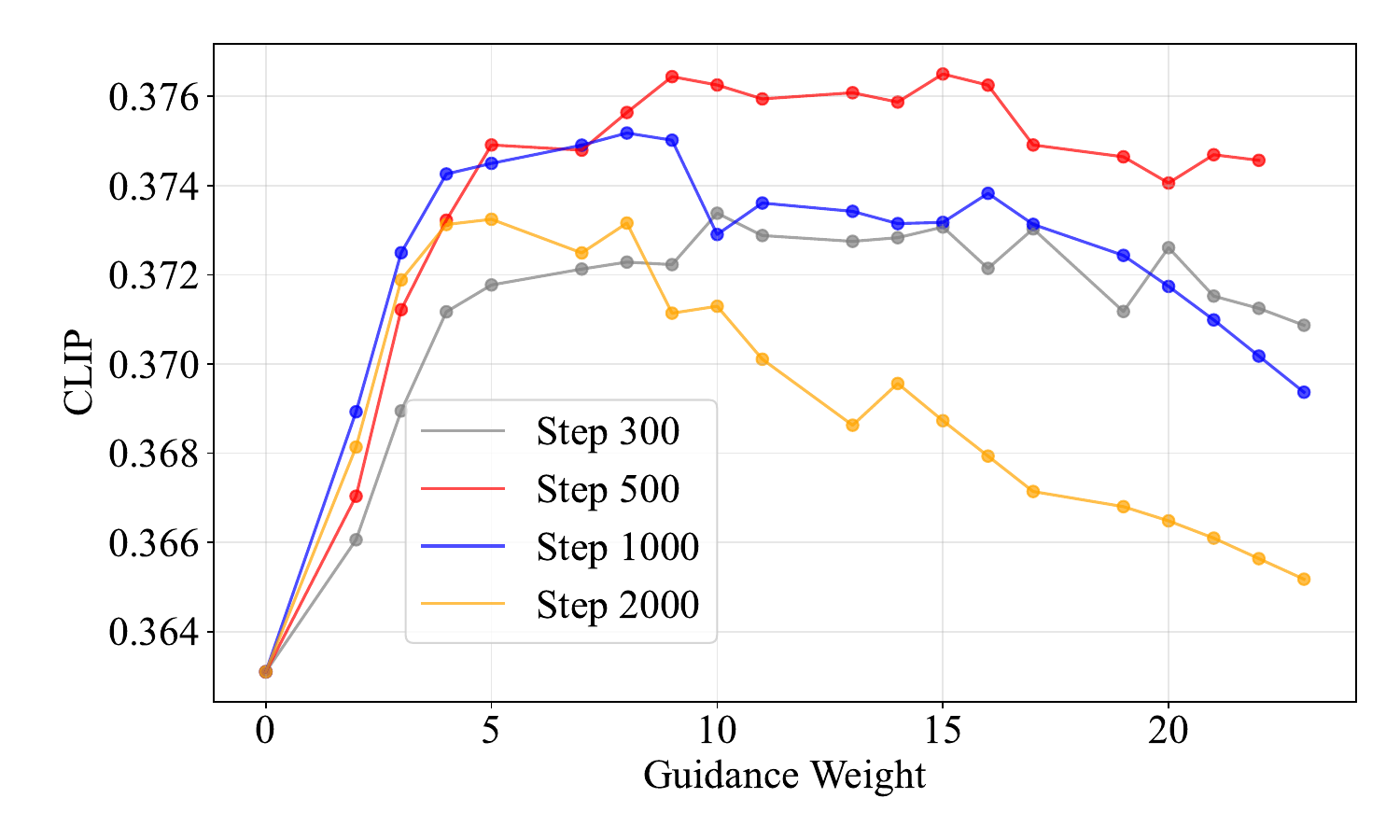} &
  \includegraphics[width=0.33\linewidth]{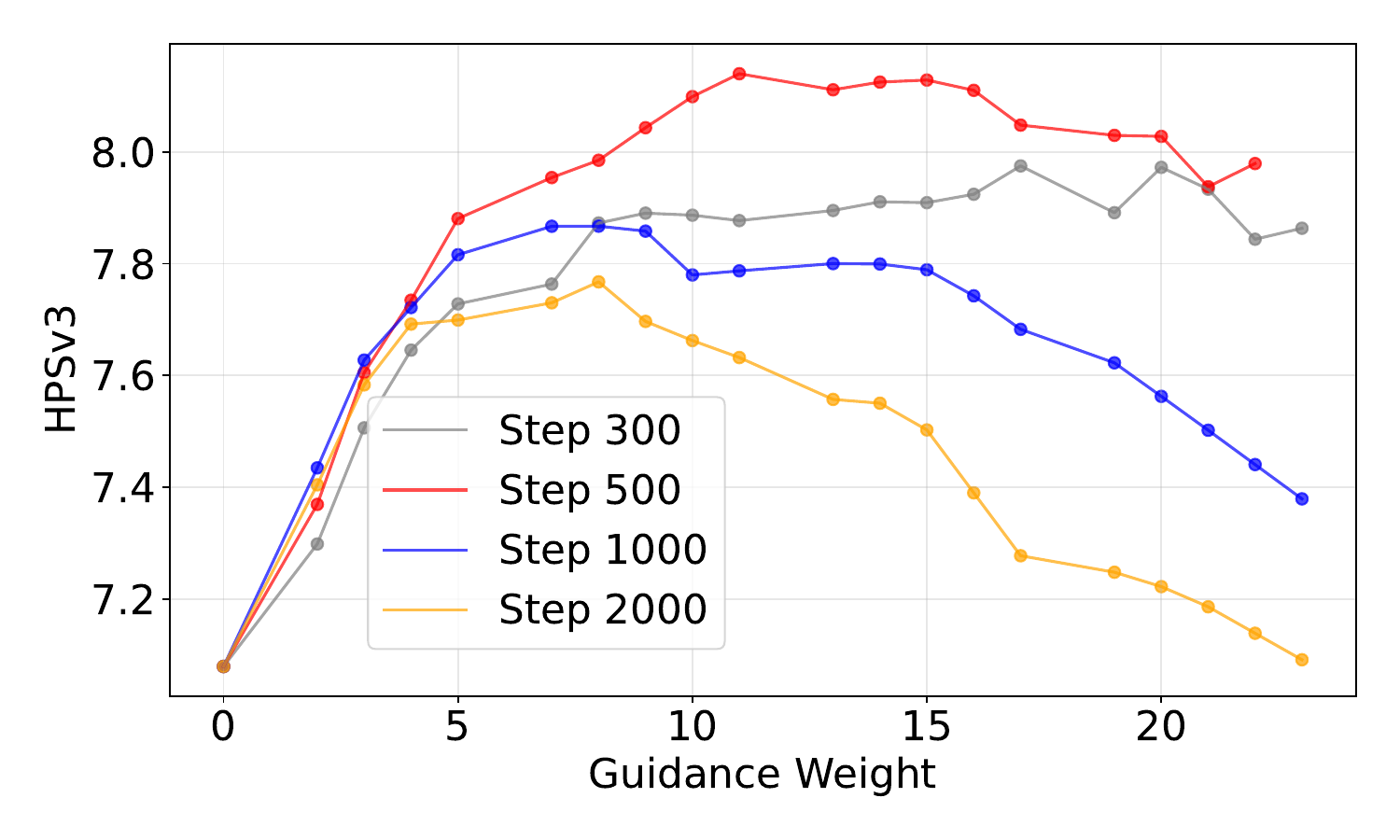} \\

\end{tabular}
}
\vspace{-4mm}
\captionsetup{labelfont=footnotesize}
\caption{
\footnotesize 
Effect of guidance weight $w$ on automatic metrics (SDXL). 
Left: PickScore; Middle: CLIP score; Right: HPSv3. 
“Step” denotes the training steps of the guidance module. 
Curves rise quickly for small $w$, 
}
\label{fig:step-weight}
\vspace{-4mm}
\end{figure*}

\noindent \textbf{Ablation on guidance weights.} Qualitatively, increasing guidance weights generally yields better reward following. as shown in Fig.~\ref{fig:weight_results}. To comprehensively quantify the effect of guidance weights in different hyperparameter settings, we measuring the performance metrics by varying the guidance weights for models finetuned on Pick-a-Pic v2 from SDXL with $300, 500, 1000$ and $2000$ steps. As shown in Fig.~\ref{fig:step-weight}, we observe that increasing guidance weights from $0$ to some moderate value (around $6$) generally leads to better reward values for all tested models, but beyond that the model performance drops. Models finetuned with less steps exhibit less amount of performance drop, which is likely due to the regularization effect of early stopping. Furthermore, Fig.~\ref{fig:fid} and \ref{fig:diversity} shows that 1) the increase of guidance weights leads to ``less natural'' images and 2) if the guidance weight is beyond certain threshold, increasing the guidance weight leads to more chaotic predictions.

\begin{table}
\footnotesize
\centering
\vspace{-1mm}
\captionof{table}{
\footnotesize 
Impact of preference data variance on alignment performance. 
“Subset” refers to training on a high-quality curated subset (low variance), 
while “Fullset” uses the full HPDv3 dataset (high variance). 
The 1st-best results are in \textbf{bold} and the 2nd-best are \uline{underlined}. 
All methods are applied to the base SDXL model.}
\label{tab:hpdv3_variance}
\renewcommand{\arraystretch}{1.2} 
\resizebox{\linewidth}{!}{
\begin{tabular}{c|c|cccccc}
\toprule
\textbf{Set} & \textbf{Method} & \textbf{PS} $\uparrow$ & \textbf{HPSv2} $\uparrow$ & \textbf{HPSv3} $\uparrow$ & \textbf{Aes} $\uparrow$ & \textbf{CLIP} $\uparrow$ & \textbf{IR} $\uparrow$ \\
\midrule
\multirow{4}{*}{Subset} 
& SDXL & 0.222 & 0.277 & 7.079 & 6.052 & 0.363 & 0.658 \\
& DPO   & 0.225 & 0.282 & 8.032 & 6.112 & 0.361 & 0.823 \\
& PGD   & \uline{0.226} & \uline{0.285} & \uline{9.459} & \uline{6.203} & \uline{0.364} & \uline{0.874} \\
& cPGD & \textbf{0.228} & \textbf{0.289} & \textbf{10.045} & \textbf{6.267} & \textbf{0.365} & \textbf{1.031} \\
\midrule
\multirow{4}{*}{Fullset} 
& SDXL & 0.222 & 0.277 & 7.079 & 6.052 & 0.363 & 0.658 \\
& DPO & 0.226 & 0.284 & 8.261 & 6.165 & \uline{0.366} & 0.918 \\
& PGD & \textbf{0.229} & \uline{0.287} & \textbf{10.065} & \textbf{6.505} & 0.364 & \uline{1.063} \\
& cPGD & \uline{0.227} & \textbf{0.290} & \uline{9.243} & \uline{6.179} & \textbf{0.373} & \textbf{1.143} \\
\bottomrule
\end{tabular}%
}
\captionsetup{labelfont=footnotesize}

\vspace{-4.5mm}
\end{table}

\noindent \textbf{Dataset quality.} Since the image distribution in the preference datasets can differ from the image distribution of the base models, here we investigate how methods are robust to different preference datasets, especially when the image quality differs a lot. In Table~\ref{tab:hpdv3_variance}, we show the results of finetuning on the full HPDv3 dataset, in which the variance of image quality is great, and on a high-quality subset. We find that our methods generally performs better in both cases, but on the high-quality subset our methods unanimously outperform the baselines. In addition, in the high-quality subset case, cPGD is unanimously better than PGD. We hypothesize that this is likely due to that cPGD imposes weaker assumptions on preference pairs as $\theta^+$ and $\theta^-$ are trained in an independent way.
Despite that the high-quality subset yields more consistent observations, using the full dataset generally leads to better reward values, in part simply due to the increased number of data points.

\section{Discussions}


\textbf{PGD as kernel method.} As shown in our experiments, PGD inference with slightly finetuned models consistently outperforms DPO methods. This behavior can be understood through the theory of neural tangent kernels (NTK)~\citep{jacot2018ntk}. In the \emph{lazy training regime}, i.e. when the finetuned model remains close to the reference model, we can write $\epsilon_\text{ref} + w(\epsilon_\text{finetuned} - \epsilon_\text{ref}) \approx \epsilon_\text{ref} + w K_\text{ref} \alpha$ where $K_\text{ref}$ is known as the NTK matrix of $\epsilon_\text{ref}$ and $\alpha$ is the vector of regression coefficients, which shows that PGD is essentially kernel regression in the NTK feature space of the reference model. Because this feature space is an intrinsic and stable property of $\epsilon_\text{ref}$, PGD inference leverages reliable features. In contrast, extended finetuning with large learning rates can push the model out of the lazy regime, causing the NTK approximation to break down and increasing the risk of overfitting on small datasets.

\begin{figure}[ht]
    \centering
    \vspace{-3pt}   
    \includegraphics[width=\linewidth]{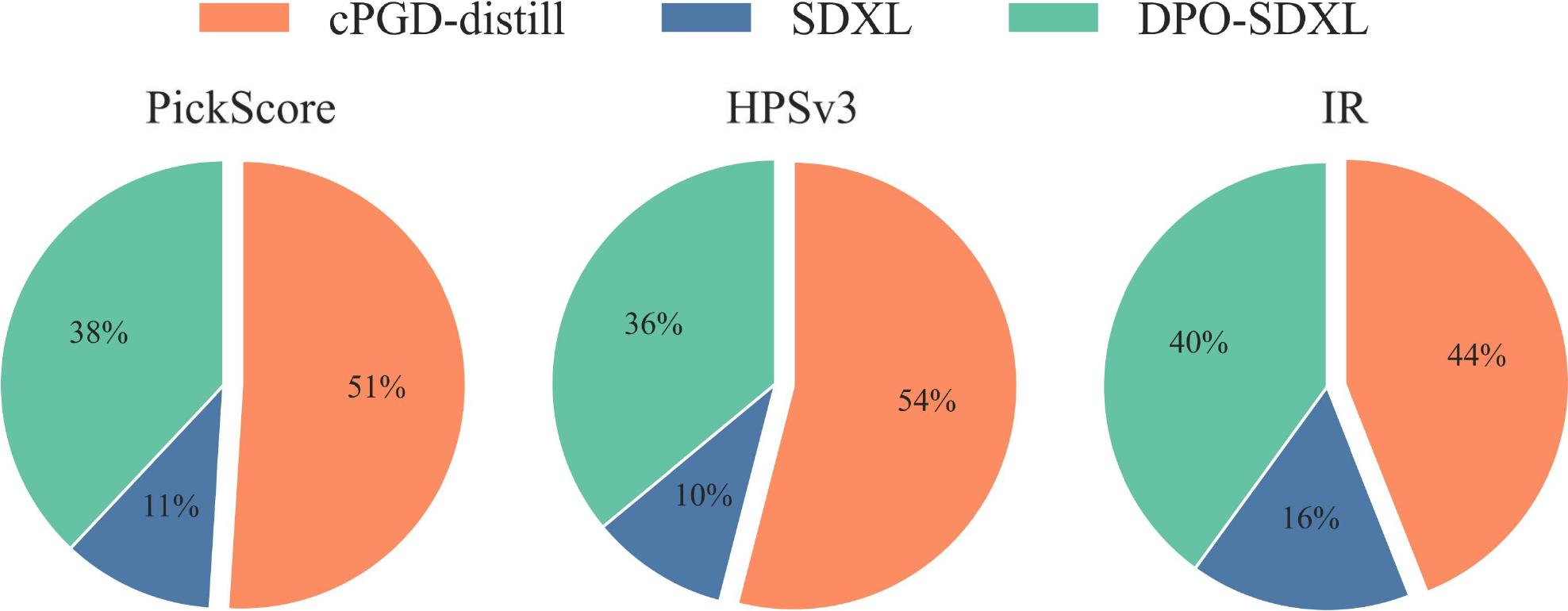}
    \captionsetup{labelfont=footnotesize}
    \caption{\footnotesize Win rate (\%) of different finetuned models on different reward models.
    For each prompt (with same random seeds), we score images from all three models and report the proportion on which each model achieves the highest score. Results show that cPGD-distill is superior.}
    \label{fig:distill_bar}
    \vspace{-5mm}
\end{figure}

\noindent \textbf{Inference cost.} While the inference time is doubled with PGD due to the need to compute outputs with the reference model, we note that it is possible to perform distillation so that one single model learns to predict the PGD outputs, as demonstrated by many other works on diffusion model distillation~\citep{salimans2022progressive, song2023consistency, meng2023distillation}. To verify this, we present our attempts in distilling a single model out of cPGD in Fig.~\ref{fig:distill_bar}, which clearly shows that the monolithic models distilled from cPGD-tuned models are still the best in all cases.

\vspace{-2mm}
\section{Conclusion}
\vspace{-2mm}

We introduced preference-guided diffusion (PGD), a simple yet effective method that better aligns diffusion models with human preference through the lens of classifier-free guidance: the finetuned model is the guidance signal of the dataset. By further take inspiration from the training of conditional diffusion models, we propose a variant called contrastive PGD (cPGD) which parameterize the finetuned module with two models independently trained on positive and negative samples, respectively. We empirically verify the effectiveness of the proposed methods on different datasets and base models.

{
    \small
    \bibliographystyle{ieeenat_fullname}
    \bibliography{main}

@String(CVPR= {IEEE Conf. Comput. Vis. Pattern Recog.})

@String(ICCV= {Int. Conf. Comput. Vis.})

@String(NIPS= {Adv. Neural Inform. Process. Syst.})

@String(ICLR = {Int. Conf. Learn. Represent.})

@String(CVPR  = {CVPR})

@String(ICCV  = {ICCV})

@String(NIPS  = {NeurIPS})

@String(ICLR  = {ICLR})

@inproceedings{lin-etal-2024-limited,
  title     = {On the Limited Generalization Capability of the Implicit Reward Model Induced by Direct Preference Optimization},
  author    = {Lin, Yong and Seto, Skyler and Ter Hoeve, Maartje and Metcalf, Katherine and Theobald, Barry-John and Wang, Xuan and Zhang, Yizhe and Huang, Chen and Zhang, Tong},
  booktitle = {Findings of the Association for Computational Linguistics: EMNLP 2024},
  pages     = {16015--16026},
  year      = {2024},
  month     = nov,
  address   = {Miami, Florida, USA},
  publisher = {Association for Computational Linguistics},
  doi       = {10.18653/v1/2024.findings-emnlp.940},
  url       = {https://aclanthology.org/2024.findings-emnlp.940/}
}

@inproceedings{blacktraining,
  title={Training Diffusion Models with Reinforcement Learning},
  author={Black, Kevin and Janner, Michael and Du, Yilun and Kostrikov, Ilya and Levine, Sergey},
  booktitle={The Twelfth International Conference on Learning Representations},
  year={2024}
}

@article{fan2023dpok,
  title={Dpok: Reinforcement learning for fine-tuning text-to-image diffusion models},
  author={Fan, Ying and Watkins, Olivia and Du, Yuqing and Liu, Hao and Ryu, Moonkyung and Boutilier, Craig and Abbeel, Pieter and Ghavamzadeh, Mohammad and Lee, Kangwook and Lee, Kimin},
  journal={Advances in Neural Information Processing Systems},
  volume={36},
  pages={79858--79885},
  year={2023}
}

@article{xu2024imagereward,
  title={Imagereward: Learning and evaluating human preferences for text-to-image generation},
  author={Xu, Jiazheng and Liu, Xiao and Wu, Yuchen and Tong, Yuxuan and Li, Qinkai and Ding, Ming and Tang, Jie and Dong, Yuxiao},
  journal={Advances in Neural Information Processing Systems},
  volume={36},
  year={2024}
}

@inproceedings{clarkdirectly,
  title={Directly Fine-Tuning Diffusion Models on Differentiable Rewards},
  author={Clark, Kevin and Vicol, Paul and Swersky, Kevin and Fleet, David J},
  booktitle={The Twelfth International Conference on Learning Representations},
  year={2024}
}

@misc{prabhudesai2023aligning,
      title={Aligning Text-to-Image Diffusion Models with Reward Backpropagation}, 
      author={Mihir Prabhudesai and Anirudh Goyal and Deepak Pathak and Katerina Fragkiadaki},
      year={2023},
      eprint={2310.03739},
      archivePrefix={arXiv},
      primaryClass={cs.CV}
}

@article{kirstain2023pick,
  title={Pick-a-pic: An open dataset of user preferences for text-to-image generation},
  author={Kirstain, Yuval and Polyak, Adam and Singer, Uriel and Matiana, Shahbuland and Penna, Joe and Levy, Omer},
  journal={Advances in Neural Information Processing Systems},
  volume={36},
  pages={36652--36663},
  year={2023}
}

@inproceedings{radford2021learning,
  title={Learning transferable visual models from natural language supervision},
  author={Radford, Alec and Kim, Jong Wook and Hallacy, Chris and Ramesh, Aditya and Goh, Gabriel and Agarwal, Sandhini and Sastry, Girish and Askell, Amanda and Mishkin, Pamela and Clark, Jack and others},
  booktitle={International conference on machine learning},
  pages={8748--8763},
  year={2021},
  organization={PmLR}
}

@misc{laionaes,
title={LAION-AESTHETICS},
howpublished={\url{https://laion.ai/blog/laion-aesthetics/}},
note={Accessed: 2023 - 11- 10},
author = {Christoph Schuhmann},
year = {2022}
}

@inproceedings{ouyang2022human,
author = {Ouyang, Long and Wu, Jeff and Jiang, Xu and Almeida, Diogo and Wainwright, Carroll L. and Mishkin, Pamela and Zhang, Chong and Agarwal, Sandhini and Slama, Katarina and Ray, Alex and Schulman, John and Hilton, Jacob and Kelton, Fraser and Miller, Luke and Simens, Maddie and Askell, Amanda and Welinder, Peter and Christiano, Paul and Leike, Jan and Lowe, Ryan},
title = {Training language models to follow instructions with human feedback},
year = {2022},
isbn = {9781713871088},
publisher = {Curran Associates Inc.},
address = {Red Hook, NY, USA},
booktitle = {Proceedings of the 36th International Conference on Neural Information Processing Systems},
articleno = {2011},
numpages = {15},
location = {New Orleans, LA, USA},
series = {NIPS '22}
}

@article{rafailov2023direct,
  title={Direct preference optimization: Your language model is secretly a reward model},
  author={Rafailov, Rafael and Sharma, Archit and Mitchell, Eric and Manning, Christopher D and Ermon, Stefano and Finn, Chelsea},
  journal={Advances in Neural Information Processing Systems},
  volume={36},
  pages={53728--53741},
  year={2023}
}

@inproceedings{wallace2024diffusion,
  title={Diffusion model alignment using direct preference optimization},
  author={Wallace, Bram and Dang, Meihua and Rafailov, Rafael and Zhou, Linqi and Lou, Aaron and Purushwalkam, Senthil and Ermon, Stefano and Xiong, Caiming and Joty, Shafiq and Naik, Nikhil},
  booktitle={Proceedings of the IEEE/CVF Conference on Computer Vision and Pattern Recognition},
  pages={8228--8238},
  year={2024}
}

@inproceedings{lialigning,
  title={Aligning Diffusion Models by Optimizing Human Utility},
  author={Li, Shufan and Kallidromitis, Konstantinos and Gokul, Akash and Kato, Yusuke and Kozuka, Kazuki},
  booktitle={The Thirty-eighth Annual Conference on Neural Information Processing Systems},
  year={2024}
}

@inproceedings{yang2024dense,
  title={A dense reward view on aligning text-to-image diffusion with preference},
  author={Yang, Shentao and Chen, Tianqi and Zhou, Mingyuan},
  booktitle={Proceedings of the 41st International Conference on Machine Learning},
  pages={55998--56032},
  year={2024}
}

@misc{hong2024marginaware,
    title={Margin-aware Preference Optimization for Aligning Diffusion Models without Reference}, 
    author={Jiwoo Hong and Sayak Paul and Noah Lee and Kashif Rasul and James Thorne and Jongheon Jeong},
    year={2024},
    eprint={2406.06424},
    archivePrefix={arXiv},
    primaryClass={cs.CV}
}

@inproceedings{
zhu2025dspo,
title={{DSPO}: Direct Score Preference Optimization for Diffusion Model Alignment},
author={Huaisheng Zhu and Teng Xiao and Vasant G Honavar},
booktitle={The Thirteenth International Conference on Learning Representations},
year={2025},
url={https://openreview.net/forum?id=xyfb9HHvMe}
}

@article{yu2022scaling,
  title={Scaling Autoregressive Models for Content-Rich Text-to-Image Generation},
  author={Yu, Jiahui and Xu, Yuanzhong and Koh, Jing Yu and Luong, Thang and Baid, Gunjan and Wang, Zirui and Vasudevan, Vijay and Ku, Alexander and Yang, Yinfei and Ayan, Burcu Karagol and others},
  journal={Trans. Mach. Learn. Res.},
  year={2022}
}

@inproceedings{ddpm,
author = {Ho, Jonathan and Jain, Ajay and Abbeel, Pieter},
title = {Denoising diffusion probabilistic models},
year = {2020},
booktitle = {Proceedings of the 34th International Conference on Neural Information Processing Systems}
}

@misc{diffusionnpo,
      title={Diffusion-NPO: Negative Preference Optimization for Better Preference Aligned Generation of Diffusion Models}, 
      author={Fu-Yun Wang and Yunhao Shui and Jingtan Piao and Keqiang Sun and Hongsheng Li},
      year={2025},
      eprint={2505.11245},
      archivePrefix={arXiv},
      primaryClass={cs.CV},
      url={https://arxiv.org/abs/2505.11245}, 
}

@article{liang2024step,
  title={Aesthetic Post-Training Diffusion Models from Generic Preferences with Step-by-step Preference Optimization},
  author={Liang, Zhanhao and Yuan, Yuhui and Gu, Shuyang and Chen, Bohan and Hang, Tiankai and Cheng, Mingxi and Li, Ji and Zheng, Liang},
  journal={arXiv preprint arXiv:2406.04314},
  year={2024}
}

@article{song2019generative,
  title={Generative modeling by estimating gradients of the data distribution},
  author={Song, Yang and Ermon, Stefano},
  journal={Advances in neural information processing systems},
  volume={32},
  year={2019}
}

@inproceedings{song2021score,
  author       = {Yang Song and
                  Jascha Sohl{-}Dickstein and
                  Diederik P. Kingma and
                  Abhishek Kumar and
                  Stefano Ermon and
                  Ben Poole},
  title        = {Score-Based Generative Modeling through Stochastic Differential Equations},
  booktitle    = {9th International Conference on Learning Representations, {ICLR} 2021,
                  Virtual Event, Austria, May 3-7, 2021},
  publisher    = {OpenReview.net},
  year         = {2021},
  url          = {https://openreview.net/forum?id=PxTIG12RRHS},
  bibsource    = {dblp computer science bibliography, https://dblp.org}
}

@inproceedings{saharia2022photo,
author = {Saharia, Chitwan and Chan, William and Saxena, Saurabh and Lit, Lala and Whang, Jay and Denton, Emily and Ghasemipour, Seyed Kamyar Seyed and Ayan, Burcu Karagol and Mahdavi, S. Sara and Gontijo-Lopes, Raphael and Salimans, Tim and Ho, Jonathan and Fleet, David J and Norouzi, Mohammad},
title = {Photorealistic text-to-image diffusion models with deep language understanding},
year = {2022},
isbn = {9781713871088},
publisher = {Curran Associates Inc.},
address = {Red Hook, NY, USA},
booktitle = {Proceedings of the 36th International Conference on Neural Information Processing Systems},
articleno = {2643},
numpages = {16},
location = {New Orleans, LA, USA},
series = {NIPS '22}
}

@misc{sdxl,
      title={SDXL: Improving Latent Diffusion Models for High-Resolution Image Synthesis}, 
      author={Dustin Podell and Zion English and Kyle Lacey and Andreas Blattmann and Tim Dockhorn and Jonas Müller and Joe Penna and Robin Rombach},
      year={2023},
      eprint={2307.01952},
      archivePrefix={arXiv},
      primaryClass={cs.CV}
}

@inproceedings{rombach2022high,
  title={High-resolution image synthesis with latent diffusion models},
  author={Rombach, Robin and Blattmann, Andreas and Lorenz, Dominik and Esser, Patrick and Ommer, Bj{\"o}rn},
  booktitle={Proceedings of the IEEE/CVF conference on computer vision and pattern recognition},
  pages={10684--10695},
  year={2022}
}

@misc{domingoenrich2025adjointmatchingfinetuningflow,
      title={Adjoint Matching: Fine-tuning Flow and Diffusion Generative Models with Memoryless Stochastic Optimal Control}, 
      author={Carles Domingo-Enrich and Michal Drozdzal and Brian Karrer and Ricky T. Q. Chen},
      year={2025},
      eprint={2409.08861},
      archivePrefix={arXiv},
      primaryClass={cs.LG},
      url={https://arxiv.org/abs/2409.08861}, 
}

@inproceedings{christiano2017deep,
  title={Deep reinforcement learning from human preferences},
  author={Christiano, Paul F and Leike, Jan and Brown, Tom and Martic, Miljan and Legg, Shane and Amodei, Dario},
  booktitle={Advances in Neural Information Processing Systems (NeurIPS)},
  year={2017}
}

@inproceedings{stiennon2020learning,
  title={Learning to summarize with human feedback},
  author={Stiennon, Nisan and Ouyang, Long and Wu, Jeff and Ziegler, Daniel M and Lowe, Ryan and Voss, Chelsea and Radford, Alec and Amodei, Dario and Christiano, Paul},
  booktitle={Advances in Neural Information Processing Systems (NeurIPS)},
  year={2020}
}

@article{azizzadenesheli2023mapo,
  title={MAPO: Model-Agnostic Preference Optimization},
  author={Azizzadenesheli, Kamyar and Sessa, Pier Giuseppe and Anand, Kartikeya and Anand, Ankesh and Fazel, Maryam},
  journal={arXiv preprint arXiv:2310.03708},
  year={2023}
}

@article{xu2024implicit,
  title={Implicit Preference Optimization: Aligning Language Models Without a Reward Model},
  author={Xu, Frank F and Li, Xuechen and Ladhak, Faisal and Durmus, Esin and Wei, Jason and Chen, Xiang Lorraine and Hashimoto, Tatsunori B},
  journal={arXiv preprint arXiv:2402.00856},
  year={2024}
}

@article{black2023training,
  title={Training diffusion models with reinforcement learning},
  author={Black, Sam and Gao, Leo and Biderman, Stella and Hallahan, Eric and Anthony, Quentin and Phang, Jason and Prakash, Shimao and Pfau, Janelle and Purohit, Shreyas and Foster, Chris and others},
  journal={arXiv preprint arXiv:2305.13301},
  year={2023}
}

@article{lee2023aligning,
  title={Aligning Text-to-Image Models using Human Feedback},
  author={Lee, Kuan-Hao and Xie, Saining and Zhang, Han and Zhang, Yiming and Zhang, Chong and Zettlemoyer, Luke and Hashimoto, Tatsunori B},
  journal={arXiv preprint arXiv:2302.12192},
  year={2023}
}

@inproceedings{dhariwal2021diffusion,
  title={Diffusion models beat GANs on image synthesis},
  author={Dhariwal, Prafulla and Nichol, Alex},
  booktitle={Advances in Neural Information Processing Systems (NeurIPS)},
  year={2021}
}

@inproceedings{ho2022classifierfree,
  title={Classifier-free diffusion guidance},
  author={Ho, Jonathan and Salimans, Tim},
  booktitle={Advances in Neural Information Processing Systems (NeurIPS) Workshop on Deep Generative Models},
  year={2022}
}

@article{nichol2021glide,
  title={GLIDE: Towards Photorealistic Image Generation and Editing with Text-Guided Diffusion Models},
  author={Nichol, Alex and Dhariwal, Prafulla and Ramesh, Aditya and Shyam, Pranav and Mishkin, Pamela and McGrew, Bob and Sutskever, Ilya and Chen, Mark},
  journal={arXiv preprint arXiv:2112.10741},
  year={2021}
}

@inproceedings{ruiz2023dreambooth,
  title={DreamBooth: Fine Tuning Text-to-Image Diffusion Models for Subject-Driven Generation},
  author={Ruiz, Nataniel and Li, Yuanzhen and Jampani, Varun and Pritch, Yael and Rubinstein, Michael and Aberman, Kfir},
  booktitle={Proceedings of the IEEE/CVF Conference on Computer Vision and Pattern Recognition (CVPR)},
  year={2023}
}

@article{liu2023plug,
  title={Plug-and-Play Diffusion Features for Text-Driven Image-to-Image Translation},
  author={Liu, Zhihao and Yu, Tan and Gu, Jiuxiang and Zhang, Ruixiang and Yang, Yi and Wang, Zhangyang and Zhou, Bolei},
  journal={arXiv preprint arXiv:2304.02883},
  year={2023}
}

@inproceedings{chefer2023attend,
  title={Attend-and-Excite: Attention-Based Semantic Guidance for Text-to-Image Diffusion Models},
  author={Chefer, Hila and Mokady, Ron and Bar, Amir and Zohar, Amit and Paiss, Roni and Wolf, Lior},
  booktitle={Proceedings of the IEEE/CVF International Conference on Computer Vision (ICCV)},
  year={2023}
}

@article{hu2021lora,
  title={LoRA: Low-Rank Adaptation of Large Language Models},
  author={Hu, Edward J. and Shen, Yelong and Wallis, Phillip and Allen-Zhu, Zeyuan and Li, Yuanzhi and Wang, Shean and Wang, Lu and Chen, Weizhu},
  journal={arXiv preprint arXiv:2106.09685},
  year={2021}
}

@article{gal2022image,
  title={An Image is Worth One Word: Personalizing Text-to-Image Generation using Textual Inversion},
  author={Gal, Rinon and Alaluf, Yuval and Atzmon, Yuval and Patashnik, Or and Bermano, Amit H. and Chechik, Gal and Cohen-Or, Daniel},
  journal={arXiv preprint arXiv:2208.01618},
  year={2022}
}

@misc{hpdv2,
      title={Human Preference Score v2: A Solid Benchmark for Evaluating Human Preferences of Text-to-Image Synthesis}, 
      author={Xiaoshi Wu and Yiming Hao and Keqiang Sun and Yixiong Chen and Feng Zhu and Rui Zhao and Hongsheng Li},
      year={2023},
      eprint={2306.09341},
      archivePrefix={arXiv},
      primaryClass={cs.CV},
      url={https://arxiv.org/abs/2306.09341}, 
}

@misc{hpdv3,
      title={HPSv3: Towards Wide-Spectrum Human Preference Score}, 
      author={Yuhang Ma and Yunhao Shui and Xiaoshi Wu and Keqiang Sun and Hongsheng Li},
      year={2025},
      eprint={2508.03789},
      archivePrefix={arXiv},
      primaryClass={cs.CV},
      url={https://arxiv.org/abs/2508.03789}, 
}

@inproceedings{rafailov2023dpo,
  title     = {Direct Preference Optimization: Your Language Model is Secretly a Reward Model},
  author    = {Rafailov, Rafael and Sharma, Archit and Mitchell, Eric and Ermon, Stefano and Finn, Chelsea},
  booktitle = {Advances in Neural Information Processing Systems (NeurIPS)},
  year      = {2023},
  url       = {https://arxiv.org/abs/2305.18290}
}

@article{wu2025dft,
  title   = {On the Generalization of SFT: A Reinforcement Learning Perspective with Reward Rectification},
  author  = {Wu, Xuehai and Zhang, Jinghan and Dong, Haonan and Li, Yujia and Hu, Wenlong and Zhao, Wayne Xin and Wen, Ji-Rong},
  journal = {arXiv preprint arXiv:2508.05629},
  year    = {2025},
  url     = {https://arxiv.org/abs/2508.05629}
}

@article{wang2023mesh,
  title={Mesh-RFT: Reframing Text-to-Image Diffusion for 3D Mesh Generation},
  author={Wang, Yifan and others},
  journal={arXiv preprint arXiv:2303.XXXX},
  year={2023}
}

@article{blattmann2023stablevideo,
  title={Stable Video Diffusion: Scaling Latent Video Diffusion Models to Large Datasets},
  author={Blattmann, Andreas and others},
  journal={arXiv preprint arXiv:2311.15127},
  year={2023}
}

@inproceedings{wu2023tuneavideo,
  title={Tune-a-Video: One-Shot Tuning of Image Diffusion Models for Text-to-Video Generation},
  author={Wu, Tianxing and Ge, Yijun and Zhang, Xintao and others},
  booktitle={ICCV},
  year={2023}
}

@article{khachatryan2023text2video,
  title={Text2Video-Zero: Text-to-Image Diffusion Models are Zero-Shot Video Generators},
  author={Khachatryan, Levon and others},
  journal={arXiv preprint arXiv:2303.13439},
  year={2023}
}

@inproceedings{jacot2018ntk,
  title={Neural Tangent Kernel: Convergence and Generalization in Neural Networks},
  author={Jacot, Arthur and Gabriel, Franck and Hongler, Clément},
  booktitle={Advances in Neural Information Processing Systems},
  volume={31},
  year={2018}
}

@inproceedings{salimans2022progressive,
  title={Progressive Distillation for Fast Sampling of Diffusion Models},
  author={Salimans, Tim and Ho, Jonathan},
  booktitle={International Conference on Learning Representations},
  year={2022}
}

@inproceedings{song2023consistency,
  title={Consistency Models},
  author={Song, Yang and Meng, Chenlin and Ermon, Stefano},
  booktitle={International Conference on Machine Learning},
  year={2023}
}

@article{meng2023distillation,
  title={On Distillation of Guided Diffusion Models},
  author={Meng, Chenlin and Song, Yang and Song, Jiaming and Wu, Jiaming and Dhariwal, Prafulla and Nichol, Alexey and Chen, Xinyang and Sohl-Dickstein, Jascha and Ermon, Stefano},
  journal={arXiv preprint arXiv:2306.05544},
  year={2023}
}

@article{bradley1952rank,
  title={Rank analysis of incomplete block designs: I. The method of paired comparisons},
  author={Bradley, Ralph Allan and Terry, Milton E.},
  journal={Biometrika},
  volume={39},
  number={3/4},
  pages={324--345},
  year={1952},
  publisher={Oxford University Press}
}

@inproceedings{heusel2017gans,
  title={GANs Trained by a Two Time-Scale Update Rule Converge to a Local Nash Equilibrium},
  author={Heusel, Martin and Ramsauer, Hubert and Unterthiner, Thomas and Nessler, Bernhard and Hochreiter, Sepp},
  booktitle={Advances in Neural Information Processing Systems (NeurIPS)},
  volume={30},
  year={2017}
}

@article{levine2018reinforcement,
  title={Reinforcement learning and control as probabilistic inference: Tutorial and review},
  author={Levine, Sergey},
  journal={arXiv preprint arXiv:1805.00909},
  year={2018}
}

@inproceedings{cherti2023reproducible,
  title        = {Reproducible scaling laws for contrastive language-image learning},
  author       = {Cherti, Mehdi and Beaumont, Romain and Wightman, Ross and Wortsman, Mitchell
                  and Ilharco, Gabriel and Gordon, Cade and Schuhmann, Christoph
                  and Schmidt, Ludwig and Jitsev, Jenia},
  booktitle    = {Proceedings of the IEEE/CVF Conference on Computer Vision and Pattern Recognition},
  pages        = {2818--2829},
  year         = {2023}
}

@inproceedings{liu2025nablagfn,
    title={Efficient Diversity-Preserving Diffusion Alignment via Gradient-Informed GFlowNets},
    author={Liu, Zhen and Xiao, Tim Z. and Liu, Weiyang and Bengio, Yoshua and Zhang, Dinghuai},
    booktitle={International Conference on Learning Representations},
    year={2025},
}

@article{song2020denoising,
  title        = {Denoising Diffusion Implicit Models},
  author       = {Song, Jiaming and Meng, Chenlin and Ermon, Stefano},
  journal      = {arXiv preprint arXiv:2010.02502},
  year         = {2020}
}

@misc{koala,
    title={KOALA: Empirical Lessons Toward Memory-Efficient and Fast Diffusion Models for Text-to-Image Synthesis}, 
    author={Youngwan Lee and Kwanyong Park and Yoorhim Cho and Yong-Ju Lee and Sung Ju Hwang},
    year={2023},
    eprint={2312.04005},
    archivePrefix={arXiv},
    primaryClass={cs.CV}
}
}


\clearpage
\setcounter{page}{1}

\section{Further Implementation Details}

\noindent\textbf{Software and precision.}
All models are implemented in \texttt{PyTorch} with \texttt{diffusers} and memory–efficient attention (\texttt{xFormers}). Mixed precision (fp16/bf16) is enabled for both training and inference.

\noindent\textbf{Sampling and guidance.}
Unless otherwise noted, SDXL uses the default 50-step  DDIM~\citep{song2020denoising} sampler; for fairness, SD1.5 is also run with 50 steps. 
Text-to-image generation employs classifier-free guidance (CFG), with the CFG weight fixed to 5.0 for SDXL and 7.5 for SD1.5. 
All other sampling hyperparameters are kept identical across methods.

\noindent\textbf{Training protocol.}
Unless noted, we follow the optimizer choices and step counts in Sec.~\ref{sec:exp_seting} (DPO/PGD main experiments: 2k training steps; distilled/cPGD ablations: 500 training steps). 
VAE weights are kept frozen; all methods share identical image resolutions and augmentations (random resize–crop and horizontal flip with probability 0.5).

\begin{figure}[hb]
\vspace{3mm}
\centering
\includegraphics[width=\linewidth]{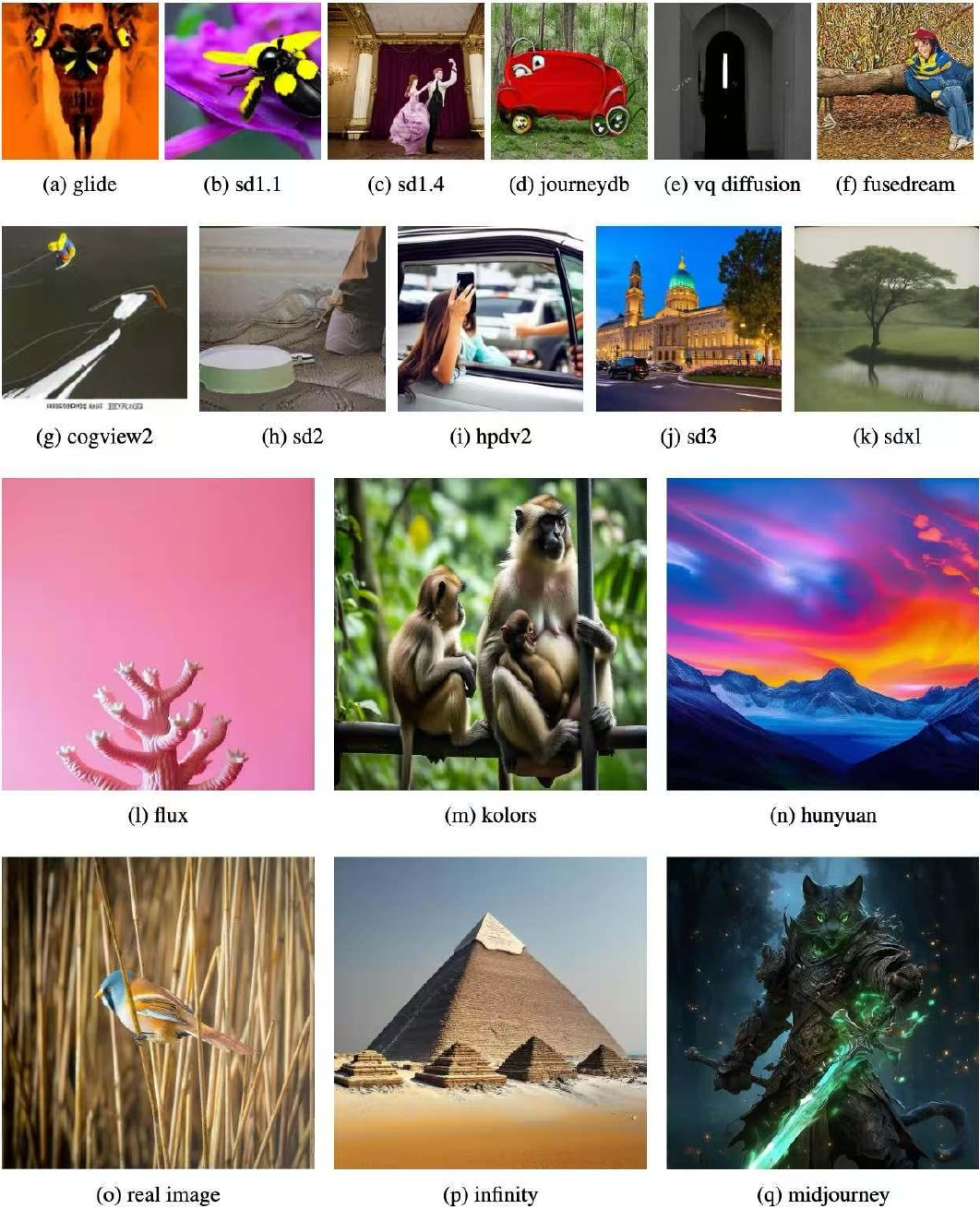}
\caption{Negative pair examples of HPDv3}
\label{fig:hpdv3_negtive_images}
\vspace{2mm}
\end{figure}

\begin{figure}[ht]
\vspace{2mm}
\centering
\includegraphics[width=\linewidth]{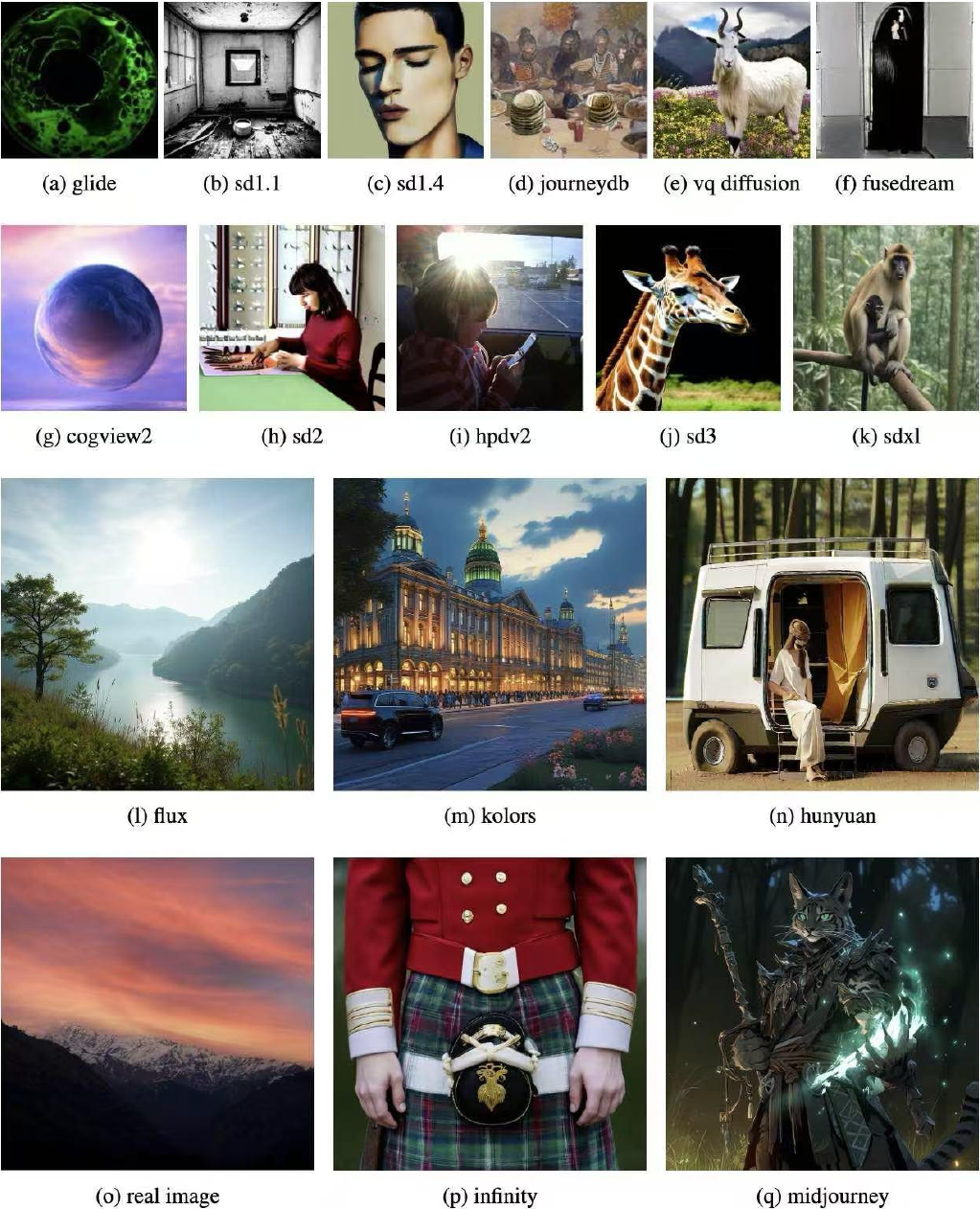}
\caption{Positive pair examples of HPDv3}
\label{fig:hpdv3_positive_images}
\vspace{-3mm}
\end{figure}

\noindent\textbf{HPDv3 subset construction.}
To obtain a \emph{curated} low-variance (in image quality) subset, we restrict HPDv3 to pairs whose two candidate images both have recorded provenance in
\{\texttt{flux}, \texttt{kolors}, \texttt{sd3}, \texttt{sdxl}, \texttt{hunyuan}, \texttt{real\_images}, \texttt{infinity}, \texttt{midjourney}\}.
Pairs are discarded if either side falls outside this list or has unknown origin.  Some examples are shown in Fig.~\ref{fig:hpdv3_negtive_images} and Fig.~\ref{fig:hpdv3_positive_images}

\noindent\textbf{Evaluation protocol.}
We compute PickScore, HPSv2, HPSv3, ImageReward, CLIP Score, and Aesthetics (Aes) using their public checkpoints and each model’s default preprocessing. 
Win rate is the percentage of prompts for which a method’s image scores higher than the base model under identical prompts, sampler settings, and a fixed random seed; for win rate, we generate one image per prompt per method.
FID and CLIP-Diversity follow Appendix
~\emph{Computation of Evaluation Metrics}: we sample $K{=}25$ prompts and, for each prompt, generate $n{=}40$ images using 40 distinct seeds, then aggregate across the $K$ prompts.

\section{Computation of Evaluation Metrics}
\label{sec:EvaluationMetrics}

\textbf{Diversity.}
To quantify the diversity of generated images, we adopt the CLIP-based diversity score \citep{domingoenrich2025adjointmatchingfinetuningflow, liu2025nablagfn}. This metric measures the variance within a set of images generated from a single prompt. Formally, for the $k$-th prompt, we generate $n{=}40$ images $\{g_i^k\}_{i=1}^{40}$ using the \texttt{open\_clip} encoder $\phi(\cdot)$~\citep{cherti2023reproducible}, and average the pairwise squared $\ell_2$ distances of their embeddings. The final diversity score is then averaged over $K{=}25$ distinct prompts:
{\small
\begin{equation*}
\mathrm{CLIP\text{-}Diversity} = \frac{1}{K} \sum_{k=1}^{K} \frac{2}{n(n-1)} \sum_{1 \le i < j \le n} \bigl\| \phi(g_i^k) - \phi(g_j^k) \bigr\|_2^{2}
\end{equation*}
}

\noindent \textbf{FID.}
We also report Fréchet Inception Distance (FID; \citet{heusel2017gans}) as a measure of distributional shift to the SDXL. The two feature distributions are:
(i) embeddings of images generated by the SDXL reference using the same prompt set and default sampling settings; and
(ii) embeddings of images generated by the method under evaluation (e.g., DPO, PGD variants) with identical prompts and seeds.
We extract 2048-d pool3 activations from a pre-trained Inception-V3 after standard preprocessing (resize/crop to $299{\times}299$ and Inception normalization), and compute FID between the two empirical distributions.

\section{More Details of the Toy 2D Example}
In this section, we give a more detailed illustration for the toy 2D experiment used in Fig.~\ref{fig:toy_exp} of the main paper to understand the behavior of DPO and PGD. 

\begin{figure}[ht]
\centering
\includegraphics[width=\linewidth]{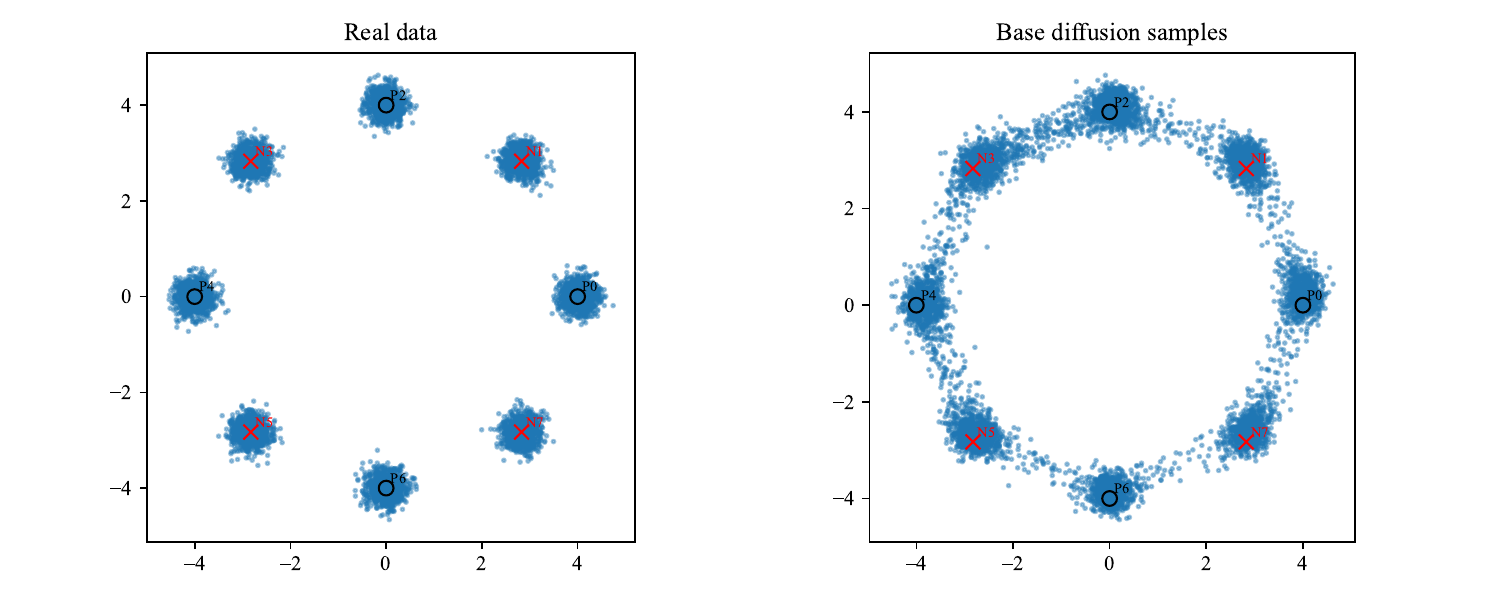}
\caption{The real 2-D data and the pretrain distribution on the Toy 2D experiment.}
\label{fig:toy2d-real}
\end{figure}

\paragraph{Setup.}
The data distribution is an 8-Gaussians mixture in $\mathbb{R}^2$: we sample points from eight well-separated Gaussian ``balls'' and assign alternating clusters as positive ($\mathrm{P}0,\mathrm{P}2,\mathrm{P}4,\mathrm{P}6$) and negative ($\mathrm{N}1,\mathrm{N}3,\mathrm{N}5,\mathrm{N}7$) samples.
We first train a small diffusion model parameterized by a 3-layer MLP on the full mixture using the standard diffusion loss; this model serves as the pretrained base policy.
Figure~\ref{fig:toy2d-real} visualizes the ground-truth data distribution and samples from the pretrained diffusion model, showing that the base model already learns a reasonable approximation of the mixture but does not distinguish between positive and negative clusters.

\begin{figure}[ht]
\centering
\includegraphics[width=\linewidth]{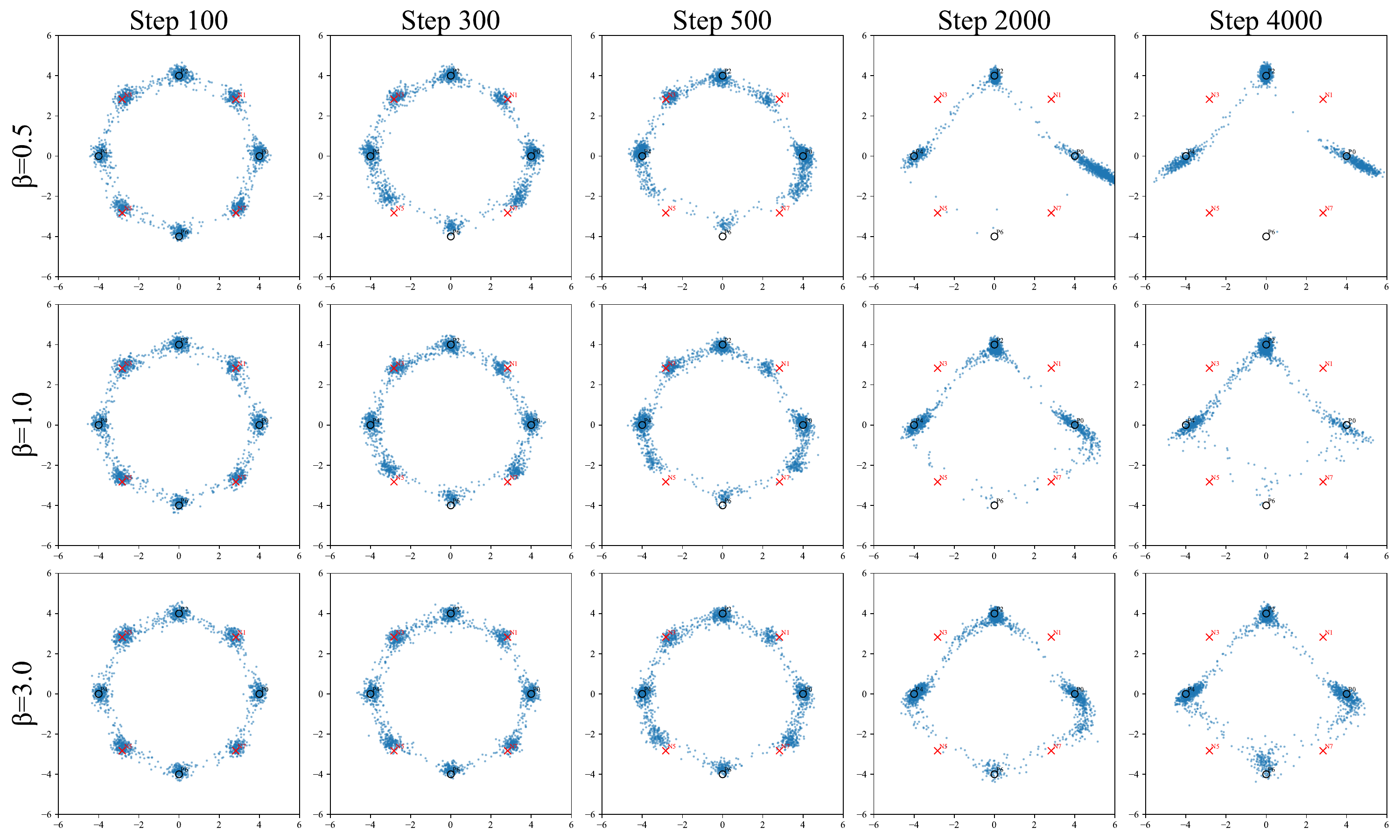}
\caption{Toy 2D experiment on DPO with different values of $\beta$.}
\label{fig:toy2d-dpo}
\vspace{-3mm}
\end{figure}


\begin{figure}[ht]
\centering
\includegraphics[width=\linewidth]{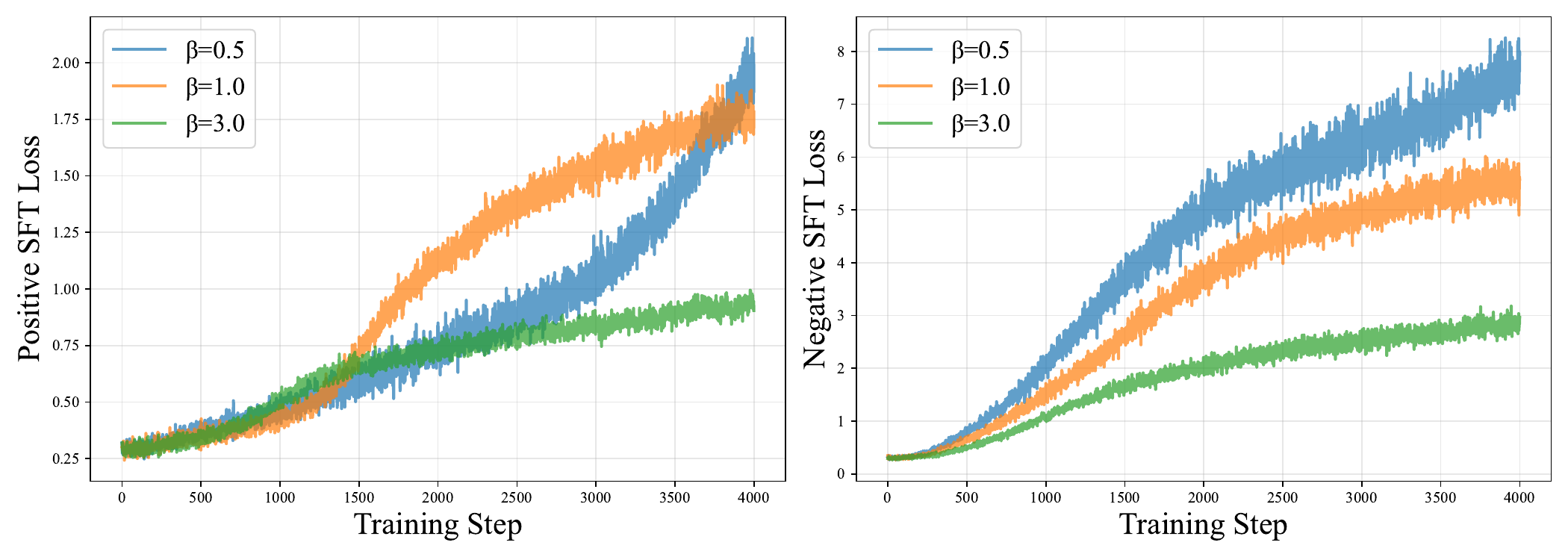}
\caption{The positive and negative SFT components of the DPO loss on the toy 2D experiment.}
\label{fig:toy2d-dpo-loss}
\vspace{-3mm}
\end{figure}

\noindent\textbf{DPO collapse and decomposition of the DPO objective.}
 We then apply Diffusion-DPO fine-tuning on randomly sampled preference pairs, using the pretrained diffusion model as the reference policy.  Figure~\ref{fig:toy2d-dpo} shows samples at multiple training steps for $\beta \in \{0.5, 1.0, 3.0\}$. We can observe that, as training proceeds, different degrees of collapse and sample imbalance emerge.
To better understand why dpo leads to instability, we decompose the Diffusion-DPO objective into its positive and negative SFT-like components.
For each preference pair $(x^+, x^-)$ with condition $c$, we track the mean log-likelihoods
\begin{align}
\mathcal{L}_{\text{pos}}(\theta) &= -\,\mathbb{E}_{(x^+,x^-)}\big[\log \pi_\theta(x^+ \mid c)\big], \\
\mathcal{L}_{\text{neg}}(\theta) &= -\,\mathbb{E}_{(x^+,x^-)}\big[\log \pi_\theta(x^- \mid c)\big],
\end{align}
which correspond to SFT-style losses on winners and losers, respectively.
Ideally, DPO should \emph{decrease} $\mathcal{L}_{\text{pos}}$ (increase $p_\theta(x^+ \mid c)$) and \emph{increase} $\mathcal{L}_{\text{neg}}$ (decrease $p_\theta(x^- \mid c)$.
However, as shown in Fig.~\ref{fig:toy2d-dpo-loss}, under large $\beta$ both quantities decrease over training, meaning that the probabilities of \emph{both} positive and negative samples are driven down.
We hypothesize that this unconstrained shrinkage of likelihood is a key reason for DPO collapse: the objective does not explicitly prevent such a degenerate solution.

\begin{figure}[ht]
\centering
\includegraphics[width=\linewidth]{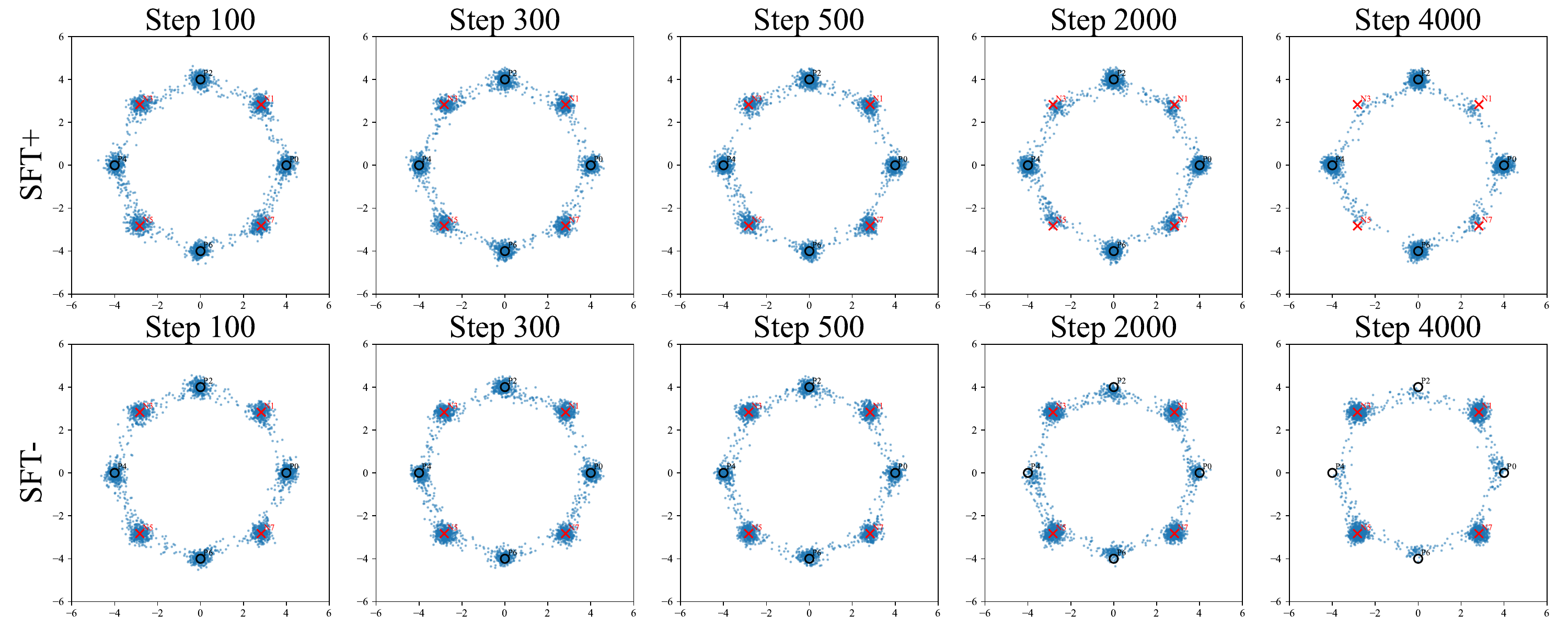}
\caption{Toy 2D experiment on SFT. The top row shows SFT on positive samples only (SFT+); the bottom row shows SFT on the negative samples (SFT-).}
\label{fig:toy2d-sft}
\vspace{-3mm}
\end{figure}

\paragraph{SFT baselines.}
We also consider a supervised fine-tuning (SFT) baseline trained with the standard diffusion loss, which is stable and requires no extra hyperparameters. In \emph{SFT+}, we discard negative samples and finetune only on positive clusters. As shown in Fig.~\ref{fig:toy2d-sft}, \emph{SFT+} trains stably and fits the multimodal distribution well, which also explains the stability and effectiveness of cPGD.

\begin{figure}[ht]
\centering
\includegraphics[width=\linewidth]{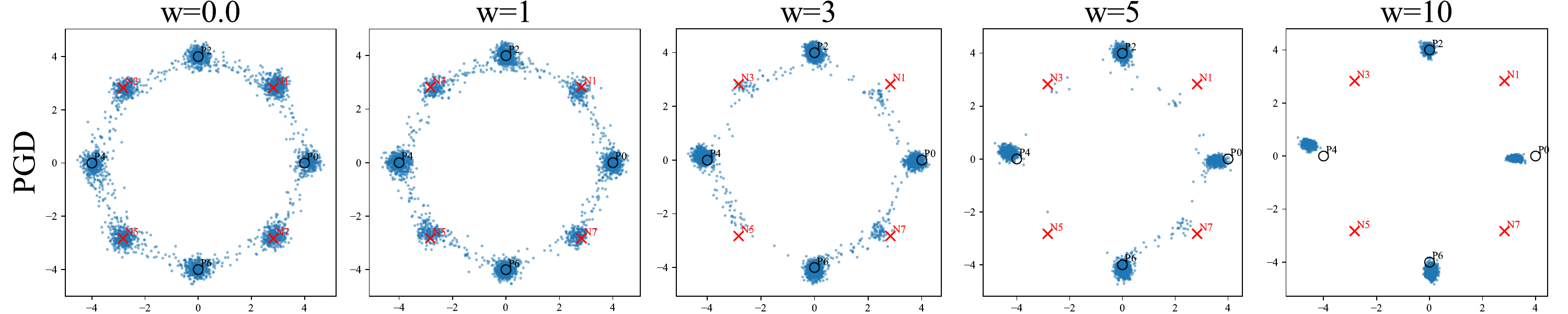}
\caption{Toy 2D experiment on PGD for different guidance weights $w$.}
\label{fig:toy2d-pgd}
\vspace{-3mm}
\end{figure}

\paragraph{Effect of PGD guidance.}
Finally, we apply Preference-Guided Diffusion (PGD) by treating the DPO-tuned model as the ``conditional'' policy and the pretrained diffusion model as the prior, and combining them via a CFG-style guidance rule at inference time.
We fix a DPO checkpoint and vary the guidance weight $w$ that scales the difference between the DPO and base predictions.
Figure~\ref{fig:toy2d-pgd} reports the resulting sample distributions for $w \in \{0,1,3,5,10\}$.
When $w=0$, PGD reduces to sampling from the base diffusion model.
As $w$ increases, probability mass is gradually shifted toward the positive clusters while the prior prevents complete collapse and preserves the global geometry of the mixture.
Extremely large $w$ reproduces the over-confident DPO behavior, whereas moderate values (e.g., $w=3$) provide a favorable trade-off: they substantially increase mass on the positive clusters while still retaining diversity and avoiding the severe mode collapse observed in pure DPO training.

\section{A New Perspective on cPGD via Maximum Entropy Learning}
\label{sec:maxent-cpgd}

In this appendix we give a maximum-entropy interpretation of cPGD and make its connection to preference learning explicit. We also explain its relationship to the RLHF objective.

\paragraph{Setting.}
Let $\mathbf{x}_t$ denote the latent at diffusion step $t$, $c$ a textual condition (prompt), and the pairwise preference label associated with a triplet $(x^w,x^l,c)$ drawn from a preference dataset $\mathcal{D}^p$, where $x^w \succ x^l$ means the winner $x^w$ is preferred over the loser $x^l$ under condition $c$.
We study the (conditional) score of the \emph{preference posterior} $p_\theta(\mathbf{x}_t \mid c,\mathcal{D}^p)$ and relate it to a contrastive combination of \emph{learnable} scores.

\paragraph{Score decomposition.}
By Bayes' rule, the conditional score decomposes into a base (likelihood) term and a preference-likelihood term:
\begin{equation}
\begin{aligned}    
\label{eq:score-decomp}
&\nabla_{\mathbf{x}_t}\log p_\theta(\mathbf{x}_t \mid c,\mathcal{D}^p)
= \\
&\nabla_{\mathbf{x}_t}\log p_\theta(\mathbf{x}_t \mid c)
+
\nabla_{\mathbf{x}_t}\log p(\mathcal{D}^p \mid \mathbf{x}_t,c),
\end{aligned}
\end{equation}
and, taking expectation over the data distribution,
\begin{equation}
\begin{aligned}
\label{eq:pref-exp}
&\nabla_{\mathbf{x}_t}\log p(\mathcal{D}^p \mid \mathbf{x}_t,c)
=\\
&\mathbb{E}_{(x^w,x^l,c)\sim\mathcal{D}^p}
\left[
\nabla_{\mathbf{x}_t}\log p(x^w \succ x^l \mid \mathbf{x}_t,c)
\right].
\end{aligned}
\end{equation}

\paragraph{From pairwise preferences to reward gradients.}
Assume a Bradley--Terry (BT) preference model with a state-dependent utility $R(\cdot \mid \mathbf{x}_t,c)$.
Let $\Delta R(\mathbf{x}_t) := R(x^{w}\mid \mathbf{x}_t,c) - R(x^{l}\mid \mathbf{x}_t,c)$ denote the reward gap between the winner and the loser.
Under the BT model we have
\begin{equation}
\label{eq:bt-grad}
\begin{aligned}
&\nabla_{\mathbf{x}_t}\log p(x^w \succ x^l \mid \mathbf{x}_t,c)\\
&= \nabla_{\mathbf{x}_t}\log \sigma\!\left(\Delta R(\mathbf{x}_t)\right) \\
&= \sigma\!\left(-\Delta R(\mathbf{x}_t)\right)\,
\nabla_{\mathbf{x}_t}\!\left[\Delta R(\mathbf{x}_t)\right] \\
&= \sigma\!\left(-\Delta R(\mathbf{x}_t)\right)\,
\Big(\nabla_{\mathbf{x}_t}R(x^w \mid \mathbf{x}_t,c)
      - \nabla_{\mathbf{x}_t}R(x^l \mid \mathbf{x}_t,c)\Big).
\end{aligned}
\end{equation}
The scalar factor $\sigma\!\left(-\Delta R(\mathbf{x}_t)\right)$ is precisely the probability of a \emph{preference inversion}, i.e.,
\(
\sigma\!\left(-\Delta R(\mathbf{x}_t)\right)
= P\big(x^{l} \succ x^{w}\,\big|\,\mathbf{x}_t,c\big),
\)
the event that the nominal loser $x^l$ is actually preferred over $x^w$.
In practice, instead of keeping this factor state-dependent, we approximate it by a dataset-level constant $w \in (0,0.5)$ that summarizes the empirical inversion rate, so that
\begin{equation}
\begin{aligned}
&\nabla_{\mathbf{x}_t}\log p(x^w \succ x^l \mid \mathbf{x}_t,c) \\
&\simeq w\,\Big(\nabla_{\mathbf{x}_t}R(x^w \mid \mathbf{x}_t,c)
               - \nabla_{\mathbf{x}_t}R(x^l \mid \mathbf{x}_t,c)\Big).
\end{aligned}
\end{equation}
This quantity is observable and can be estimated from repeated human re-annotation (or from a fitted Bradley--Terry model).
Because image preferences are inherently subjective, the inversion probability is strictly positive and dataset-dependent.
For example, the Pick-a-Pic paper~\citep{kirstain2023pick} reports a re-annotation agreement of roughly $70\%$, which implies an inversion rate of about $30\%$; hence a data-driven and practically meaningful setting is $w \approx 0.3$.
In our experiments we therefore treat $w$ as a fixed constant (unless otherwise stated), and note that moderate perturbations around $0.3$ do not materially affect performance.
The restriction $w<0.5$ preserves the sign of the guidance direction and avoids overcompensating the base score in the presence of annotation noise.

\noindent\textbf{Relationship between maximum-entropy RL and the RLHF objective.}
Consider the standard RLHF objective with a reference policy $\pi_\text{ref}$:
\begin{equation}
\label{eq:maxent}
\max_{\pi_\theta}~
\mathbb{E}_{\mathbf{x}_0 \sim \pi_\theta(\cdot\mid \mathbf{c})}
\left[ R(\mathbf{c},\mathbf{x}_0) \right]
- \beta\, D_{\mathrm{KL}}\!\left(\pi_\theta(\cdot\mid \mathbf{c}) \,\Vert\, \pi_\text{ref}(\cdot\mid \mathbf{c})\right).
\end{equation}
Expanding the KL divergence gives
\begin{equation}
\begin{aligned}
&D_{\mathrm{KL}}(\pi_\theta \Vert \pi_\text{ref})
=
\mathbb{E}_{\mathbf{x}_0 \sim \pi_\theta}\!\left[
\log \frac{\pi_\theta(\mathbf{x}_0 \mid \mathbf{c})}
     {\pi_\text{ref}(\mathbf{x}_0 \mid \mathbf{c})}
\right] \\
&=
\underbrace{\mathbb{E}_{\mathbf{x}_0 \sim \pi_\theta}\big[\log \pi_\theta(\mathbf{x}_0 \mid \mathbf{c})\big]}_{-\mathcal{H}(\pi_\theta)}
-\underbrace{\mathbb{E}_{\mathbf{x}_0 \sim \pi_\theta}\big[\log \pi_\text{ref}(\mathbf{x}_0 \mid \mathbf{c})\big]}_{-H(\pi_\theta,\pi_\text{ref})},
\end{aligned}
\end{equation}
where $\mathcal{H}(\pi_\theta)$ is the Shannon entropy of $\pi_\theta$, and
$H(\pi_\theta,\pi_\text{ref})$ is the cross-entropy between $\pi_\theta$ and $\pi_\text{ref}$.
Substituting this into \eqref{eq:maxent} yields
\begin{equation}
\label{eq:maxent-obj}
\max_{\pi_\theta}~
\mathbb{E}\!\left[R(\mathbf{c},\mathbf{x}_0)\right]
+ \beta\,\mathcal{H}(\pi_\theta)
- \beta\,H(\pi_\theta,\pi_\text{ref}).
\end{equation}

This expression shows that RLHF balances three effects:
maximizing expected reward, maximizing the entropy of $\pi_\theta$, and
penalizing deviation from the reference distribution via the cross-entropy term.
A particularly illuminating special case is when the reference policy is
\emph{uniform} over a finite support $\mathcal{X}$, i.e.,
$\pi_\text{ref}(\mathbf{x}_0\mid\mathbf{c}) = 1/|\mathcal{X}|$.
Then
\begin{equation}
H(\pi_\theta,\pi_\text{ref})
= -\mathbb{E}_{\mathbf{x}_0\sim\pi_\theta}\log \tfrac{1}{|\mathcal{X}|}
= \log |\mathcal{X}|,
\end{equation}
which is a constant independent of $\pi_\theta$.
In this case, the RLHF objective \eqref{eq:maxent-obj} is equivalent (up to an additive constant) to the classical maximum-entropy RL objective
\begin{equation}
\label{eq:maxent-simplified}
\max_{\pi_\theta}~
\mathbb{E}\!\left[R(\mathbf{c},\mathbf{x}_0)\right]
+ \beta\,\mathcal{H}(\pi_\theta).
\end{equation}

In our setting, $\pi_\text{ref}$ is not uniform but given by a pretrained base diffusion model.
The cross-entropy term $H(\pi_\theta,\pi_\text{ref})$ can be accounted for at inference time by using a reference-guided sampler (analogous to classifier-free guidance), so that training primarily optimizes the reward-plus-entropy part \eqref{eq:maxent-simplified} while the reference model supplies the base distribution at sampling time.

\paragraph{Reward--likelihood link.}
The MaxEnt optimum satisfies the Boltzmann form
$\pi^\star(\mathbf{x}_0\mid \mathbf{c}) \propto \exp\{\frac{1}{\beta} R(\mathbf{c},\mathbf{x}_0)\}$.
Equivalently, for any candidate $x_0$,
\begin{equation}
\label{eq:reward-lik}
R(\mathbf{x}_0 \mid \mathbf{x}_t,c) = \beta\,\log p(\mathbf{x}_0 \mid \mathbf{x}_t,c) - Z(\mathbf{x}_t,c),
\end{equation}
where $Z$ is a normalizer independent of $x$ (but possibly dependent on $(\mathbf{x}_t,c)$).
Substituting~\eqref{eq:reward-lik} into~\eqref{eq:bt-grad} yields

\begin{equation}
\begin{aligned}
\label{eq:pref-loglik}
&\nabla_{\mathbf{x}_t}\log\widehat{p}(x \mid \mathbf{x}_t,c)
= \\
&w\,\beta\left[
\nabla_{\mathbf{x}_t}\log p(x^w \mid \mathbf{x}_t,c) 
- \nabla_{\mathbf{x}_t}\log p(x^l \mid \mathbf{x}_t,c)
\right].
\end{aligned}
\end{equation}

\paragraph{Bayes expansion to conditional scores.}
By Bayes' rule,
\begin{equation}
\begin{aligned}
\label{eq:bayes}
&\log p(\mathbf{x}_0 \mid \mathbf{x}_t,c)
=\\
&\log p(\mathbf{x}_t \mid \mathbf{x}_0,c)
+ \log p(\mathbf{x}_0 \mid c) - \log p(\mathbf{x}_t \mid c).
\end{aligned}
\end{equation}
Taking $\nabla_{\mathbf{x}_t}$ and subtracting the loser from the winner gives
\begin{equation}
\begin{aligned}
\label{eq:cond-score-diff}
&\nabla_{\mathbf{x}_t}\log p(x^w \mid \mathbf{x}_t,c)
- \nabla_{\mathbf{x}_t}\log p(x^l \mid \mathbf{x}_t,c)
= \\
&\nabla_{\mathbf{x}_t}\log p(\mathbf{x}_t \mid x^w,c)
- \nabla_{\mathbf{x}_t}\log p(\mathbf{x}_t \mid x^l,c),
\end{aligned}
\end{equation}
since the terms $\nabla_{\mathbf{x}_t}\log p(\mathbf{x}_t\mid c)$ cancel.
Combining \eqref{eq:score-decomp}, \eqref{eq:pref-loglik}, and \eqref{eq:cond-score-diff}, we obtain
\begin{equation}
\label{eq:posterior-score-identity}
\begin{aligned}
&\nabla_{\mathbf{x}_t}\log \widehat{p}(\mathbf{x}_t \mid c,y)
=
\nabla_{\mathbf{x}_t}\log p(\mathbf{x}_t \mid c) \\
&\quad+ w\beta\Big[
\nabla_{\mathbf{x}_t}\log p(\mathbf{x}_t \mid x^w,c)
- \nabla_{\mathbf{x}_t}\log p(\mathbf{x}_t \mid x^l,c)
\Big].
\end{aligned}
\end{equation}

\begin{table*}[t]
\centering
\small
\setlength{\tabcolsep}{6pt}
\begin{tabular}{l|cccccc|c}
\toprule
\textbf{Method} & \textbf{PS$\uparrow$} & \textbf{HPSv2$\uparrow$} & \textbf{HPSv3$\uparrow$} & \textbf{Aes$\uparrow$} & \textbf{CLIP$\uparrow$} & \textbf{IR$\uparrow$} & \textbf{Avg.$\uparrow$} \\
\midrule
SDXL      & 50.0 & 50.0 & 50.0 & 50.0 & 50.0 & 50.0 & 50.0 \\
NPO  & 58.7 & 59.2 & 69.1 & 52.1 & 37.5 & 53.5 & 55.0 \\
MaPO & 55.9 & 65.3 & 61.8 & \textbf{68.2} & 50.2 & 68.2 & 61.6 \\
DPO  & 71.7 & 77.6 & 67.9 & 53.3 & 61.6 & 65.8 & 66.3 \\
\rowcolor{blue!15}
PGD-merge & \textbf{87.0} & \textbf{88.9} & \textbf{84.4} & 59.9 & 63.7 & \textbf{77.6} & \textbf{76.9} \\
\rowcolor{blue!15}
cPGD-merge& 76.2 & 84.2 & 72.4 & 45.0 & \textbf{ 68.9} & 71.7 & 69.7 \\
\bottomrule
\end{tabular}
\caption{Win rates(\%) of preference optimization methods against the SDXL model on the Pick-a-Pic v2 test set. Model
checkpoints for other methods are provided by their respective authors. The 1st-best results are bolded.}
\label{tab:merge_winrate}
\end{table*}

\noindent \textbf{Contrastive score decomposition for cPGD.}
\label{prop:cpgd}
Under the BT preference model and the MaxEnt link in \eqref{eq:reward-lik}, the conditional score of the preference posterior satisfies the approximation \eqref{eq:posterior-score-identity}.

\noindent \textbf{Proof sketch.}
Apply the score decomposition \eqref{eq:score-decomp}, approximate the preference-likelihood gradient via \eqref{eq:bt-grad}, invoke the MaxEnt reward--likelihood link \eqref{eq:reward-lik}, and use the Bayes expansion \eqref{eq:bayes} to turn gradients of $\log p(x \mid \mathbf{x}_t,c)$ into a \emph{difference of conditional scores} \eqref{eq:cond-score-diff}. Substituting into \eqref{eq:score-decomp} yields \eqref{eq:posterior-score-identity}.

\paragraph{Practical cPGD estimator.}
We approximate each score in \eqref{eq:posterior-score-identity} with the score of a learned diffusion model:
\begin{equation}
\begin{aligned}
\label{eq:cpgd-estimator}
&\nabla_{\mathbf{x}_t}\log \widehat{p}(\mathbf{x}_t \mid c,y)
=
\underbrace{\nabla_{\mathbf{x}_t}\log p_{\theta_0}(\mathbf{x}_t \mid c)}_{\text{base score}}
+ \\
&\hat{w}\,\Big[
\underbrace{\nabla_{\mathbf{x}_t}\log p_{\theta^w}(\mathbf{x}_t \mid c)}_{\text{positive score}}
-
\underbrace{\nabla_{\mathbf{x}_t}\log p_{\theta^l}(\mathbf{x}_t \mid c)}_{\text{negative score}}
\Big],
\end{aligned}
\end{equation}
where $\theta_0$ is the base model trained on the generic corpus, while $\theta^w$ and $\theta^l$ are separately fine-tuned on winners and losers, respectively (with the standard diffusion objective under condition $c$). The scalar $\hat{w} =  w\beta$ absorbs scale and step-size factors and may be step-dependent, i.e., $\hat{w}_t$.

\paragraph{Relation to PGD and CFG.}
If we tie $\theta^l\!=\!\theta_0$, \eqref{eq:cpgd-estimator} reduces to a PGD-style \emph{single-branch} guidance (base plus a ``preferred'' branch). If, further, we replace $(\theta^w,\theta^l)$ with $(\theta_{cond},\theta_{uncond})$ and tune $\hat{w}$, we recover the functional form of classifier-free guidance (CFG), while cPGD keeps the \emph{pairwise preference semantics} in the guidance difference.

\paragraph{Assumptions and approximations.}
\begin{itemize}
\item \textbf{BT model.} Pairwise preferences follow a sigmoid link on reward gaps; the local sigmoid factor is absorbed into $w$.
\item \textbf{Score surrogates.} The conditional scores $\nabla_{\mathbf{x}_t}\log p(\mathbf{x}_t \mid x^{(\cdot)},c)$ are approximated by diffusion models trained on the corresponding subsets (winners/losers) with condition $c$.
\item \textbf{Scaling.} $\hat{w}$ absorbs $\beta$ and numerical factors from the sampler; tuning can be metric-specific but is typically stable over a range.
\end{itemize}

\section{Collapsing Multi-Model Guidance into a Single Checkpoint via Taylor Linearization}
\label{sec:model_merging}

While PGD and cPGD apply preference guidance at inference time by combining multiple models, we find that their effect can often be approximated by a \emph{single} checkpoint obtained via simple weight-space interpolation. This provides a practical alternative that preserves most preference gains while eliminating the multi-model sampling overhead.

Let $f_{\theta}(x_t, c)$ denote the diffusion model output used by the sampler (e.g., noise or velocity prediction, equivalently the score up to a known transform), parameterized by $\theta$, at time step $t$ with condition $c$.
Denote the base SDXL parameters by $\theta_0$ and a preference-optimized ``winner'' model by $\theta_{+}$.
The PGD-guided prediction can be written as
\begin{equation}
    f_{\text{PGD}}
    \;=\;
    f_{\theta_0}
    \;+\;
    \lambda \bigl(f_{\theta_{+}} - f_{\theta_0}\bigr),
\end{equation}
where $\lambda$ is the PGD guidance weight.

\paragraph{First-order Taylor approximation.}
Consider a first-order Taylor expansion of $f_{\theta}$ around $\theta_0$:
\begin{equation}
    f_{\theta_0 + \Delta\theta}
    \;\approx\;
    f_{\theta_0}
    \;+\;
    J_{\theta_0}\Delta\theta,
    \quad
    J_{\theta_0}
    =
    \left.\frac{\partial f_{\theta}}{\partial \theta}\right|_{\theta=\theta_0}.
\end{equation}
Applying this to $\theta_{+}$ gives
\begin{equation}
    f_{\theta_{+}} - f_{\theta_0}
    \;\approx\;
    J_{\theta_0}(\theta_{+}-\theta_0),
\end{equation}
and substituting back into the PGD update yields
\begin{equation}
    f_{\text{PGD}}
    \;\approx\;
    f_{\theta_0}
    +
    J_{\theta_0}\bigl(\lambda(\theta_{+}-\theta_0)\bigr)
    \;\approx\;
    f_{\theta_0+\lambda(\theta_{+}-\theta_0)}.
\end{equation}
This motivates a merged checkpoint that approximates PGD using a single model,
\begin{equation}
    \theta_{\text{PGD-merge}}
    \;=\;
    \theta_0 + \alpha(\theta_{+}-\theta_0),
\end{equation}
where $\alpha$ is a scalar merging coefficient.

\paragraph{Second-order error under strong (extrapolative) guidance.}
In our setting, the PGD guidance weight is typically chosen from a wide range $\lambda \in [0,10]$, with best performance often near $\lambda \text{ in } [5, 10]$. This places PGD in a \emph{strong, extrapolative} regime ($\lambda>1$), where higher-order deviations from the local linear model can be amplified.

A second-order Taylor expansion with remainder gives
\begin{equation}
    f_{\theta_0+\Delta\theta}
    \;=\;
    f_{\theta_0}
    \;+\;
    J_{\theta_0}\Delta\theta
    \;+\;
    \frac{1}{2}\, \mathcal{H}_{\theta_0}[\Delta\theta,\Delta\theta]
    \;+\;
    \mathcal{O}(\|\Delta\theta\|^3),
    \label{eq:taylor2}
\end{equation}
where $\mathcal{H}_{\theta_0}[\Delta\theta,\Delta\theta]$ denotes the \emph{directional} second derivative of $f_\theta$ at $\theta_0$ along $\Delta\theta$ (a vector in the same output space as $f$).
Let $\Delta\theta = \theta_{+}-\theta_0$, $a := J_{\theta_0}\Delta\theta$, and $b := \mathcal{H}_{\theta_0}[\Delta\theta,\Delta\theta]$. Then
\begin{align}
    f_{\theta_{+}} - f_{\theta_0}
    &=
    a + \frac{1}{2} b + \mathcal{O}(\|\Delta\theta\|^3), \\
    f_{\text{PGD}}
    &=
    f_{\theta_0}
    + \lambda a
    + \frac{1}{2}\lambda b
    + \mathcal{O}(\lambda\|\Delta\theta\|^3),
    \label{eq:pgd_2nd}
\end{align}
whereas evaluating the merged checkpoint at $\theta_0+\alpha\Delta\theta$ yields
\begin{equation}
    f_{\theta_0+\alpha\Delta\theta}
    =
    f_{\theta_0}
    + \alpha a
    + \frac{1}{2}\alpha^2 b
    + \mathcal{O}(\alpha^3\|\Delta\theta\|^3).
    \label{eq:merge_2nd}
\end{equation}
The leading-order mismatch is therefore
\begin{equation}
    f_{\text{PGD}} - f_{\theta_0+\alpha\Delta\theta}
    \;\approx\;
    (\lambda-\alpha)\,a
    \;+\;
    \frac{1}{2}(\lambda-\alpha^2)\,b,
    \label{eq:mismatch_2nd}
\end{equation}
up to higher-order terms.
Importantly, even setting $\alpha=\lambda$ does \emph{not} eliminate the curvature-induced mismatch:
\begin{equation}
    f_{\text{PGD}} - f_{\theta_0+\lambda\Delta\theta}
    \;\approx\;
    \frac{1}{2}\lambda(1-\lambda)\,b,
    \label{eq:curv_residual}
\end{equation}
whose magnitude grows as $\mathcal{O}(\lambda^2)$ in the extrapolative regime $\lambda>1$.
Thus, when $\lambda$ is large (e.g., $\lambda\approx 10$), higher-order effects can dominate and the best-performing merged checkpoint is obtained by treating $\alpha$ as an \emph{effective scale} that absorbs these amplified nonlinearities under the sampling distribution of $(x_t,t,c)$, rather than enforcing $\alpha=\lambda$.

\paragraph{Extension to cPGD.}
For cPGD, we additionally use a ``loser'' model with parameters $\theta_{-}$, and the guided prediction takes the form
\begin{equation}
    f_{\text{cPGD}}
    \;=\;
    f_{\theta_0}
    \;+\;
    \lambda \bigl(f_{\theta_{+}} - f_{\theta_{-}}\bigr).
\end{equation}
Applying the same linearization around $\theta_0$ gives
\begin{equation}
    f_{\theta_{+}} - f_{\theta_{-}}
    \;\approx\;
    J_{\theta_0}(\theta_{+}-\theta_{-}),
    \quad
    \Rightarrow\quad
    f_{\text{cPGD}}
    \;\approx\;
    f_{\theta_0+\lambda(\theta_{+}-\theta_{-})},
\end{equation}
which motivates the merged checkpoint
\begin{equation}
    \theta_{\text{cPGD-merge}}
    \;=\;
    \theta_0 + \alpha(\theta_{+}-\theta_{-}).
\end{equation}
Analogous to \eqref{eq:mismatch_2nd}, a second-order expansion introduces curvature terms of the form
$\mathcal{H}_{\theta_0}[\theta_{+}-\theta_0,\theta_{+}-\theta_0]$ and
$\mathcal{H}_{\theta_0}[\theta_{-}-\theta_0,\theta_{-}-\theta_0]$,
so $\alpha$ again serves as a practical knob that absorbs higher-order deviations from the local linear model.

\paragraph{Merging coefficients and evaluation.}
We perform a lightweight sweep over $\alpha$ and fix the best-performing values throughout our experiments:
\textbf{$\alpha = 6$ for PGD-merge} and \textbf{$\alpha = 10$ for cPGD-merge}.
Once merged, $\theta_{\text{PGD-merge}}$ and $\theta_{\text{cPGD-merge}}$ are used as standard SDXL-style generators without any additional guidance model.
Table~\ref{tab:merge_winrate} reports win rates (\%) of all methods against SDXL on the Pick-a-Pic v2 test set.
Model checkpoints for NPO, MaPO, and DPO are taken from the official releases.

\paragraph{Results and analysis.}
Compared to the SDXL reference row (fixed at $50\%$ by construction), all preference-optimized baselines improve the average win rate, with DPO reaching $66.3\%$.
PGD-merge further boosts the average win rate to \textbf{$76.9\%$}, a gain of more than $10$ percentage points over DPO.
It also attains the highest scores on most individual metrics, including PS ($87.0\%$), HPSv2 ($88.9\%$), HPSv3 ($84.4\%$), and IR ($77.6\%$), indicating strong overall preference alignment and instruction-following.
cPGD-merge achieves a slightly lower average win rate of \textbf{$69.7\%$}, but remains clearly above all prior baselines and exhibits the best CLIP win rate ($68.9\%$), suggesting improved text--image consistency.
Overall, the two merged models dominate or match existing preference-optimization methods on most metrics, showing that even a simple first-order weight-space merging scheme can recover most of the benefits of explicit PGD/cPGD guidance, while collapsing the multi-model inference pipeline into a single checkpoint.


\section{Plug-and-play across Different Architectures}

\begin{figure}[ht]
\centering
\includegraphics[width=\linewidth]{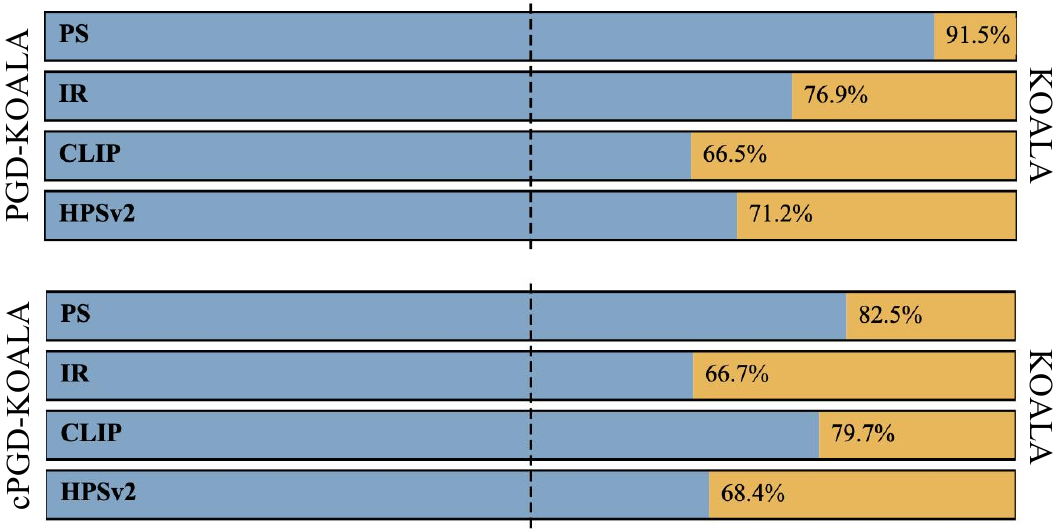}
 \caption{\textbf{Plug-and-play transfer to \textsc{KOALA} using an \textsc{SDXL} PGD/cPGD score module.}
    Pairwise preference rates (\%) of \textsc{PGD-\!KOALA} (top) and \textsc{cPGD-\!KOALA} (bottom) against the original \textsc{KOALA} baseline under four evaluation metrics (PS, IR, CLIP, HPSv2). The dashed vertical line denotes parity at 50\%; values above 50\% indicate that the guided model is preferred. All gains are obtained \emph{without} any training or fine-tuning of \textsc{KOALA}, highlighting the plug-and-play nature of PGD/cPGD under shared latent-space (VAE) alignment.}
    \label{fig:KOALA}
\end{figure}

To assess whether PGD/cPGD can be deployed in a truly plug-and-play manner across model architectures, we apply our framework to \textsc{KOALA}~\citep{koala}, a text-to-image model whose parameterization differs from \textsc{SDXL}. Concretely, we reuse the \textsc{SDXL} score module as an external guidance component during \textsc{KOALA} sampling. This transfer is enabled by a key compatibility condition: \textsc{KOALA} and \textsc{SDXL} share the same VAE and thus operate in an identical latent space. Such latent-space alignment is essential for our method, since PGD/cPGD is score-based and relies on consistent latent representations to inject guidance without modifying the target model. As shown in Fig.~\ref{fig:KOALA}, integrating the \textsc{SDXL} score module improves \textsc{KOALA} consistently across PS, IR, CLIP, and HPSv2, while requiring \emph{no} additional training or fine-tuning on \textsc{KOALA}. These results demonstrate that our approach generalizes effectively across architectures when the latent space is shared.

\section{Human Preference Study}
\label{sec:human-pref}

\paragraph{Setup and protocol.}
We conduct a human preference evaluation using a bilingual (English/Chinese) web interface; instructions are shown in both languages with English taking precedence. For each question, we display six images generated from the \emph{same} text prompt by six systems (Raw, DPO, MaPO, NPO, PGD, cPGD). Images are \emph{anonymized} and labeled only by random letters. To mitigate order and identity biases, (i) the six thumbnails are randomly permuted per question, and (ii) the letter--method mapping is re-sampled for every question (hence letters are not consistent across questions). Participants are instructed to \textbf{select 1--3 images} they prefer according to the stated criteria:
\emph{Text Alignment} (faithfulness to the prompt text), \emph{Semantic Alignment} (scene/logic consistency), and \emph{Aesthetic Preference} (composition, lighting, style).

\paragraph{Subjects and materials.}
We evaluate \num{55} prompts with \num{20} participants, yielding \num{55}\,$\times$\,\num{20}\,=\,\num{1100} prompt--participant decision units. Because multiple selections are allowed, the total number of recorded choices (“votes”) exceeds the number of questions: we collect \num{1848} valid votes in total (\(\approx\)1.68 selections per decision unit; \(\approx\)33.6 selections per prompt).

\paragraph{Aggregation metric.}
Each checked option contributes one vote to the corresponding method. We report the \emph{vote share} of each method, defined as its total votes divided by the grand total of votes across all methods.

\begin{table}[ht]
\centering
\small
\begin{tabular}{lcc}
\toprule
Method & Votes & Vote share (\%) \\
\midrule
Raw   & 208 & 18.9 \\
DPO   & 324 & 29.5 \\
MaPO  & 197 & 17.9 \\
NPO   & 256 & 23.3 \\
PGD   & \textbf{500} & \textbf{45.5} \\
cPGD  & \uline{363} & \uline{33.0} \\
\midrule
Total & 1848 & 168.0 \\
\bottomrule
\end{tabular}
\caption{\textbf{Summary of human preference votes.} Participants could select 1--3 images per question, hence the total number of votes exceeds the number of questions.}
\label{tab:human-pref}
\end{table}

\paragraph{Results.}
As shown in Table~\ref{tab:human-pref}, PGD receives 500 votes, corresponding to a 45.5\% selection rate, which is a relative improvement of +54.2\% over DPO (324 votes, 29.5\%). cPGD obtains 363 votes (33.0\%), an +8.2\% relative gain over DPO. Raw and MaPO are selected in 208 (18.9\%) and 197 (17.9\%) cases, respectively, while NPO receives 256 votes (23.3\%). Because participants could select 1--3 images per question, the percentages in Table~\ref{tab:human-pref} sum to 168\%, indicating that annotators chose on average 1.68 images per question.

It is worth noting that this multiple-selection protocol, while largely fair, may slightly exaggerate the apparent margin of PGD and attenuate the observed advantage of cPGD over DPO. When several plausible images are presented, annotators tend to favor the option that is visually or stylistically most distinctive, rather than the one that is strictly the best aligned with the text prompt. This behavior is consistent with salience bias and the von Restorff (isolation) effect: annotators tend to overweight visually salient styles, which may slightly exaggerate PGD’s margin and understate cPGD’s advantage over DPO under the multi-choice protocol.

\section{Additional Ablation Studies}
\label{sec:more_ablation}

\begin{figure}[ht]
\centering
\includegraphics[width=\linewidth]{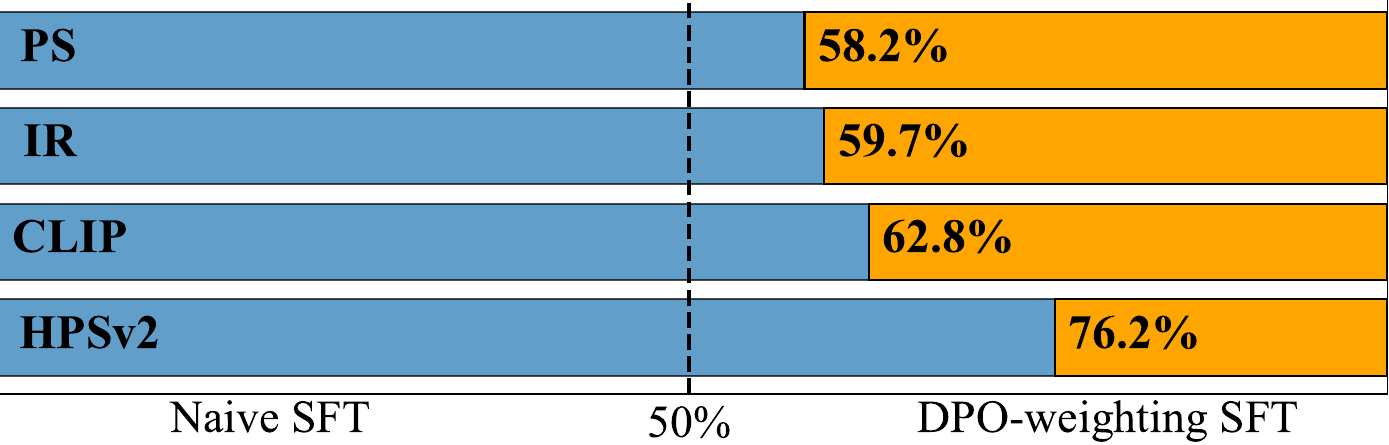}
\caption{Comparison of SFT training with and without $w_{\mathrm{DPO}}$ weighting. 
Blue bars indicate naive SFT, and orange bars indicate weighted SFT. 
All results are obtained under the cPGD inference strategy with the same guidance weight.}
\label{fig:lossWeight}
\end{figure}

\noindent \textbf{Effect of $w_{\mathrm{DPO}}$. weighting For cPGD} As shown in Fig.~\ref{fig:lossWeight}, re-weighting the SFT objective with $w_{\mathrm{DPO}}$ consistently lifts win rates above the 50\% chance level under the same cPGD guidance weight—58.2\% on PickScore, 59.7\% on ImageReward, 62.8\% on CLIP, and 76.2\% on HPSv2. In contrast, na\"ive SFT hovers around 50\%, indicating that uniform supervision underutilizes the signal in preference data. The DPO-derived weights emphasize examples with larger preference margins, aligning the SFT update with the “preference direction” exploited at inference and yielding uniform gains without changing the sampler, prompts, or seeds.

\paragraph{Partial-step guidance: compute-quality trade-off.}

\noindent\textbf{Setup.}
Under \emph{cPGD}, we keep the default SDXL sampler at 50 diffusion steps and
apply guidance only to the first $s$ high-noise steps, falling back to the base
update afterwards:
\[
w_t \;=\;
\begin{cases}
w, & t \le s,\\
0, & t > s,
\end{cases}
\qquad t=1,\dots,50.
\]
We report PickScore win rate (vs.\ SDXL-base, higher is better) and the
\emph{time ratio} measured as wall-clock relative to vanilla SDXL 50-step
sampling (i.e., $1.0$ equals unguided SDXL).

\begin{table}[ht]
\centering
\caption{Guiding only the first $s$ of 50 steps on SDXL (cPGD).}
\label{tab:partial-steps-transposed}
\setlength{\tabcolsep}{10pt}
\resizebox{\linewidth}{!}{
\begin{tabular}{c|cc}
\toprule
\textbf{Guided steps $s$} &
\makecell{\textbf{PickScore}\\\textbf{win rate (\%)}} &
\makecell{\textbf{Time ratio}\\\textbf{($\times$ vs.\ SDXL)}} \\
\midrule
10 & 70.2 & 1.4 \\
20 & 72.4 & 1.8 \\
30 & 74.8 & 2.2 \\
40 & 76.9 & 2.6 \\
50 & 79.1 & 3.0 \\
\bottomrule
\end{tabular}
}
\label{tab:partial-steps}
\end{table}

\noindent\textbf{Findings.}
Table~\ref{tab:partial-steps} exhibits a smooth Pareto frontier:
as $s$ increases, reward improves monotonically while compute grows roughly
linearly. Notably, \emph{even guiding only the first 10 steps} already delivers
a strong gain (70.2\% win rate) at just $1.4\times$ the cost of vanilla SDXL.
Moreover, $s{=}30$ recovers about $94.6\%$ of the full 50-step cPGD reward
(74.8 vs.\ 79.1) at only $\sim\!73\%$ of its compute (2.2 vs.\ 3.0),
and $s{=}40$ reaches $97.2\%$ of full reward at $\sim\!87\%$ of the compute.
In practice, selecting $s$ around 30--40 provides a favorable balance between
preference adherence and efficiency, while $s{=}10$ is a compelling low-cost
setting when latency is critical.

\noindent\textbf{Distillation.} To improve efficiency, we compress our guidance into a single checkpoint via a simple distillation procedure. Concretely, on Pick-a-Pic v2 we train for 500 steps using the standard $\epsilon$-prediction loss
\begin{equation*}
\mathcal{L}_{\text{distill}}
= \mathbb{E}_{(x_0,c),\,t,\,\epsilon}\!
\left[\;\big\|\;\hat{\epsilon}
- \epsilon_\phi(x_0,\epsilon,t,c)\;\big\|_2^2\right],
\end{equation*}
where $\hat{\epsilon}$ is obtained from the cPGD guidance formula. We then evaluate the distilled cPGD checkpoint against the raw SDXL base model and a DPO-tuned SDXL on three preference proxies (PickScore, HPSv3, and ImageReward). Figure~\ref{fig:distill_bar} reports \emph{win rates} (\%), measured as the fraction of prompts for which a method outperforms the SDXL base under identical seeds. The distilled model consistently surpasses DPO on PickScore and HPSv3 while remaining competitive on ImageReward, demonstrating that the benefits of preference guidance can be preserved even when compressed into a single offline checkpoint.

In our implementation, we use a purely offline distillation protocol: the targets $\hat{\epsilon}$ are computed once on the original Pick-a-Pic v2 pairs $(x_0,c)$, without generating new samples.
This makes distillation fast and simple, and reduces dependence on external factors such as prompt quality or sampling heuristics.
In principle, higher-quality (or more curated) preference data and online variants that interleave sampling and distillation could further improve the fidelity of the distilled checkpoint.

\begin{figure}[t]
\vspace{1mm}
\centering
\vspace{-1pt}   

\resizebox{\linewidth}{!}{
\begin{tabular}{@{}c c@{}}
  \includegraphics[width=0.499\linewidth]{ 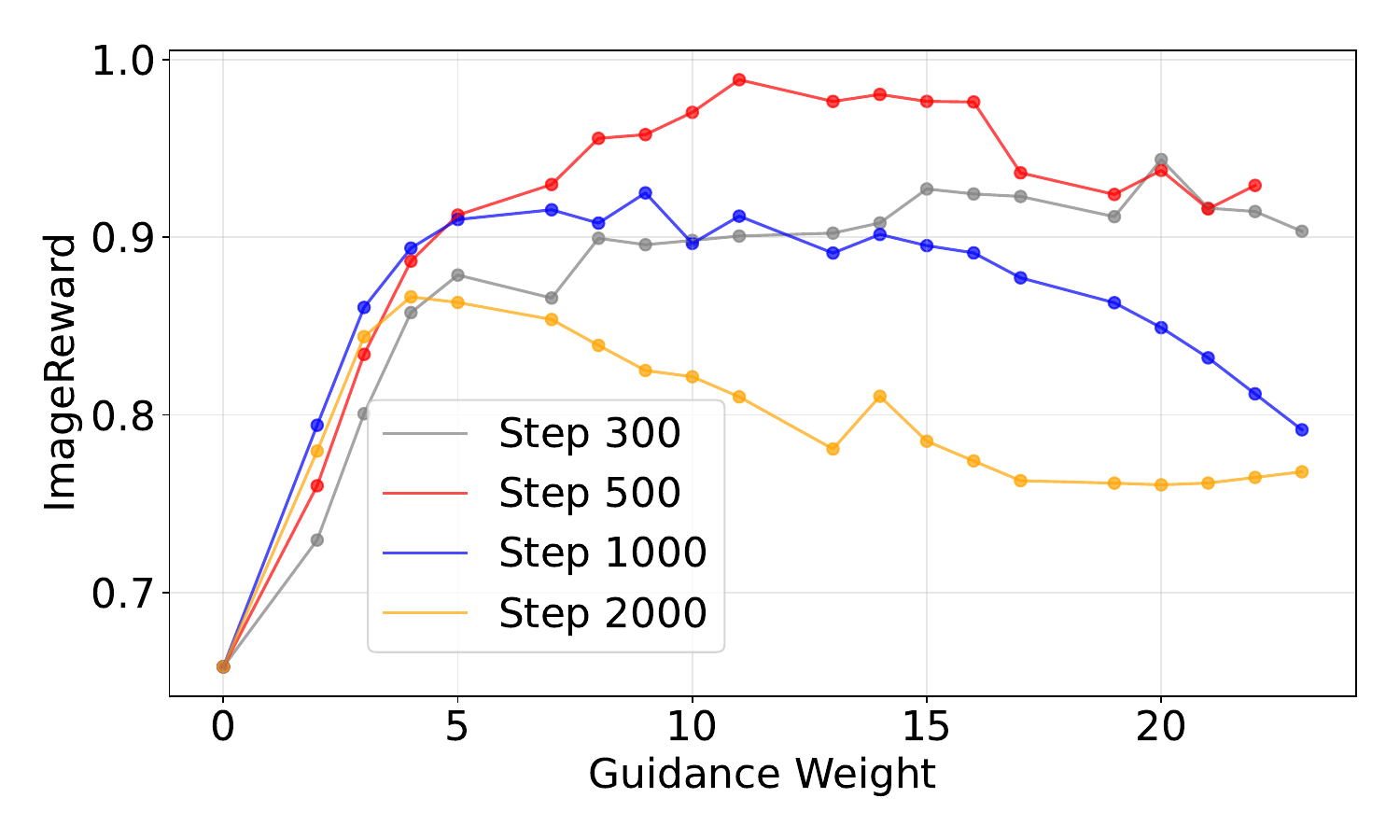} &
  \includegraphics[width=0.499\linewidth]{ 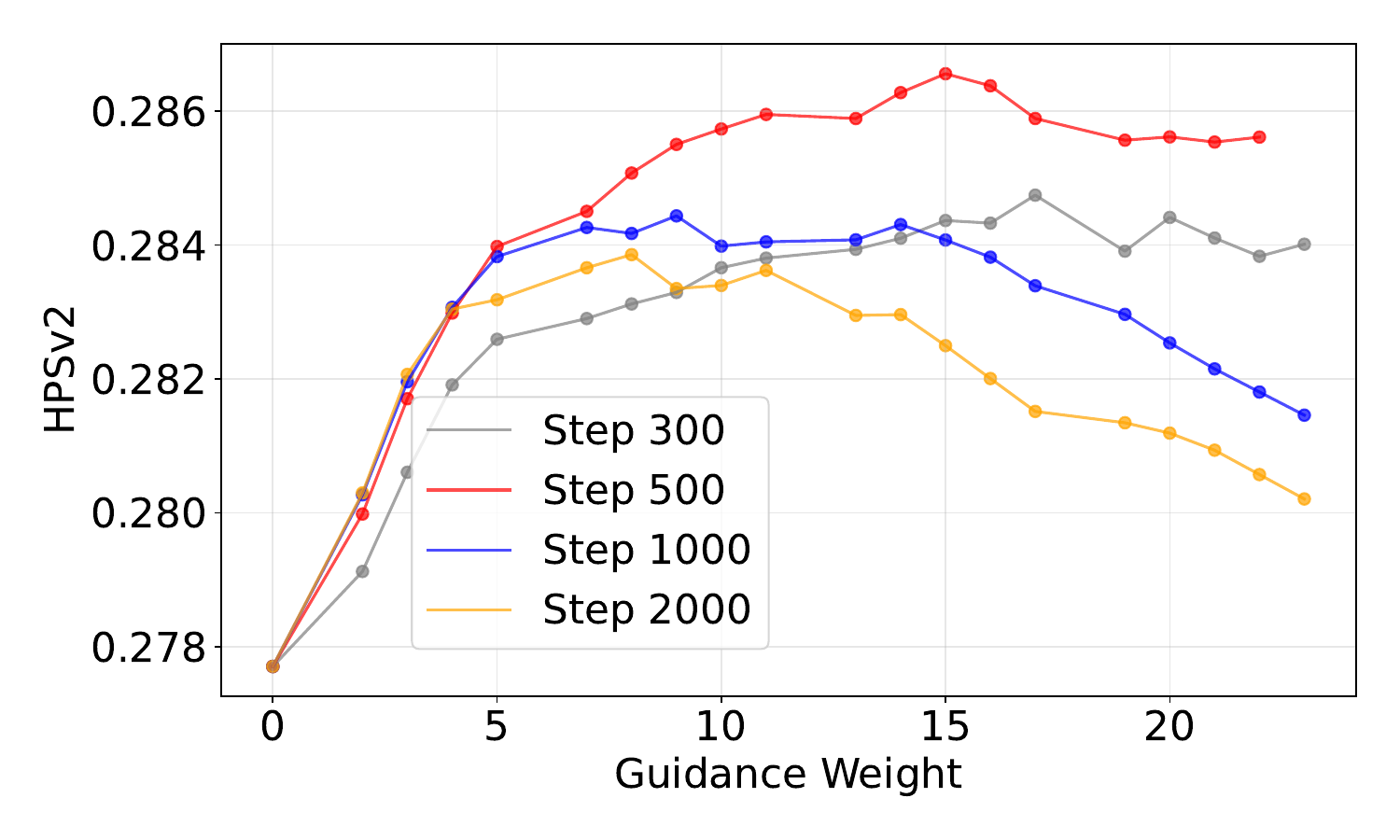} \\
\end{tabular}
}
\captionsetup{labelfont=footnotesize}
\caption{
\footnotesize 
Effect of guidance weight $w$ on automatic metrics (SDXL). 
Left: ImageReward; Right: HPSv2. 
“Step” denotes the training steps of the guidance module. 
Curves rise quickly for small $w$, 
}
\label{fig:step-weight-v2}
\end{figure}

\noindent\textbf{More metrics on guidance weights.}
As shown in Figures~\ref{fig:step-weight} and~\ref{fig:step-weight-v2}, the curves under different preference metrics exhibit very similar trends as we vary the guidance weight $w$.
This indicates that severe reward hacking is uncommon in our setting: improvements on one proxy metric typically do not come at the expense of others.
It also suggests that tuning $w$ does not incur a large overhead from having to separately fit each metric, since a single choice of $w$ transfers well across all evaluated preference metrics.

\section{Full Quantitative Results}

We report complete per-metric win rates (WR, \%) across three prompt suites for both SDXL and SD~1.5. 
Results on SDXL with Pick-a-Pic~v2, HPDv2, and Parti-Prompts appear in Tables~\ref{tab:sdxl_winrates_1}--\ref{tab:sdxl_winrates_3}; 
SD~1.5 results are given in Tables~\ref{tab:sd15_winrates_1}--\ref{tab:sd15_winrates_3}. 
Across all benchmarks, our guided variants outperform baselines on preference-aligned proxies (PickScore, HPSv2/v3, ImageReward), yielding higher unweighted \emph{average} win rate.

Within each block, the row marked ``--'' is the reference policy evaluated against itself and therefore equals $50\%$ by construction. 
Numbers above $50\%$ indicate systematic wins over the base under identical prompts, sampler settings, and fixed seed. 
The \emph{Average} column is the unweighted mean over PS, HPSv2, HPSv3, Aes, CLIP, and IR.

\noindent\textbf{High-level trends.}
(1) On SDXL, both PGD and cPGD consistently lift the underlying models across all three suites, with larger gains when guiding stronger preference-tuned bases (e.g., DPO/MaPO/SPO). 
(2) On SD~1.5, the same pattern holds: PGD brings uniform improvements across metrics, while cPGD recovers most of the PGD gains with lower compute.
(3) Improvements concentrate on PS/HPS/IR, while Aes/CLIP remain competitive—consistent with guidance emphasizing the learned preference direction rather than style drift.

\begin{table*}[t]
\caption{Win rates of preference optimization methods against the SDXL model on the Pick-a-Pic
v2 test set benchmark. Model checkpoints for other methods are provided by
their respective authors.}
\centering
\resizebox{0.7\textwidth}{!}{
    \begin{tabular}{l c | c c c c c c | c}
    \toprule
    \textbf{Base Model} &{\makecell{Inference \\ Strategy}}  & \textbf{ PS} $\uparrow$ & \textbf{ HPSv2} $\uparrow$ & \textbf{ HPSv3} $\uparrow$ & \textbf{ Aes} $\uparrow$ & \textbf{ CLIP} $\uparrow$ & \textbf{ IR} $\uparrow$  &  \textbf{ Average} $\uparrow$ \\
    \midrule
                            & --    & 50.00\% & 50.00\% & 50.00\% & 50.00\% & 50.00\% & 50.00\% & 50.00\% \\
                            & cPGD & 79.95\% & 80.19\% & 77.12\% & 50.94\% & 64.86\% & 69.81\% & 70.48\% \\
                            & PGD & 78.77\% & 79.01\% & 73.58\% & 51.89\% & 63.68\% & 69.34\% & 69.38\% \\
\multirow{-4}{*}{SDXL} & NPO   & 58.73\% & 59.20\% & 69.10\% & 52.12\% & 37.50\% & 53.54\% & 55.03\% \\ \midrule
                            & --    & 71.70\% & 77.59\% & 67.92\% & 53.30\% & 61.56\% & 65.80\% & 66.31\% \\
                            & cPGD & 80.90\% & 77.59\% & 84.67\% & 63.92\% & 58.73\% & 64.86\% & 71.78\% \\
                            & PGD & 83.25\% & 85.38\% & 85.61\% & 59.67\% & 62.26\% & 73.58\% & 74.96\% \\
\multirow{-4}{*}{DPO-SDXL}  & NPO   & 76.89\% & 81.84\% & 81.37\% & 53.77\% & 57.78\% & 70.75\% & 70.40\% \\ \midrule
                            & --    & 55.90\% & 65.33\% & 61.79\% & 68.16\% & 50.24\% & 68.16\% & 61.60\% \\
                            & cPGD & 77.36\% & 78.77\% & 72.88\% & 69.10\% & 59.43\% & 72.41\% & 71.66\% \\
\multirow{-3}{*}{MaPO-SDXL} & PGD & 80.42\% & 81.60\% & 81.60\% & 75.71\% & 51.42\% & 72.17\% & 73.82\% \\ \midrule
                            & --    & 89.39\% & 83.02\% & 95.99\% & 81.84\% & 33.25\% & 78.77\% & 77.04\% \\
                            & cPGD & 92.92\% & 88.44\% & 96.70\% & 78.77\% & 53.77\% & 84.43\% & 82.51\% \\
\multirow{-3}{*}{SPO-SDXL}  & PGD & 92.22\% & 86.08\% & 96.46\% & 82.08\% & 42.45\% & 81.37\% & 80.11\% \\
    \bottomrule
    \end{tabular}
}
\label{tab:sdxl_winrates_1}
\end{table*}

\begin{table*}[t]
\caption{Win rates of preference optimization methods against the SDXL model on the HPD
v2 test set benchmark. Model checkpoints for other methods are provided by
their respective authors.}
\centering
\resizebox{0.7\textwidth}{!}{
    \begin{tabular}{l c | c c c c c c | c}
    \toprule
    \textbf{Base Model} &{\makecell{Inference \\ Strategy}}  & \textbf{ PS} $\uparrow$ & \textbf{ HPSv2} $\uparrow$ & \textbf{ HPSv3} $\uparrow$ & \textbf{ Aes} $\uparrow$ & \textbf{ CLIP} $\uparrow$ & \textbf{ IR} $\uparrow$  &  \textbf{ Average} $\uparrow$ \\
    \midrule
                            & --    & 50.00\% & 50.00\% & 50.00\% & 50.00\% & 50.00\% & 50.00\% & 50.00\% \\
                            & cPGD & 76.50\% & 75.50\% & 71.25\% & 52.25\% & 59.00\% & 69.00\% & 67.25\% \\
                            & PGD & 85.50\% & 73.50\% & 78.75\% & 69.25\% & 46.75\% & 73.00\% & 71.13\% \\
\multirow{-4}{*}{SDXL} & NPO   & 63.75\% & 65.00\% & 78.75\% & 60.50\% & 42.75\% & 55.50\% & 61.04\% \\ \midrule
                            & --    & 66.75\% & 73.75\% & 66.00\% & 59.75\% & 55.25\% & 73.00\% & 65.75\% \\
                            & cPGD & 75.25\% & 79.50\% & 71.25\% & 57.75\% & 61.00\% & 73.00\% & 69.63\% \\
                            & PGD & 84.25\% & 85.50\% & 82.25\% & 66.75\% & 55.00\% & 75.25\% & 74.83\% \\
\multirow{-4}{*}{DPO-SDXL}  & NPO   & 78.00\% & 86.00\% & 78.00\% & 60.75\% & 53.25\% & 76.25\% & 72.04\% \\ \midrule
                            & --    & 55.75\% & 72.50\% & 68.00\% & 69.25\% & 50.75\% & 65.50\% & 63.63\% \\
                            & cPGD & 73.75\% & 81.25\% & 77.00\% & 67.75\% & 57.00\% & 70.75\% & 71.25\% \\
\multirow{-3}{*}{MaPO-SDXL} & PGD & 86.50\% & 79.25\% & 83.75\% & 74.25\% & 50.25\% & 73.00\% & 74.50\% \\ \midrule
                            & --    & 93.00\% & 90.00\% & 97.25\% & 86.00\% & 32.75\% & 77.25\% & 79.38\% \\
                            & cPGD & 93.25\% & 91.75\% & 97.00\% & 81.25\% & 42.75\% & 81.25\% & 81.21\% \\
\multirow{-3}{*}{SPO-SDXL}  & PGD & 95.00\% & 90.75\% & 96.00\% & 84.50\% & 34.50\% & 77.25\% & 79.67\% \\
    \bottomrule
    \end{tabular}
}
\label{tab:sdxl_winrates_2}
\end{table*}

\begin{table*}[t]
\caption{Win rates of preference optimization methods against the SDXL model on the Parti-Prompts
set benchmark. Model checkpoints for other methods are provided by
their respective authors.}
\centering
\resizebox{0.7\textwidth}{!}{
    \begin{tabular}{l c | c c c c c c | c}
    \toprule
    \textbf{Base Model} &{\makecell{Inference \\ Strategy}}  & \textbf{ PS} $\uparrow$ & \textbf{ HPSv2} $\uparrow$ & \textbf{ HPSv3} $\uparrow$ & \textbf{ Aes} $\uparrow$ & \textbf{ CLIP} $\uparrow$ & \textbf{ IR} $\uparrow$  &  \textbf{ Average} $\uparrow$ \\
    \midrule
                            & --    & 50.00\% & 50.00\% & 50.00\% & 50.00\% & 50.00\% & 50.00\% & 50.00\% \\
                            & cPGD & 75.80\% & 80.27\% & 77.82\% & 57.23\% & 62.01\% & 74.02\% & 71.19\% \\
                            & PGD & 78.74\% & 78.37\% & 78.19\% & 68.75\% & 53.98\% & 69.06\% & 71.18\% \\
\multirow{-4}{*}{SDXL} & NPO   & 55.02\% & 56.37\% & 62.07\% & 55.09\% & 38.97\% & 51.53\% & 53.18\% \\ \midrule
                            & --    & 63.97\% & 70.28\% & 64.03\% & 57.54\% & 58.46\% & 69.73\% & 64.00\% \\
                            & cPGD & 73.90\% & 79.23\% & 72.79\% & 59.38\% & 64.09\% & 74.45\% & 70.64\% \\
                            & PGD & 80.76\% & 83.88\% & 81.74\% & 67.65\% & 57.48\% & 76.10\% & 74.60\% \\
\multirow{-4}{*}{DPO-SDXL}  & NPO   & 70.59\% & 78.37\% & 77.51\% & 63.36\% & 56.25\% & 70.83\% & 69.49\% \\ \midrule
                            & --    & 52.02\% & 64.40\% & 58.52\% & 72.37\% & 48.22\% & 65.01\% & 60.09\% \\
                            & cPGD & 72.49\% & 81.13\% & 76.90\% & 70.10\% & 58.21\% & 74.39\% & 72.20\% \\
\multirow{-3}{*}{MaPO-SDXL} & PGD & 78.86\% & 77.82\% & 79.60\% & 77.76\% & 53.55\% & 72.67\% & 73.38\% \\ \midrule
                            & --    & 87.81\% & 85.48\% & 92.16\% & 88.11\% & 31.74\% & 74.94\% & 76.71\% \\
                            & cPGD & 91.97\% & 90.32\% & 93.87\% & 83.64\% & 50.37\% & 81.31\% & 81.91\% \\
\multirow{-3}{*}{SPO-SDXL}  & PGD & 91.36\% & 87.75\% & 93.87\% & 88.36\% & 47.98\% & 77.33\% & 81.11\% \\
    \bottomrule
    \end{tabular}
}
\label{tab:sdxl_winrates_3}
\end{table*}

\begin{table*}[t]
\caption{Win rates of preference optimization methods against the SD1.5 model on the Pick-a-Pic
v2 test set benchmark. Model checkpoints for other methods are provided by
their respective authors.}
\centering
\resizebox{0.7\textwidth}{!}{
    \begin{tabular}{l c | c c c c c c | c}
    \toprule
    \textbf{Base Model} &{\makecell{Inference \\ Strategy}}  & \textbf{ PS} $\uparrow$ & \textbf{ HPSv2} $\uparrow$ & \textbf{ HPSv3} $\uparrow$ & \textbf{ Aes} $\uparrow$ & \textbf{ CLIP} $\uparrow$ & \textbf{ IR} $\uparrow$  &  \textbf{ Average} $\uparrow$ \\
    \midrule
                          & --    & 50.00\% & 50.00\% & 50.00\% & 50.00\% & 50.00\% & 50.00\% & 50.00\% \\
                           & cPGD & 76.89\% & 71.70\% & 71.93\% & 63.21\% & 59.91\% & 72.17\% & 69.30\% \\
\multirow{-3}{*}{SD1.5}     & PGD & 78.30\% & 71.23\% & 67.92\% & 62.26\% & 58.49\% & 63.68\% & 66.98\% \\ \midrule
                           & --    & 76.42\% & 67.69\% & 66.27\% & 65.09\% & 55.90\% & 60.61\% & 65.33\% \\
                           & cPGD & 79.01\% & 81.84\% & 75.47\% & 71.70\% & 62.26\% & 76.18\% & 74.41\% \\
\multirow{-3}{*}{DPO-SD1.5} & PGD & 79.25\% & 70.28\% & 66.27\% & 66.27\% & 59.43\% & 63.21\% & 67.45\% \\ \midrule
                           & --    & 72.64\% & 78.07\% & 76.18\% & 68.63\% & 58.49\% & 75.00\% & 71.50\% \\
                           & cPGD & 76.65\% & 80.19\% & 75.94\% & 70.51\% & 60.38\% & 74.29\% & 72.99\% \\
\multirow{-3}{*}{KTO-SD1.5} & PGD & 81.60\% & 83.25\% & 80.19\% & 70.28\% & 61.08\% & 77.36\% & 75.63\% \\ \midrule
                           & --    & 71.23\% & 62.97\% & 64.86\% & 68.16\% & 38.68\% & 61.08\% & 61.16\% \\
                           & cPGD & 82.31\% & 81.84\% & 75.47\% & 71.23\% & 60.61\% & 76.18\% & 74.61\% \\
\multirow{-3}{*}{SPO-SD1.5} & PGD & 79.72\% & 70.28\% & 69.58\% & 69.81\% & 44.34\% & 67.92\% & 66.94\% \\
    \bottomrule
    \end{tabular}
}
\label{tab:sd15_winrates_1}
\end{table*}

\begin{table*}[t]
\caption{Win rates of preference optimization methods against the SD1.5 model on the HPD
v2 test set benchmark. Model checkpoints for other methods are provided by
their respective authors.}
\centering
\resizebox{0.7\textwidth}{!}{
    \begin{tabular}{l c | c c c c c c | c}
    \toprule
    \textbf{Base Model} &{\makecell{Inference \\ Strategy}}  & \textbf{ PS} $\uparrow$ & \textbf{ HPSv2} $\uparrow$ & \textbf{ HPSv3} $\uparrow$ & \textbf{ Aes} $\uparrow$ & \textbf{ CLIP} $\uparrow$ & \textbf{ IR} $\uparrow$  &  \textbf{ Average} $\uparrow$ \\
    \midrule
                          & --    & 50.00\% & 50.00\% & 50.00\% & 50.00\% & 50.00\% & 50.00\% & 50.00\% \\
                           & cPGD & 80.50\% & 88.00\% & 84.00\% & 73.00\% & 61.25\% & 79.50\% & 77.71\% \\
\multirow{-3}{*}{SD1.5}     & PGD & 75.25\% & 68.00\% & 67.00\% & 67.25\% & 52.25\% & 65.50\% & 65.88\% \\ \midrule
                           & --    & 76.25\% & 69.75\% & 68.00\% & 66.50\% & 57.50\% & 65.25\% & 67.21\% \\
                           & cPGD & 80.25\% & 81.00\% & 79.00\% & 70.25\% & 60.75\% & 75.75\% & 74.50\% \\
\multirow{-3}{*}{DPO-SD1.5} & PGD & 79.00\% & 73.75\% & 71.75\% & 66.00\% & 54.75\% & 68.50\% & 68.96\% \\ \midrule
                           & --    & 73.50\% & 85.00\% & 85.00\% & 69.50\% & 58.75\% & 77.75\% & 74.92\% \\
                           & cPGD & 77.00\% & 86.50\% & 82.25\% & 68.00\% & 58.00\% & 79.50\% & 75.21\% \\
\multirow{-3}{*}{KTO-SD1.5} & PGD & 80.50\% & 88.50\% & 84.75\% & 71.75\% & 58.75\% & 80.50\% & 77.46\% \\ \midrule
                           & --    & 77.25\% & 72.75\% & 74.00\% & 73.75\% & 29.25\% & 63.25\% & 65.04\% \\
                           & cPGD & 80.75\% & 78.75\% & 82.25\% & 71.25\% & 44.50\% & 75.00\% & 72.08\% \\
\multirow{-3}{*}{SPO-SD1.5} & PGD & 81.75\% & 74.25\% & 75.00\% & 75.25\% & 35.00\% & 67.00\% & 68.04\% \\
    \bottomrule
    \end{tabular}
}
\label{tab:sd15_winrates_2}
\end{table*}

\begin{table*}[t]
\caption{Win rates of preference optimization methods against the SD1.5 model on the Parti-Prompts
set benchmark. Model checkpoints for other methods are provided by
their respective authors.}
\centering
\resizebox{0.7\textwidth}{!}{
    \begin{tabular}{l c | c c c c c c | c}
    \toprule
    \textbf{Base Model} &{\makecell{Inference \\ Strategy}}  & \textbf{ PS} $\uparrow$ & \textbf{ HPSv2} $\uparrow$ & \textbf{ HPSv3} $\uparrow$ & \textbf{ Aes} $\uparrow$ & \textbf{ CLIP} $\uparrow$ & \textbf{ IR} $\uparrow$  &  \textbf{ Average} $\uparrow$ \\
    \midrule
                            & --    & 50.00\% & 50.00\% & 50.00\% & 50.00\% & 50.00\% & 50.00\% & 50.00\% \\
                           & cPGD & 66.36\% & 76.90\% & 68.87\% & 68.14\% & 58.39\% & 71.02\% & 68.28\% \\
\multirow{-3}{*}{SD1.5}     & PGD & 68.01\% & 65.01\% & 59.56\% & 58.15\% & 54.96\% & 56.86\% & 60.43\% \\ \midrule
                           & --    & 67.34\% & 64.83\% & 64.46\% & 62.19\% & 53.55\% & 61.03\% & 62.23\% \\
                           & cPGD & 73.84\% & 74.08\% & 69.61\% & 69.00\% & 59.01\% & 67.34\% & 68.81\% \\
\multirow{-3}{*}{DPO-SD1.5} & PGD & 74.20\% & 67.40\% & 65.01\% & 63.05\% & 55.51\% & 62.87\% & 64.68\% \\ \midrule
                           & --    & 66.61\% & 78.31\% & 71.94\% & 68.75\% & 53.31\% & 71.26\% & 68.36\% \\
                           & cPGD & 66.24\% & 77.14\% & 69.49\% & 68.44\% & 55.88\% & 72.18\% & 68.23\% \\
\multirow{-3}{*}{KTO-SD1.5} & PGD & 72.12\% & 80.33\% & 72.79\% & 72.43\% & 55.15\% & 73.84\% & 71.11\% \\ \midrule
                           & --    & 68.63\% & 61.21\% & 64.15\% & 71.94\% & 37.75\% & 61.64\% & 60.89\% \\
                           & cPGD & 74.82\% & 71.94\% & 71.32\% & 73.90\% & 47.49\% & 72.86\% & 68.72\% \\
\multirow{-3}{*}{SPO-SD1.5} & PGD & 74.20\% & 66.54\% & 66.24\% & 72.24\% & 46.69\% & 67.03\% & 65.49\% \\
    \bottomrule
    \end{tabular}
}
\label{tab:sd15_winrates_3}
\end{table*}

\begin{figure*}[t]
\centering
\begin{tabular}{c c c c}
 \multicolumn{4}{c}{\includegraphics[width=0.7\linewidth]{ 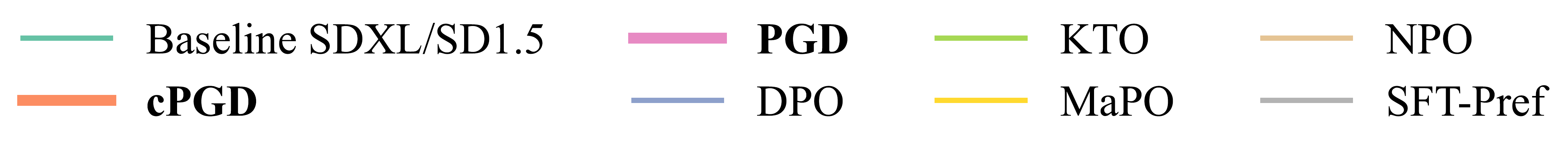}} \\[1ex]
  
  \parbox[t]{0.03\linewidth}{\centering \raisebox{10ex}{\rotatebox{90}{SDXL}}
} &
  \includegraphics[width=0.29\linewidth]{ 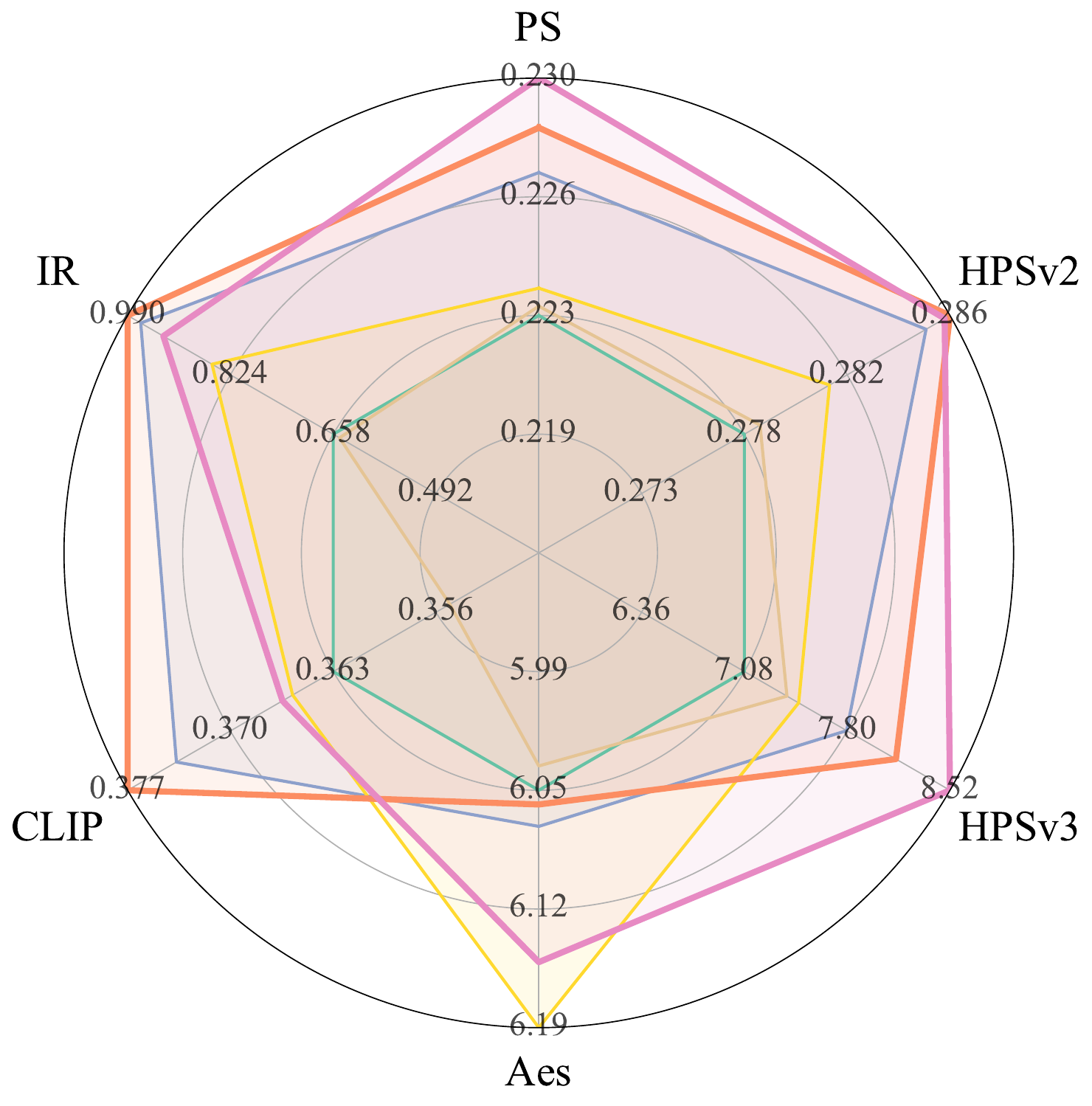} &
  \includegraphics[width=0.29\linewidth]{ 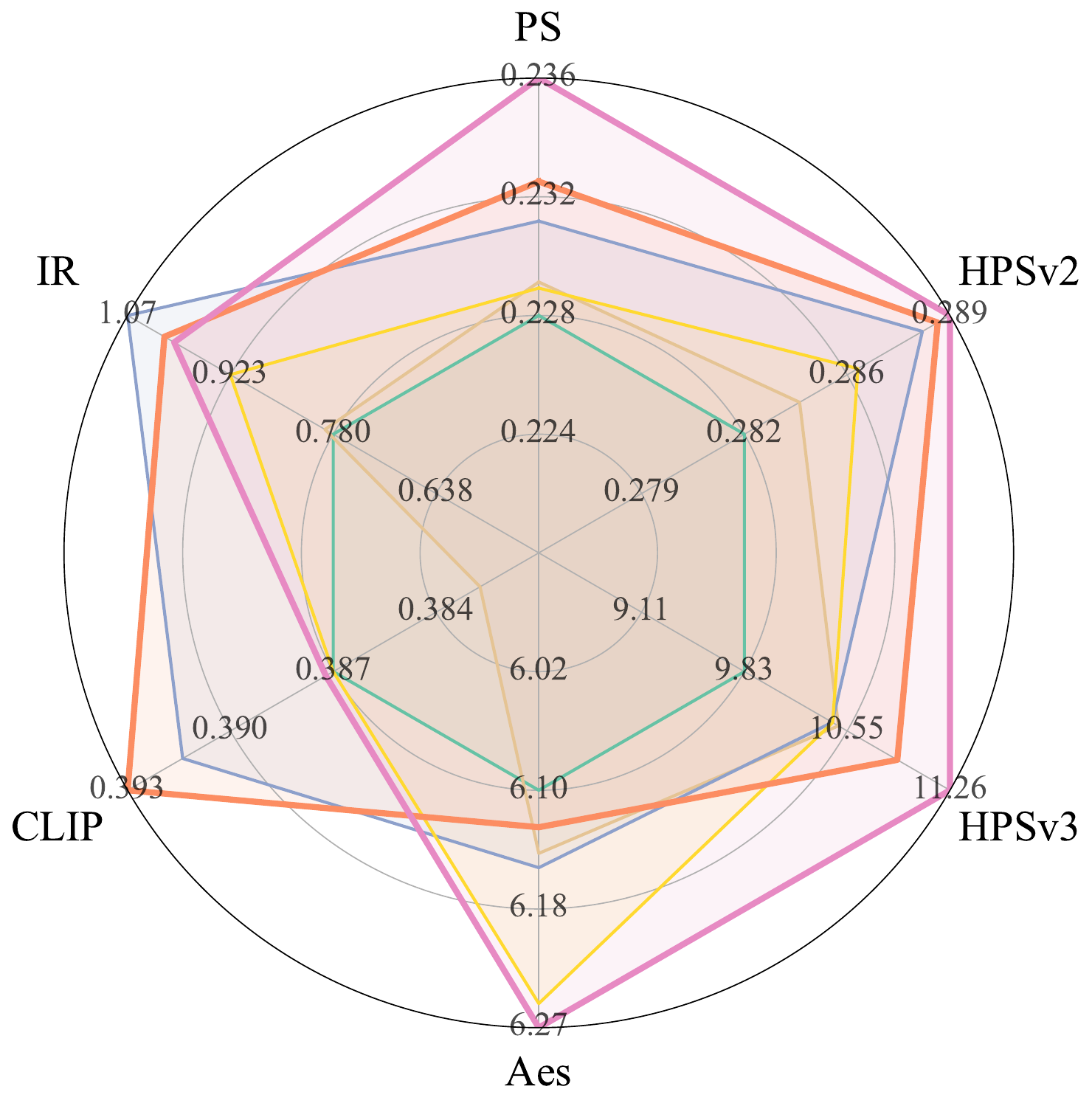} &
  \includegraphics[width=0.29\linewidth]{ 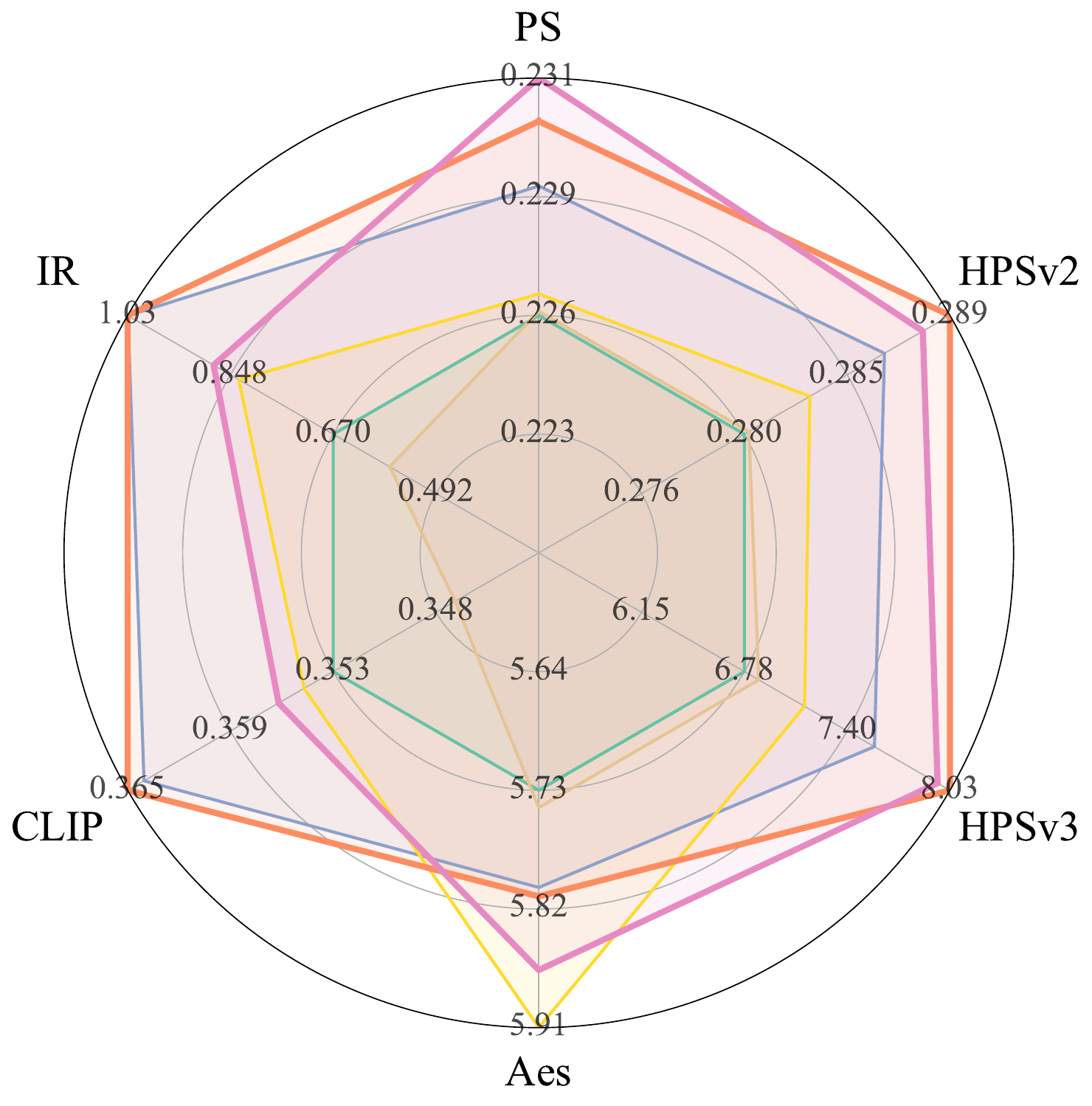} \\[1ex]

  \parbox[t]{0.03\linewidth}{\centering \raisebox{10ex}{\rotatebox{90}{SD1.5}}
} &
  \includegraphics[width=0.29\linewidth]{ 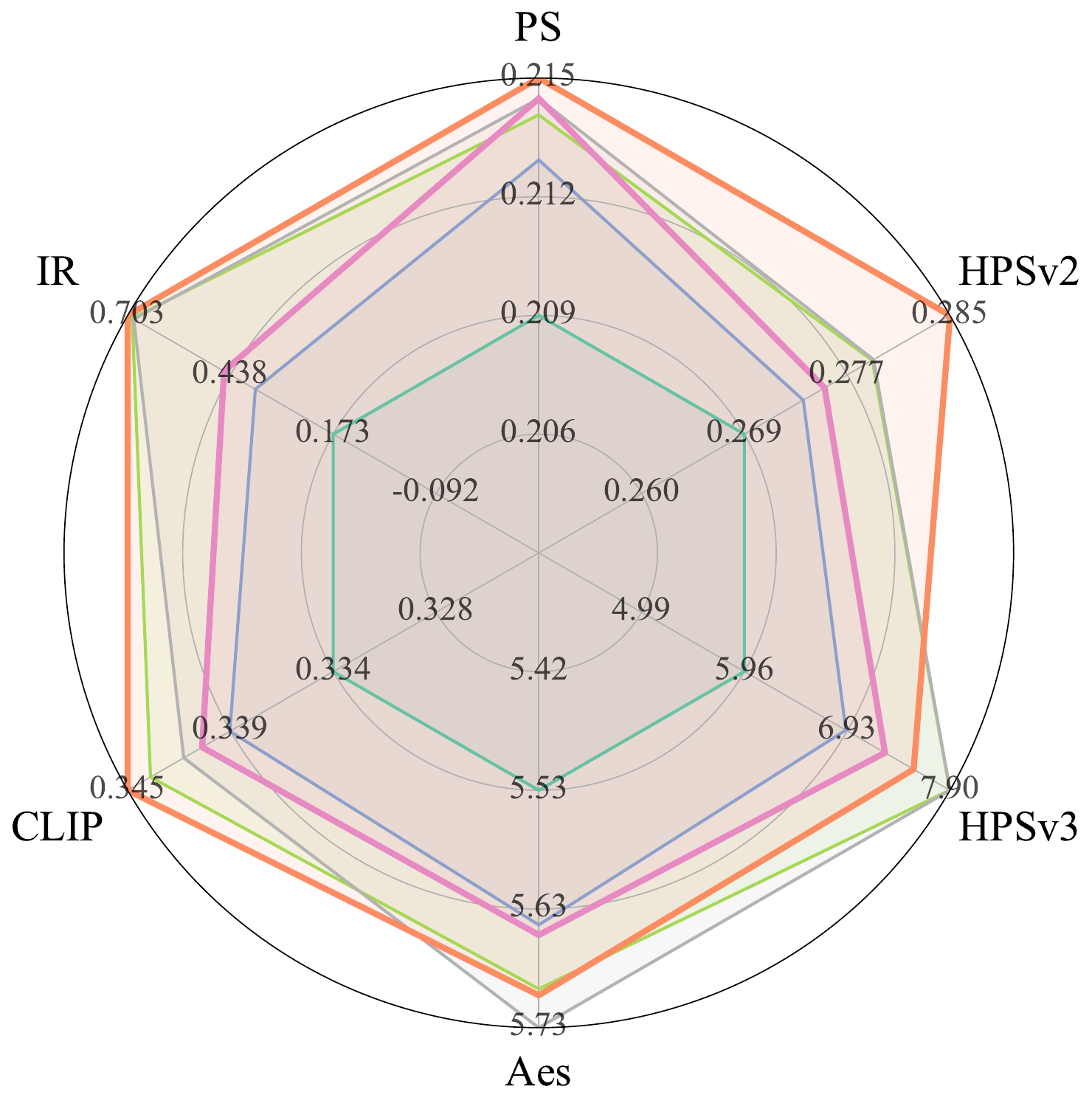} &
  \includegraphics[width=0.29\linewidth]{ 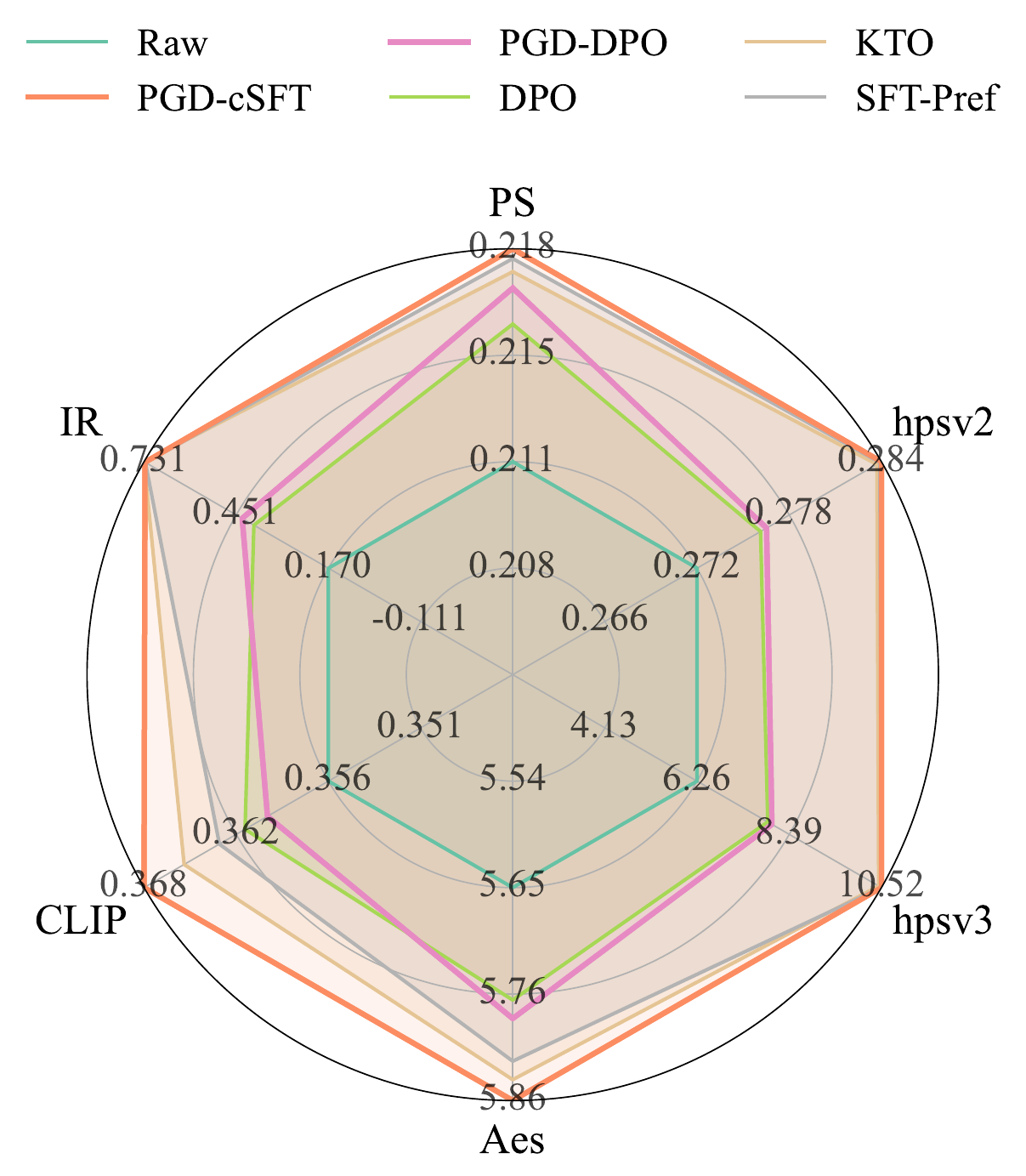} &
  \includegraphics[width=0.29\linewidth]{ 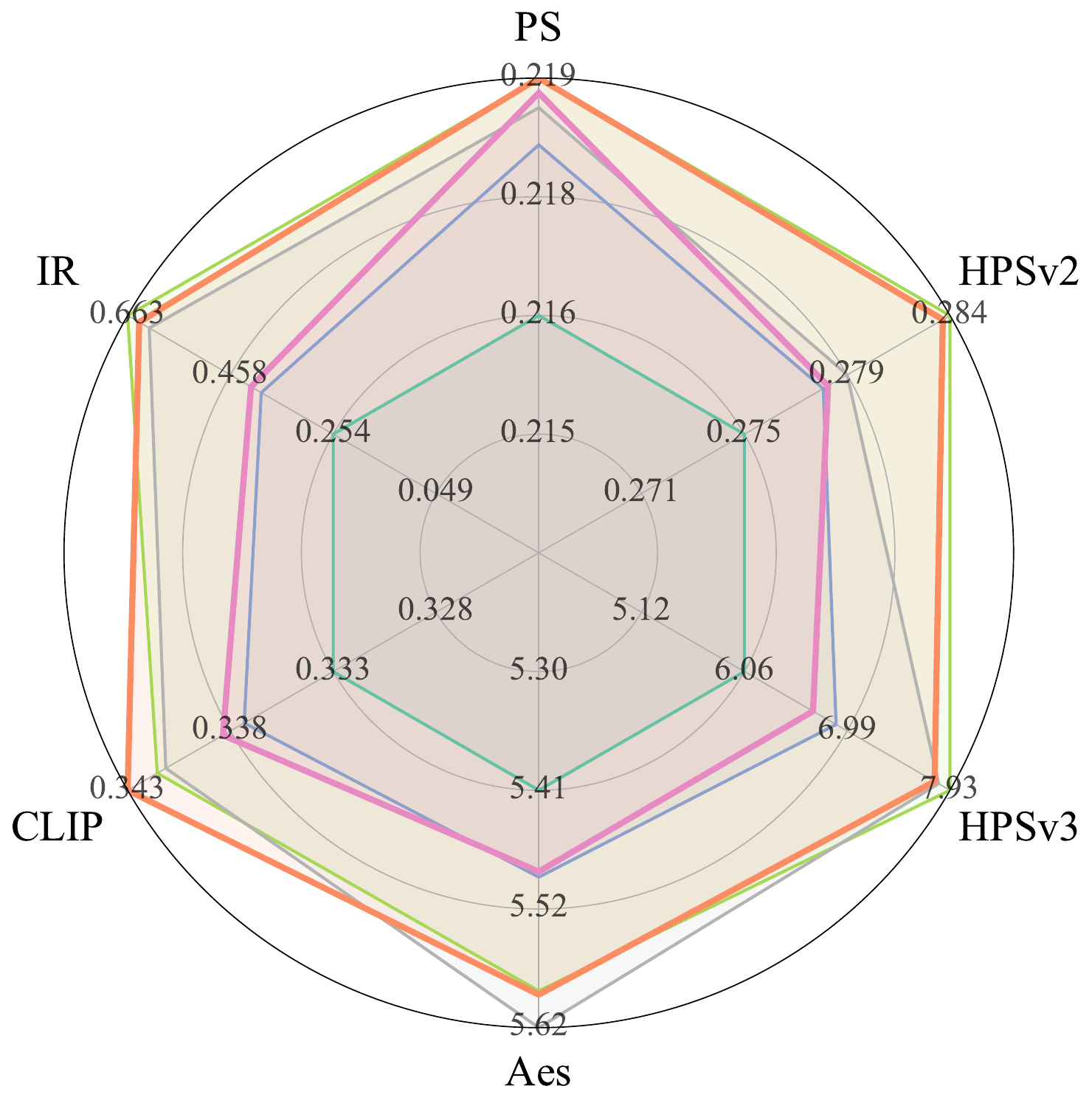} \\

  & Picapicv2 test & HPDv2 test & Parti-Prompts \\
\end{tabular}

\caption{ General Results with SDXL, SD1.5.}
\label{fig:general_results_appendix}
\end{figure*}

\begin{figure*}[ht!]
    \centering
    \includegraphics[width=1\textwidth]{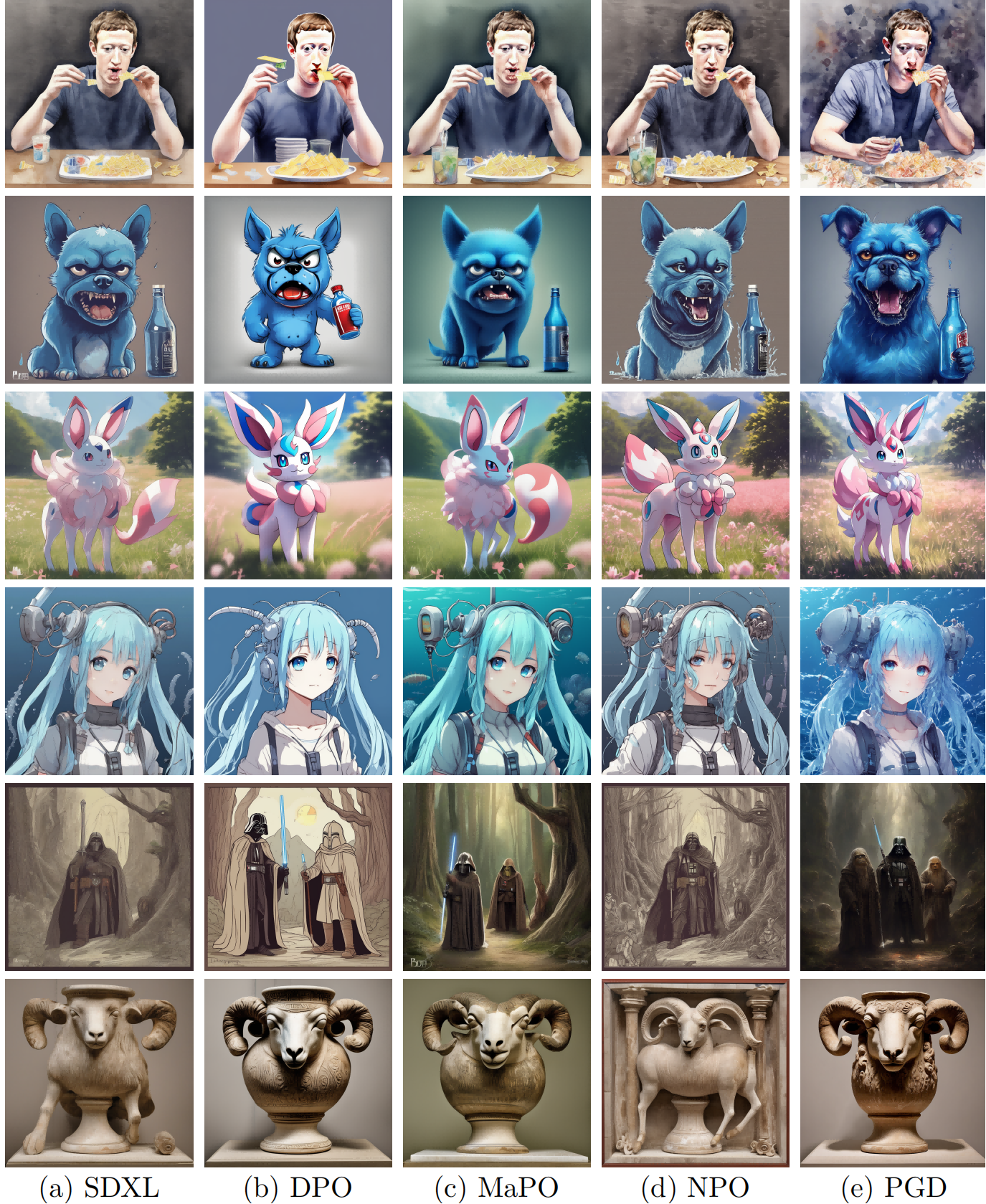}
    \caption{Comparison of preference-optimization methods on SDXL. Columns show outputs from the base model (SDXL), DPO, MaPO, NPO and PGD.}
    \label{fig:l1comparePGD}
\end{figure*}
\begin{figure*}[ht!]
    \centering
    \includegraphics[width=1\textwidth]{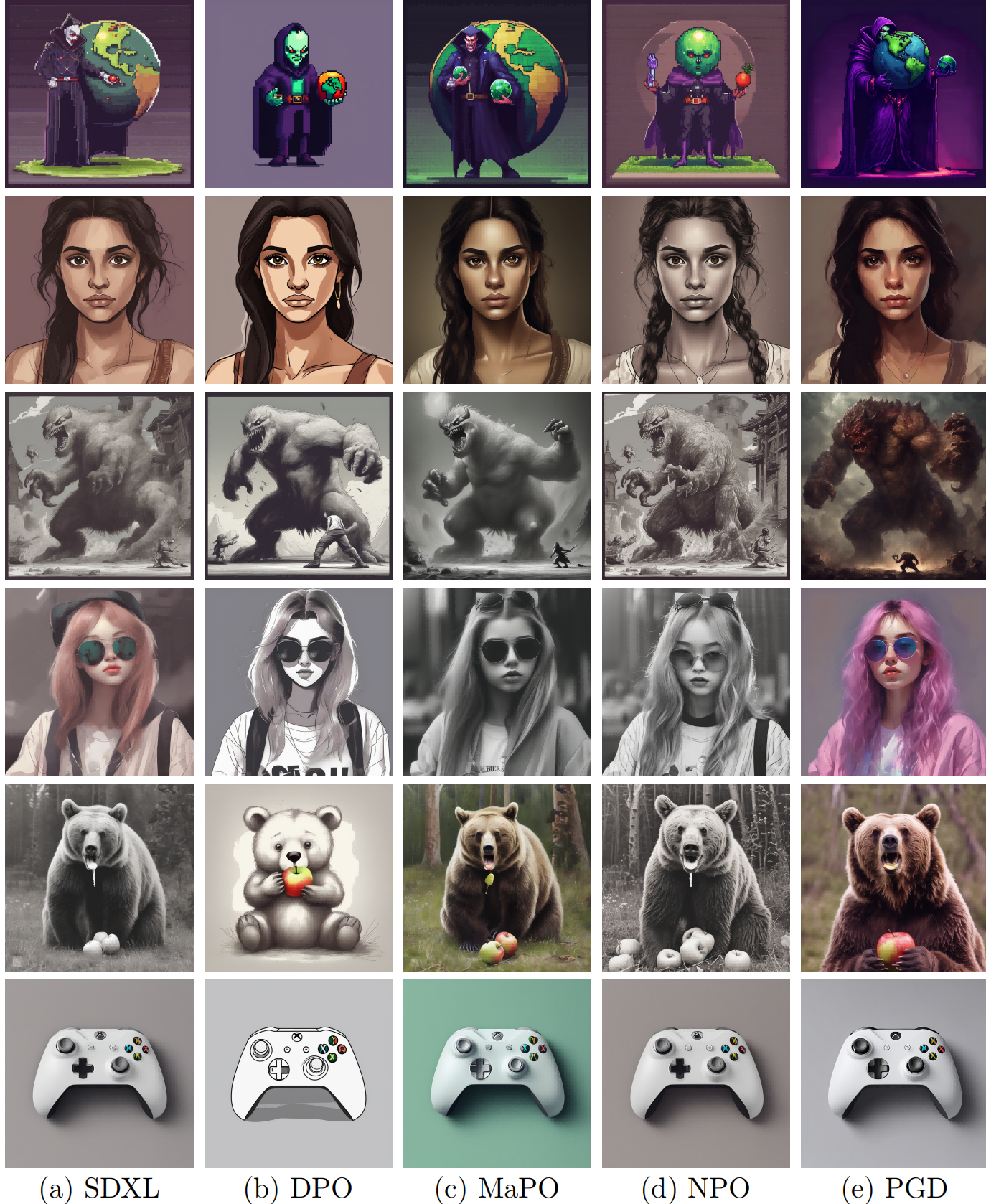}
    \caption{Comparison of preference-optimization methods on SDXL. Columns show outputs from the base model (SDXL), DPO, MaPO, NPO and PGD.}
    \label{fig:l2comparePGD}
\end{figure*}
\begin{figure*}[ht!]
    \centering
    \includegraphics[width=1\textwidth]{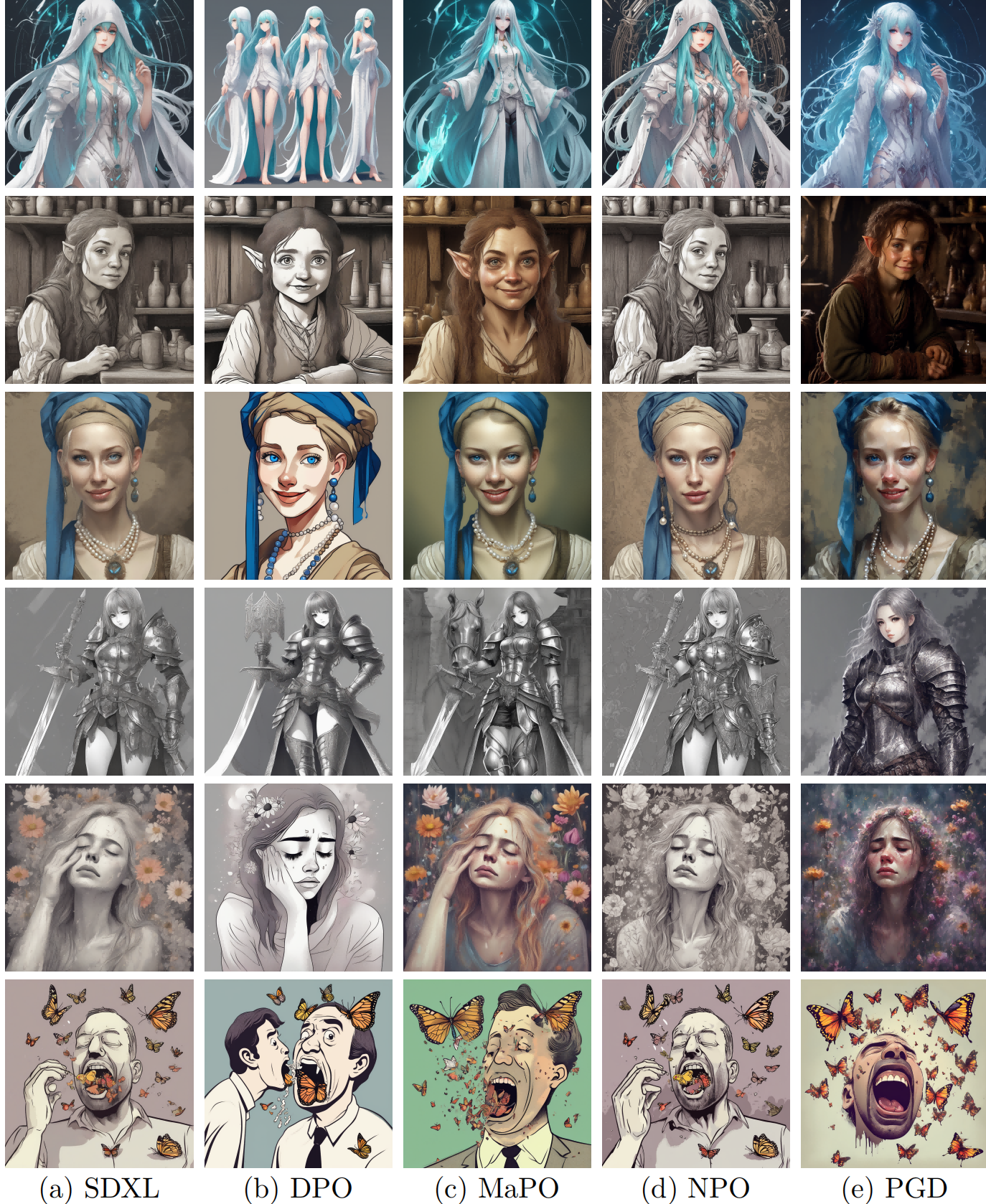}
    \caption{Comparison of preference-optimization methods on SDXL. Columns show outputs from the base model (SDXL), DPO, MaPO, NPO and PGD.}
    \label{fig:l3comparePGD}
\end{figure*}
\begin{figure*}[ht!]
    \centering
    \includegraphics[width=1\textwidth]{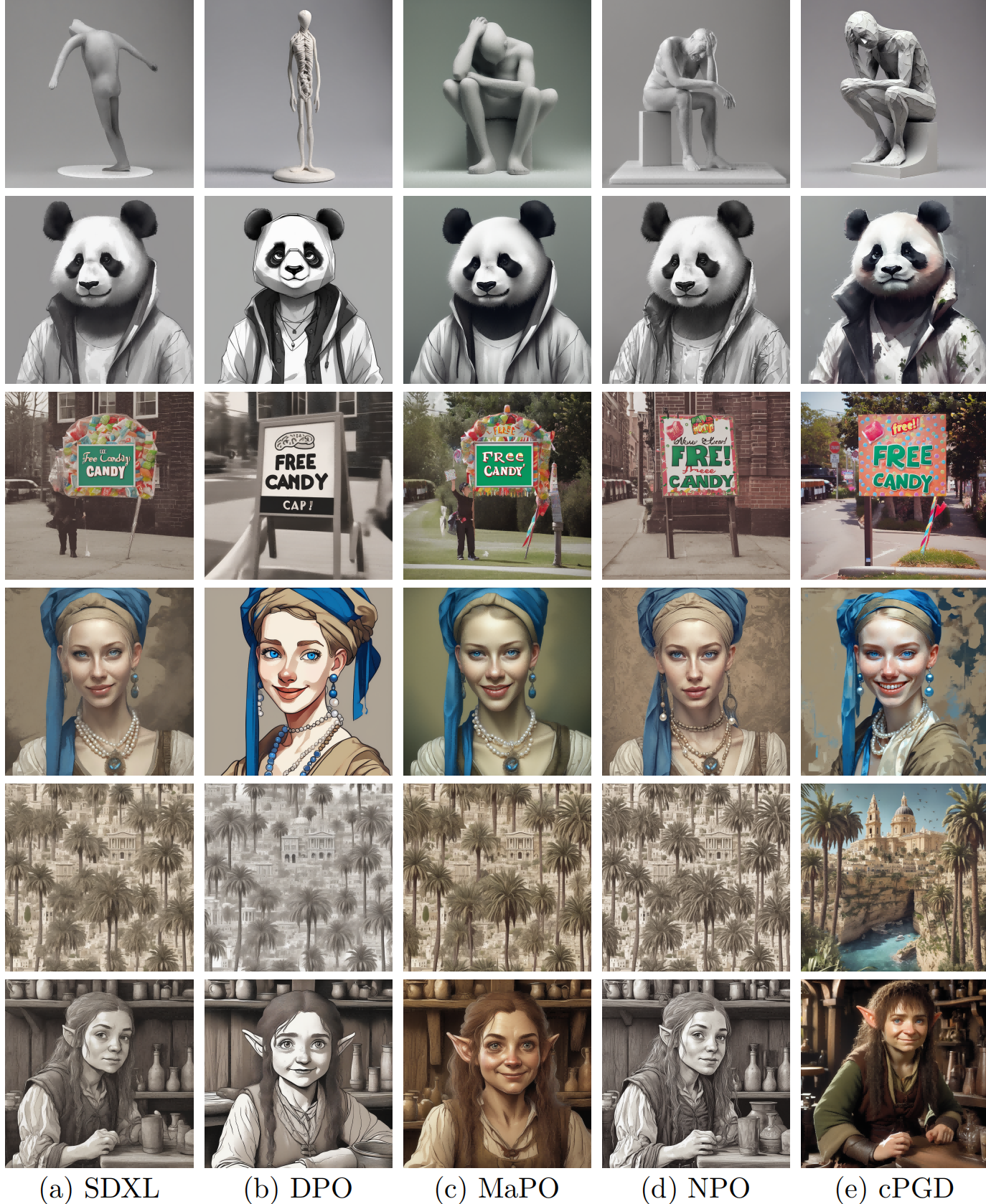}
    \caption{Comparison of preference-optimization methods on SDXL. Columns show outputs from the base model (SDXL), DPO, MaPO, NPO and cPGD.}
    \label{fig:ll1comparePGD}
\end{figure*}
\begin{figure*}[ht!]
    \centering
    \includegraphics[width=1\textwidth]{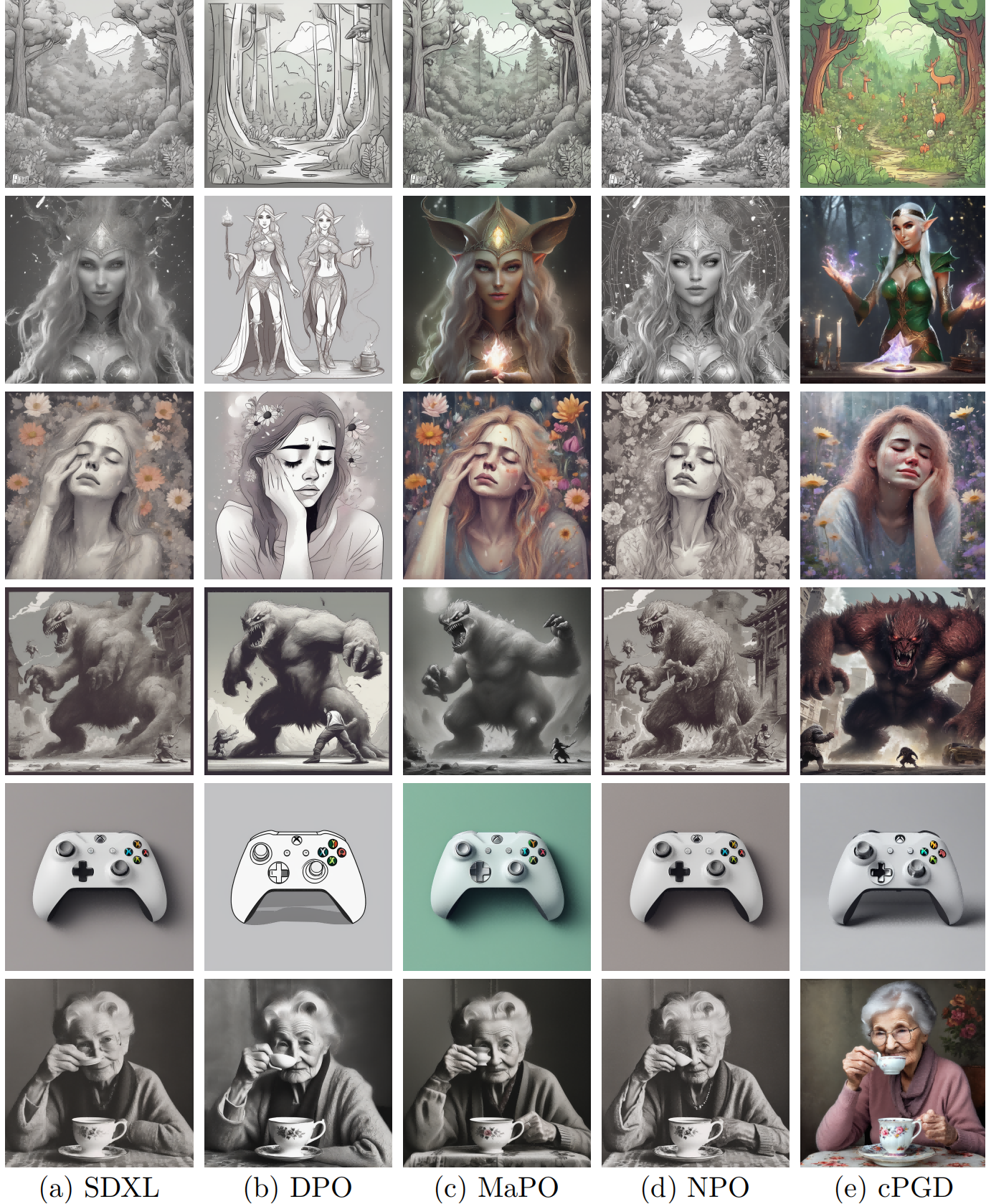}
    \caption{Comparison of preference-optimization methods on SDXL. Columns show outputs from the base model (SDXL), DPO, MaPO, NPO and cPGD.}
    \label{fig:ll2comparePGD}
\end{figure*}
\begin{figure*}[ht!]
    \centering
    \includegraphics[width=1\textwidth]{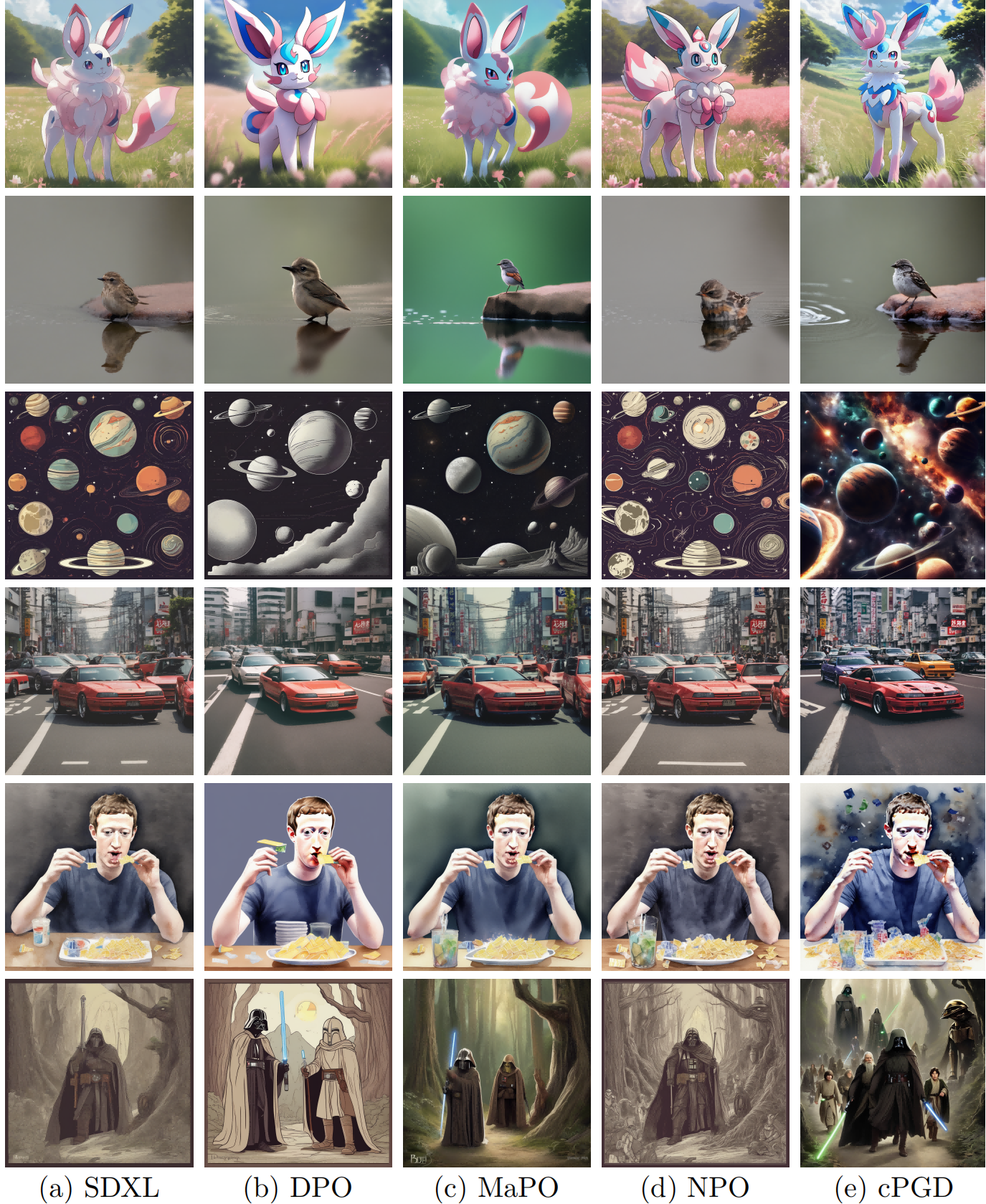}
    \caption{Comparison of preference-optimization methods on SDXL. Columns show outputs from the base model (SDXL), DPO, MaPO, NPO and cPGD.}
    \label{fig:ll3comparePGD}
\end{figure*}

Figures~\ref{fig:l1comparePGD}--\ref{fig:ll3comparePGD} provide additional side-by-side comparisons for PGD and cPGD against representative baselines across varied prompts. 

%


\end{document}